\documentclass{article}

\usepackage[preprint]{neurips_2026}

\usepackage[utf8]{inputenc} 
\usepackage[T1]{fontenc}    
\usepackage{hyperref}       
\usepackage{url}            
\usepackage{booktabs}       
\usepackage{amsfonts}       
\usepackage{nicefrac}       
\usepackage{microtype}      
\usepackage{xcolor}         

\usepackage{graphicx}
\usepackage{amsmath}
\usepackage{makecell}
\usepackage{pifont}
\usepackage{placeins}
\usepackage{threeparttable}
\usepackage{tikz}
\usepackage{xcolor}
\usepackage{graphicx} 
\usetikzlibrary{
    positioning,
    arrows.meta,
    shapes.geometric,
    fit
}
\usepackage{multirow}

\usepackage{tikz}
\usepackage{xcolor}
\usepackage{graphicx}
\usepackage{fontawesome5}

\usetikzlibrary{
    positioning,
    arrows.meta,
    shapes.geometric,
    shadows.blur,
    fit,
    calc
}

\colorlet{red}{black}
\colorlet{brown}{black}
\colorlet{blue}{black}
\colorlet{orange}{black}

\usepackage{tikz}
\usetikzlibrary{positioning, calc, arrows.meta}

\newcommand{\cmark}{\textcolor{green}{\ding{51}}} 
\newcommand{\xmark}{\textcolor{red}{\ding{55}}}   

\title{Towards Principled Continual Anomaly Detection: A Systematic Framework and Benchmark Scenarios}
\author{%
  Kamil Faber
    \\
  Faculty of Computer Science\\
  AGH University of Krakow \\
  Krakow, Poland \\
  \texttt{kfaber@agh.edu.pl} \\
  \And
  Mateusz Smendowski \\
  Faculty of Computer Science\\
  AGH University of Krakow \\
  Krakow, Poland \\
  \texttt{smendowski@agh.edu.pl} \\
  \AND
  Roberto Corizzo \\
  Department of Computer Science \\
  American University \\
  Washington DC, US \\
  \texttt{rcorizzo@american.edu} \\
}

\begin{document}

\maketitle

\begin{abstract}
{\color{blue}
Continual anomaly detection (CAD) studies how models can adapt to evolving data distributions while retaining performance on previously observed regimes. CAD benchmarks, however, depend critically on how tasks are defined, filtered, ordered, and validated. In tabular domains, task boundaries are rarely given, and arbitrary splits can create unlearnable, redundant, or overly transferable tasks that obscure genuine continual-learning behavior. To this end, we introduce a systematic framework for reproducible benchmark scenario design from existing tabular anomaly-detection datasets. The framework discovers candidate tasks, filters unsuitable tasks, and derives principled orderings that expose diverse dynamics. The framework allows us to deliver five benchmark-ready scenarios from three large-scale cybersecurity anomaly detection datasets, yielding both single-dataset and multi-dataset CAD settings. 

}
\end{abstract}

\section{Introduction}
{\color{blue}
Anomaly detection (AD) models are increasingly deployed in non-stationary environments, where the notion of normality itself shifts over time~\citep{lu2023review}. Continual anomaly detection (CAD) studies this setting by requiring detectors to adapt to emerging regimes while retaining useful knowledge about previously observed ones~\citep{faber2024lifelong}. CAD inherits specific challenges of anomaly detection: it is commonly one-class, unsupervised, or semi-supervised; anomalies are rare while normality itself drifts independently of the observed anomalies~\citep{pang2021deep}.


{
A key aspect to consider is that conventional anomaly detection datasets do not automatically yield strong continual scenarios: when tasks have low diversity, the induced task sequence may exhibit limited distributional separation, and high performance can partly reflect an easy or weakly differentiated scenario rather than a broadly solved CAD problem~\cite{dragoi2022anoshift,amalapuram2024spider}.
Unlike vision-based CAD, where object categories can provide semantically meaningful units of progression~\citep{lee2025continual}, tabular anomaly detection datasets rarely include validated task boundaries. As a result, CAD scenarios are often constructed from chronological windows, metadata fields, or manual partitions. These choices are convenient, but not necessarily valid, as different time periods may represent nearly identical normal regimes, while substantial distributional shifts may occur within a single period. Consequently, the resulting task sequence may fail to induce meaningful continual-learning dynamics~\citep{amalapuram2024spider,chin2024continual}.
%
%
%
Consequently, tabular CAD lacks reliable, reusable benchmark scenarios and principled mechanisms for transforming existing anomaly-detection datasets into validated CAD scenarios.
}

This paper addresses the scenario-design problem directly. Rather than treating task sequences as a minor experimental detail, we treat scenario construction as a first-class benchmark-design problem. Given one or more existing tabular anomaly-detection datasets, the framework discovers candidate tasks, evaluates their learnability and transfer structure with single-task experts, filters unsuitable tasks, and derives principled orderings that expose different continual-learning dynamics. 


Rather than providing an exhaustive leaderboard of CAD methods, this paper focuses on the benchmark-design problem itself. We introduce a systematic framework for constructing and validating CAD scenarios, and deliver five  single-dataset and multi-dataset benchmark scenarios.
%
%
 Each scenario is equipped with six orderings based on curriculum, generalization and drift levels.
%
Together, the framework and released scenarios provide a reproducible methodological foundation for future CAD benchmarking.
Our contributions are as follows: \\
\noindent - We formalize a \textbf{principled framework for reproducible benchmark-design}, which transforms existing tabular anomaly detection datasets into continual scenarios through candidate task discovery, single-task expert analysis, iterative task filtering, scenario ordering, scenario selection, and validation.

\noindent - We define \textbf{six principled task-ordering} families that expose complementary continual-learning dynamics from the same retained task set. Moreover, we define three scenario quality properties for CAD scenarios: feasibility, non-triviality, and forgetting. 

\noindent - We deliver \textbf{five CAD benchmark scenarios} from three large-scale tabular anomaly detection datasets from the cybersecurity domain, spanning within-dataset regime shifts and cross-dataset continual adaptation. We provide ready-to-use benchmark artifacts, including task splits, train-test splits, and final orderings, which establish a reproducible foundation for future CAD benchmarking.




}

\section{Related Works}

{\color{blue}
\textbf{Anomaly Detection Benchmarks}: Existing anomaly detection benchmarks have improved reproducibility by standardizing datasets, metrics, and evaluation protocols~\citep{han2022adbench,lu2023review,arodi2024cableinspect}. However, they primarily evaluate anomaly detectors under static, stream-oriented, or domain-specific protocols. They do not address the CAD-specific question of continual adaptation with knowledge retention.

\textbf{Continual Learning Benchmarks and Tasks}: 
Continual learning benchmarks have shown that benchmark design and evaluation protocols strongly determine which forms of forgetting, transfer, and adaptation become observable~\citep{lin2021clear}, arguing that the benchmark design is inseparable from the claims made about lifelong learning systems~\citep{parisi2019continual}. 
In domains such as image classification and reinforcement learning, CL benchmarks typically define progression through image classes, games, or robotic tasks, often relying on natural boundaries or artificial splits such as partitioning a class set into sequential tasks~\citep{lin2021clear,parisi2019continual,wang2024comprehensive}. Similarly, In vision-oriented CAD, task splits can often be grounded in semantically meaningful units such as object categories or product types. 
However, in tabular CAD, such splits are much less straightforward, creating a central challenge for constructing meaningful CL scenarios from existing datasets.

\textbf{Continual Anomaly Detection and Tabular CAD Scenarios}: 
Continual anomaly detection (CAD) has emerged as a distinct research area at the intersection of anomaly detection and continual learning~\citep{faber2024lifelong}. Existing work focus on both vision-oriented CAD, often built on image anomaly-detection benchmarks with semantically meaningful object, product, or defect structure~\citep{hu2025replaycad,lee2025continual}, and tabular CAD, where such structure is usually less explicit. Recent CAD methods study replay-based retention, prompt-based adaptation, task-aware parametrization, and cybersecurity-oriented continual training~\citep{faber2023vlad,channappayya2023augmented,liu2024unsupervised,zhang2025task,amalapuram2024spider,amalapuram2024soul}. However, the evaluation side of tabular CAD remains less mature than the methodological side. Many widely used tabular anomaly detection datasets were designed for static anomaly detection rather than CL evaluation. Consequently, they do not provide validated task boundaries, principled task orderings, or evidence that the induced tasks are learnable, non-redundant, and capable of exposing forgetting or transfer.
This creates a specific benchmark-design problem. In tabular CAD, a chronological window or data source may appear to define a natural task, but such partitions do not necessarily correspond to meaningful changes.
AnoShift~\citep{dragoi2022anoshift} is an important positive example, showing that time-based partitioning can yield meaningful continual scenarios when the temporal structure is sufficiently rich~\citep{amalapuram2024spider}.
 However, many commonly used intrusion-detection datasets, such as CICIDS2017 or CICIDS2018, cover much shorter collection periods and do not automatically provide the same level of validated time structure.
This issue was illustrated by~\citet{channappayya2023augmented,amalapuram2024spider}, who observed that time-based splits can exhibit limited task diversity.
A different approach was proposed by \citep{faber2024lifelong}, where clustering-based options were leveraged for constructing continual scenarios from originally non-continual data \citep{faber2024lifelong}. However, this work does not provide any further validation of created tasks  to guarantee challenging scenarios. 

These issues motivate the need for a more comprehensive framework in which task discovery is only one component, complemented by explicit filtering, ordering, and validation criteria. To this end, we propose a framework that addresses that gap by making scenario construction a principled and reproducible part of CAD evaluation. Moreover, we deliver five CAD benchmark scenarios from three large-scale intrusion detection datasets.

}

\section{A Framework for CAD Scenario Benchmark Design}
{\color{blue}

{\color{brown}
We propose a framework for transforming one or more existing tabular anomaly-detection datasets into validated continual anomaly detection (CAD) scenarios. Designed for datasets that were not originally collected as continual-learning benchmarks, the framework does not merely split a dataset into arbitrary tasks, it constructs CAD scenarios whose tasks are empirically meaningful, suitable for anomaly detection, and organized according to an explicit evaluation rationale.
The framework supports both single-dataset and multi-dataset scenarios, enabling within-dataset regime shifts as well as cross-dataset continual adaptation. Its output is not an arbitrary split, but a set of reusable CAD scenarios with empirically validated task structure, principled orderings, task statistics, and validation results.
A graphical overview of our framework is shown in Figure \ref{fig:methodology-overview}. To support the reader, we also provide a notation reference in Appendix~\ref{app:notation}.
Our code is available at: \url{https://github.com/lifelonglab/CAD-Benchmarks-Framework}. The benchmark scenarios are available at: \url{https://huggingface.co/collections/lifelonglab/tabular-cad-benchmarks}. 
}

}



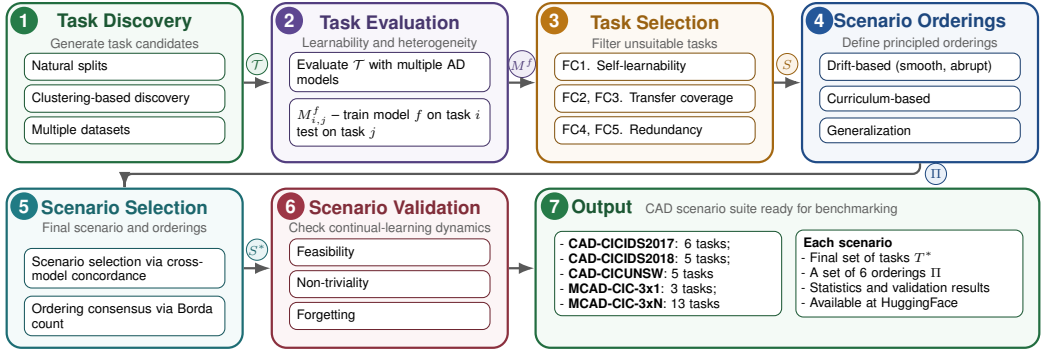
\begin{figure*}[t]
\centering
\resizebox{0.98\textwidth}{!}{%
\begin{tikzpicture}[
    font=\sffamily\footnotesize,
    >=Latex,
    stage/.style={
        draw=#1!75!black,
        fill=#1!5,
        rounded corners=8pt,
        very thick,
        minimum width=4.45cm,
        minimum height=3cm,
        text width=4cm,
        align=center,
        inner sep=6pt
    },
    stagewide/.style={
        draw=#1!75!black,
        fill=#1!5,
        rounded corners=8pt,
        very thick,
        minimum width=9.5cm,
        minimum height=3cm,
        text width=8cm,
        align=left,
        inner sep=6pt
    },
    card/.style={
        draw=#1!60!black,
        fill=white,
        rounded corners=4pt,
        line width=0.55pt,
        text width=3.5cm,
        align=left,
        inner sep=4pt,
        font=\sffamily\scriptsize
    },
    cardc/.style={
        draw=#1!60!black,
        fill=white,
        rounded corners=4pt,
        line width=0.55pt,
        text width=3.8cm,
        align=center,
        inner sep=4pt,
        font=\sffamily\scriptsize
    },
    mini/.style={
        draw=#1!60!black,
        fill=white,
        rounded corners=4pt,
        line width=0.55pt,
        text width=1.40cm,
        minimum height=0.75cm,
        align=center,
        inner sep=4pt,
        font=\sffamily\scriptsize
    },
    num/.style={
        circle,
        fill=#1!85!black,
        text=white,
        font=\bfseries\sffamily,
        minimum size=0.52cm,
        inner sep=0pt
    },
    badge/.style={
        circle,
        draw=#1!70!black,
        fill=#1!10,
        text=#1!70!black,
        font=\bfseries\sffamily\scriptsize,
        minimum size=0.40cm,
        inner sep=0pt
    },
    title/.style={
        font=\bfseries\sffamily\normalsize,
        text=black!85,
        align=center
    },
    subtitle/.style={
        font=\sffamily\scriptsize,
        text=black!62,
        align=center
    },
    connector/.style={
        ->,
        very thick,
        draw=black!65,
        rounded corners=4pt
    }
]

\definecolor{greenC}{RGB}{45,145,85}
\definecolor{purpleC}{RGB}{111,84,170}
\definecolor{orangeC}{RGB}{219,139,24}
\definecolor{blueC}{RGB}{42,102,180}
\definecolor{tealC}{RGB}{38,145,155}
\definecolor{redC}{RGB}{185,64,82}


\node[stage=greenC] (disc) at (0,0) {};
\node[num=greenC] at ($(disc.north west)+(0.35,-0.35)$) {1};

\node[title, text=greenC!60!black] at ($(disc.north)+(0,-0.38)$)
{Task Discovery};

\node[subtitle] at ($(disc.north)+(0,-0.78)$)
{Generate task candidates};

\node[badge=greenC] at ($(disc.east)+(0.25, 0.35)$) {$\mathcal{T}$};

\node[card=greenC] at ($(disc.center)+(0,0.3)$)
{{Natural splits}};

\node[card=greenC] at ($(disc.center)+(0,-0.3)$)
{{Clustering-based discovery}};

\node[card=greenC] at ($(disc.center)+(0,-0.9)$)
{{Multiple datasets}};


\node[stage=purpleC] (eval) at (5,0) {};
\node[num=purpleC] at ($(eval.north west)+(0.35,-0.35)$) {2};

\node[title, text=purpleC!65!black] at ($(eval.north)+(0,-0.38)$)
{Task Evaluation};

\node[subtitle] at ($(eval.north)+(0,-0.78)$)
{Learnability and heterogeneity};

\node[badge=purpleC] at ($(eval.east)+(0.25,0.35)$) {$M^f$};

\node[card=purpleC] at ($(eval.center)+(0,0.2)$)
{Evaluate $\mathcal{T}$ with multiple AD models };

\node[card=purpleC] at ($(eval.center)+(0,-0.75)$)
{
$M^f_{i,j}$ -- train model $f$ on task $i$ test on task $j$\\[4pt]
};


\node[stage=orangeC] (select) at (10,0) {};
\node[num=orangeC] at ($(select.north west)+(0.35,-0.35)$) {3};

\node[title, text=orangeC!70!black] at ($(select.north)+(0,-0.38)$)
{Task Selection};

\node[subtitle] at ($(select.north)+(0,-0.78)$)
{Filter unsuitable tasks};

\node[badge=orangeC] at ($(select.east)+(0.25,0.35)$) {$S$};

\node[card=orangeC] at ($(select.center)+(0,0.3)$)
{FC1. Self-learnability};

\node[card=orangeC] at ($(select.center)+(0,-0.3)$)
{FC2, FC3. Transfer coverage};

\node[card=orangeC] at ($(select.center)+(0,-0.9)$)
{FC4, FC5. Redundancy};



\node[stage=blueC] (construct) at (15,0) {};
\node[num=blueC] at ($(construct.north west)+(0.35,-0.35)$) {4};

\node[title, text=blueC!70!black] at ($(construct.north)+(0,-0.38)$)
{Scenario Orderings};

\node[subtitle] at ($(construct.north)+(0,-0.78)$)
{Define principled orderings};

\node[badge=blueC] at ($(construct.south)+(0.3,-0.2)$) {$\Pi$};

\node[card=blueC] at ($(construct.center)+(0,0.3)$)
{Drift-based (smooth, abrupt)};

\node[card=blueC] at ($(construct.center)+(0,-0.3)$)
{Curriculum-based};

\node[card=blueC] at ($(construct.center)+(0,-0.9)$)
{Generalization};


\node[stage=tealC] (rank) at (0,-3.5) {};
\node[num=tealC] at ($(rank.north west)+(0.35,-0.35)$) {5};

\node[title, text=tealC!70!black] at ($(rank.north)+(0,-0.38)$)
{Scenario Selection};

\node[subtitle] at ($(rank.north)+(0,-0.78)$)
{Final scenario and orderings};

\node[badge=tealC] at ($(rank.east)+(0.25,0.35)$) {$S^{*}$};


\node[card=tealC] at ($(rank.center)+(0,0)$)
{Scenario selection via cross-model concordance};

\node[card=tealC] at ($(rank.center)+(0,-0.9)$)
{Ordering consensus via Borda count};

\node[stage=redC] (valid) at (5,-3.5) {};
\node[num=redC] at ($(valid.north west)+(0.38,-0.35)$) {6};

\node[title, text=redC!70!black] at ($(valid.north)+(0.1,-0.38)$)
{Scenario Validation};

\node[subtitle] at ($(valid.north)+(0,-0.78)$)
{Check continual-learning dynamics};



\node[card=redC] at ($(valid.center)+(0,0.3)$)
{
Feasibility
};

\node[card=redC] at ($(valid.center)+(0,-0.3)$)
{
Non-triviality
};

\node[card=redC] at ($(valid.center)+(0,-0.9)$)
{
Forgetting
};












\node[stagewide=greenC] (out) at (12.5,-3.5) {};
\node[num=greenC] at ($(out.north west)+(0.38,-0.35)$) {7};

\node[title, text=greenC!60!black] at ($(out.north)+(-3.5,-0.38)$)
{Output};

\node[subtitle] at ($(out.north)+(-0.3,-0.42)$)
{CAD scenario suite ready for benchmarking};


\node[card=greenC, text width=4cm] at ($(out.center)+(-2.25,-0.1)$)
{
- \textbf{CAD-CICIDS2017}: 6 tasks; \\
- \textbf{CAD-CICIDS2018}: 5 tasks; \\ 
- \textbf{CAD-CICUNSW}: 5 tasks  \\ 
- \textbf{MCAD-CIC-3x1}: 3 tasks; \\
- \textbf{MCAD-CIC-3xN}: 13 tasks};

\node[card=greenC, text width=4cm] at ($(out.center)+(2.3,-0.1)$)
{\textbf{Each scenario}\\
- Final set of tasks $T^{*}$ \\
- A set of 6 orderings $\Pi$ \\
- Statistics and validation results \\
- Available at HuggingFace};

\draw[connector] (disc.east) -- (eval.west);
\draw[connector] (eval.east) -- (select.west);

\draw[connector] (select.east) -- (construct.west);
\draw[connector] (construct.south) --  ++(0,-0.2) -| ($(rank.north)+(0,0.15)$) -- (rank.north);
\draw[connector] (rank.east) -- (valid.west);

\draw[connector] (valid.east) -- (out.west);

\end{tikzpicture}%
}
\caption{Detailed overview of the proposed framework for transforming anomaly detection datasets into validated continual anomaly detection scenarios.}
\label{fig:methodology-overview}
\end{figure*}

\subsection{Task discovery}
{\color{blue}

{\color{red}
Task discovery generates candidate task sets from one or more preprocessed tabular anomaly-detection datasets. 
Let
$\mathcal{D}=\{D^{(1)},\ldots,D^{(R)}\}$ denote the available datasets, where each
$D^{(r)}=\{(x_i,y_i)\}_{i=1}^{N_r}$ and $y_i\in\{0,1\}$ indicates normal or anomalous samples.
Each candidate task set $T_i=\{\tau_1,\ldots,\tau_K\}$ is built on top of tasks consisting of a regime of normal data together with the anomaly samples in the test subset:
$
\tau_k =
\left(
D^{\mathrm{norm}}_{k,\mathrm{train}},
D^{\mathrm{norm}}_{k,\mathrm{test}},
D^{\mathrm{anom}}_{k,\mathrm{test}}
\right)
$. 
The key requirement is that candidate tasks should correspond to materially distinct learning conditions rather than arbitrary slices of the dataset. 
Additional formal details and examples are provided in Appendix~\ref{app:task-discovery-details}.

We consider three task-discovery mechanisms:\\
- \textbf{Natural split}, applicable when the dataset provides an inherent temporal, contextual, or acquisition-based structure; for example, CICIDS2017 can be partitioned by capture day. \\
- \textbf{A family of clustering-based splits}, where task structure is inferred from the data distribution. We consider three variants: jointly clustering normal and anomalous samples, clustering only normal samples and assigning anomalies to the nearest induced cluster, and clustering only normal samples while distributing anomalies randomly. We provide more details in Appendix~\ref{appendix:sec:clustering}. \\
- \textbf{Multi-dataset} option, constructing the scenario on top of multiple datasets, where each dataset can be treated as a single task or decomposed into multiple tasks.


}

{\color{brown}
}


\subsection{Tasks evaluation and selection}
{\color{blue}
Task discovery produces candidate task sets, but neither natural boundaries nor clustering-based partitions guarantee that the resulting tasks are meaningful for continual anomaly detection. We therefore evaluate each candidate task set to ensure that tasks are learnable, sufficiently distinct, and not redundant.\\
%
%
\textbf{Task evaluation}
We assess task heterogeneity through single-task experts (STEs). For each candidate task and each model $f \in F$, we train an STE on that task and evaluate it on all candidate tasks, yielding a cross-task performance matrix $M^f$ where $M^f_{i,j}$ denotes training on task $i$ and evaluation on task $j$. This empirical view exposes learnability, transfer, and redundancy directly. The models used at this stage are distinct from those used later in scenario validation (see Appendix~\ref{appendix:sec:models}).

\textbf{Tasks Selection}
%
We define five filtering criteria (FCs) to exclude candidate tasks that could lead to misleading evaluation: tasks that are not learnable, tasks that are too easy because they are covered or dominated by many other tasks, and tasks whose transfer profiles are redundant. Because task removal changes the cross-task transfer structure, filtering is performed iteratively: at each pass, we recompute all criteria, remove the task flagged by the largest number of criteria, and stop only when no remaining task violates any criterion. \\
\textbf{FC1. Self-learnability} - \textit{retaining only tasks that are individually learnable.} \\
A candidate task $k$ is retained only if it is sufficiently learnable in isolation by at least one model $f$:
\begin{equation}
    \exists f \in F  \quad M^f_{k,k} - b_k > \gamma_l
\end{equation}
where $\gamma_l$ denotes the minimum acceptable gain above the random baseline defined as $b_k$. 
\\
\textbf{FC2. Limited incoming transfer coverage} - \textit{retaining only tasks posing a challenge for other STEs}\\
A task $k$ should not be solvable by too many STEs trained on other tasks, as it would make the task $k$ trivial. Since we consider multiple models, we first define the best specialist performance for task $k$ across all models:
$S_k^{\star} = \max_{f \in F} M^f_{k,k}.$
Then, for each model $f \in F$, we require the number of tasks whose STEs cover task $k$ at at least a fraction $\gamma_t$ of that best specialist to be less than $P_t$. Formally:
\begin{equation}
\max_{f \in F} |\mathcal{C}_k^f| \leq P_t, \quad \text{where } \quad
    \mathcal{C}_k^f =
    \left\{
        i \in \{1,\dots,K\} \setminus \{k\}
        \;\middle|\;
        \frac{M^f_{i,k}}{S_k^{\star}} \geq \gamma_t
    \right\}.
\end{equation}
Here, $\gamma_t$ is the retained-performance threshold (e.g., $\gamma_t = 0.9$ for 90\% of the best specialist), and $P_t$ is the maximum acceptable number of other tasks that may cover task $k$ under any model. 
%
\\
\textbf{FC3. Limited outgoing transfer coverage} - \textit{filtering tasks that lead to trivial scenario} \\
Learning task $k$ should not lead to a model being able to solve too many tasks, as this would lead to a trivial scenario. 
To this end, an STE trained on task $k$ should not achieve performance on other tasks $i, i !=k$ similar to STE's trained on the specific task $i$. 
More formally, we define:
%
\begin{equation}
\max_{f \in F} |\mathcal{D}_k^f| \leq P_d, \quad \text{where } \quad
    \mathcal{D}_k^f =
    \left\{
        i \in \{1,\dots,K\} \setminus \{k\}
        \;\middle|\;
        \frac{M^f_{k,i}}{S_i^{\star}} \geq \gamma_d
    \right\}.
\end{equation}
where $\gamma_d$ is a relative dominance threshold and $P_d$ is the maximum acceptable number of tasks that task $k$ may dominate. 
%
\\
\textbf{FC4. Source profile redundancy} - \textit{filtering tasks with nearly identical source transfer profile} \\
Two tasks $k$ and $j$ are considered redundant as sources if their transfer behavior across other tasks is similar:
\begin{equation}
    \sum_{f \in F} \frac{\frac{1}{K}\sum_{m=1}^{K} |M^f_{k,m} - M^f_{j,m}|}{|F|} < \delta_r,
\end{equation}
where $\delta_r$ is the maximum allowed average difference between the two transfer profiles.
%
\\
\textbf{FC5. Target profile redundancy} - \textit{filtering tasks with nearly identical target transfer profile} \\
Similarly, two tasks $k$ and $j$ are considered redundant as targets if they are solved similarly by all models. 
\begin{equation}
    \sum_{f \in F} \frac{\frac{1}{K}\sum_{m=1}^{K} |M^f_{m,k} - M^f_{m,j}|}{|F|} < \delta_r
\end{equation}


\subsection{Scenario orderings}

{\color{blue}
Task selection stage creates multiple scenario candidates $\mathcal{S} = \{s_1, \ldots, s_m\}$, where each scenario candidate $s_i$ corresponds to the specific set of tasks $T_i$. While task selection determines the presence of a meaningful cross-task structure, ordering tasks within the scenario determines how this structure is exposed to the model, directly influencing the observed continual learning dynamics \cite{faber2024mnist}.
The same set of tasks can induce different amounts of forgetting, transfer, and adaptation difficulty depending on the sequence in which the model encounters them. 
To this end, we define a family of principled ordering strategies $\Pi$, each inducing different continual learning dynamics.
We compute orderings independently for each model $f$, yielding model-specific ordering families $\Pi^f$. Final benchmark orderings are obtained only after cross-model aggregation in the scenario selection stage, reducing dependence on any single detector's inductive bias.

\textbf{1. Drift-based ordering}:
Given an STE task performance matrix $M^f$, \textbf{a smooth-drift ordering 
$\pi^{f}_{\mathrm{SD}}$} is obtained by minimizing the cumulative distance between consecutive tasks, whereas \textbf{abrupt-drift ordering} $\pi^{f}_{\mathrm{AD}}$ maximizes large shifts between consecutive tasks. More formally:
\begin{equation}
    \pi^{f}_{\mathrm{SD}}
    =
    \arg\min_{\pi}
    \sum_{t=1}^{K-1} D_{\pi_t,\pi_{t+1}},
    \ \  
        \pi^{f}_{\mathrm{AD}}
    =
    \arg\max_{\pi}
    \sum_{t=1}^{K-1} D_{\pi_t,\pi_{t+1}},
    \ \
    \text{where} \ D_{i,j} = 1 - \frac{M^f_{i,j} + M^f_{j, i}}{2}.
\end{equation}

\textbf{2. Curriculum ordering}: Curriculum orderings present tasks by increasing or decreasing empirical difficulty, testing whether models benefit from structured task progression \cite{faber2024mnist}. We use STE self-performance as a difficulty proxy, $d_i = 1 - M^f_{i,i}$, and define
%
an ascending curriculum ordering $\pi_C$ and descending curriculum ordering $\pi_{RC}$ as:
\begin{equation}
    \pi^f_{C}:  d_{\pi_1} \leq d_{\pi_2} \leq \dots \leq d_{\pi_K},
    \quad
    \pi^f_{RC}:  d_{\pi_1} \geq d_{\pi_2} \geq \dots \geq d_{\pi_K},
\end{equation}
%

\textbf{3. Generalization-based ordering}: Generalization orderings rank tasks based on the ability of an STE trained on task $i$ to generalize to other tasks. In contrast to curriculum ordering, which assesses the difficulty of the task, this criterion captures the overall usefulness of a task as a source of transferable knowledge across the entire task set. 
We define an ordering by increasing generalization $\pi_{GI}$  and decreasing generalization $\pi_{GD}$ leveraging the average off-diagonal performance of its STE $g_i$:
\begin{equation}
    \pi^f_{GI}: g_{\pi_1} \leq \dots \leq g_{\pi_K}, 
    \quad
    \pi^f_{GD}: g_{\pi_1}  \geq \dots \geq g_{\pi_K},
    \quad
    \text{where} \ 
        g_i =
    \frac{1}{K-1}
    \sum_{\substack{j=1 \\ j \neq i}}^{K}
    M^f_{i,j},
\end{equation}
%
{
These ordering strategies allow the benchmark to separate different continual learning questions. Smooth-drift orderings test gradual adaptation, abrupt-drift orderings stress stability under large regime changes, curriculum orderings test whether methods benefit from increasing difficulty, and generalization-based orderings expose whether early broad tasks reduce or obscure later forgetting.
}


\subsection{Final scenario and ordering selection}

After the previous stages, we have scenario candidates $s_1,\dots,s_m$ together with model-specific orderings $\Pi^f$ for each $s_i$. 
We then select the scenario whose ordering structure is most consistent across models and aggregate the resulting orderings into a final consensus.

\textbf{Scenario Selection via Cross-Model Rank Concordance.}
For each scenario candidate, we quantify cross-model agreement over the induced orderings using Kendall's $W$ \citep{field2005k}. We then select the scenario whose ordering structure yields the highest mean concordance across the ordering families. Full details are provided in Appendix~\ref{app:scenario-selection-details}. \\
\textbf{Final Consensus Ordering via Borda Count.}
Given the selected split $s^*$, we aggregate the per-model orderings using Borda count \citep{rothe2019borda}. This yields a single consensus ordering for each of six ordering families by ranking tasks according to their aggregate model ranks. Full details of this process are provided in Appendix~\ref{app:scenario-selection-details}.

\subsection{Scenario Validation}
\label{sec:scenario-validation}
Scenario validation evaluates the complete task sequence after discovery, filtering, ordering, and scenario selection. Our purpose in this study is not to rank CAD methods exhaustively, but to verify that the accepted scenario is useful as a benchmark.


{\color{red}



We use four diagnostic reference strategies. \emph{Naive} denotes sequential training without an explicit retention mechanism. \emph{Cumulative} updates models using all data observed so far and serves as a non-continual reference with access to past data. \emph{MSTE} is an oracle pool of single-task experts, one per task, and measures task-specific learnability limiting forgetting. \emph{Replay} augments sequential training with stored samples from previous tasks and serves as a practical retention-based continual-learning reference.
We analyze each scenario through aggregate performance across all tasks, using metrics such as ROC-AUC and normalized PR-AUC, together with the following criteria. More details about the evaluation protocol can be found in Appendix~\ref{appendix:sec:evaluation_protocol}. 
We do not define universal pass/fail thresholds, since the required levels of feasibility, non-triviality, and forgetting depend on the dataset, domain, and intended use. Users of the framework should therefore report these diagnostics explicitly and justify scenario acceptance based on the three criteria below.


- \textbf{Feasibility}: A scenario should be learnable by at least one strong reference strategy, such as MSTE or Cumulative. This verifies that the tasks are not intrinsically degenerate and that meaningful detection performance is achievable when forgetting is controlled or past data are available.\\
%
- \textbf{Non-triviality}: A valid continual scenario should not be solved by Naive sequential training alone. We assess this through the gap between Naive and stronger references such as Replay, Cumulative, and MSTE, verifying that the scenario is learnable yet still requires retention, adaptation, or task-specific specialization.\\
%
- \textbf{Forgetting}: A meaningful continual scenario should induce measurable forgetting under Naive training. We assess this using the Forgetting Measure (see Appendix~\ref{appendix:sec:evaluation_protocol}). Low forgetting may indicate that the scenario is too weak, redundant, or insufficiently demanding for continual-learning evaluation.

{\color{brown}

}
} 

%
%

{\color{red}




 \section{Proposed Benchmark CAD Scenarios} {\color{blue}

\subsection{Benchmark scenarios}
We leverage the proposed framework to create five benchmark-ready CAD scenarios, which constitute one of the core artifacts of this work. The suite includes three single-dataset scenarios, \textbf{CAD-CICIDS2017}~\citep{sharafaldin2018cicids}, \textbf{CAD-CICIDS2018}~\citep{sharafaldin2018cicids}, and \textbf{CAD-CICUNSW}~\citep{mohammadian2024poisoning}, and two multi-dataset scenarios, \textbf{MCAD-CIC-3x1} and \textbf{MCAD-CIC-3xN}. 
The single-dataset scenarios capture within-dataset regime changes, while the two multi-dataset variants extend the evaluation to cross-dataset continual adaptation. In particular, MCAD-CIC-3x1 treats each dataset as one task, whereas MCAD-CIC-3xN combines the discovered task structure across datasets into a longer and more heterogeneous sequence. Both multi-dataset scenarios are not manually defined, but are processed through all stages of our framework, including task selection and ordering.
These scenarios are derived from three large-scale tabular cybersecurity datasets: CICIDS2017, CICIDS2018, and CIC-UNSW-NB15. CICIDS2017 and CICIDS2018 contain diverse intrusion-detection traffic and attack behaviors collected across multiple periods \citep{sharafaldin2018cicids}, while CIC-UNSW-NB15 provides a complementary network-security setting with distinct traffic characteristics and attack types \citep{moustafa2015unswnb15}. More details about the datasets can be found in Appendix~\ref{appendix:sec:dataset}. While the proposed framework is domain-general, the released benchmark suite focuses on large-scale cybersecurity anomaly detection, where validated continual scenarios are particularly lacking~\citep{amalapuram2024spider}.

The final scenarios differ substantially in scale, number of tasks, anomaly ratio, and task structure, as summarized in Table~\ref{tab:scenario_overview}. The single-dataset scenarios isolate continual anomaly detection within a coherent source domain, whereas the multi-dataset scenarios increase distributional heterogeneity by combining concepts across datasets. As a result, the released assets cover several sources of practical difficulty, including class imbalance, imbalance between task sizes, small retained tasks. More details about per-task size and anomaly ratio can be found in Appendix~\ref{appendix:sec:final-scenarios}. 


  \begin{table}[t]     
  \centering     
  \setlength{\tabcolsep}{21pt}
  \small
  \caption{Overview of the five released CAD scenarios, including number of tasks, total number of samples, and test anomaly ratio.}
  \label{tab:scenario_overview}     
  \begin{tabular}{l c c c}         
  \toprule         
  \textbf{Scenario} &  \textbf{\# Tasks} & \textbf{\# Data samples} & \textbf{Anomaly ratio (test)} \\         
  \midrule         
  CAD-CICIDS2017 & 6 & 2,076,848 &  18.77\% \\
  CAD-CICIDS2018 & 5 &  2,590,771 &  28.04\% \\         
  CAD-CICUNSW  & 5  & 1,084,928  & 12.76\% \\
  MCAD-CIC-3x1 & 3 & 17,915,569 & 10.42\% \\
  MCAD-CIC-3xN & 13 & 3,581,792 & 27.36\% \\
  \bottomrule    
  \end{tabular} 
  \end{table}

For each scenario, we provide six orderings over the same validated task set: curriculum-up, curriculum-down, generalization-up, generalization-down, smooth drift, and abrupt drift. These orderings are designed to expose different aspects of continual learning behavior while reducing the risk that future conclusions are driven by a single favorable task sequence. 

The released artifacts therefore include not only the filtered tasks, but also standardized train/test splits, task metadata, exact orderings, and scenario definitions ready for evaluation. The scenarios are available at \url{https://huggingface.co/collections/lifelonglab/tabular-cad-benchmarks}. The source code for our framework is available at \url{hhttps://github.com/lifelonglab/CAD-Benchmarks-Framework}.
The Appendix~\ref{app:dataset_progression} provides the full construction process for each scenario, including task candidate discovery process, filtering decisions and single-task expert validation, task-level sample statistics. 

 }

{\color{blue}

%

%



\subsection{Experimental protocol}
Our experimental setup is organized into two phases. In the first phase, which corresponds to framework processing up to the scenario selection, where we rely on ROC-AUC as the main metric. At this stage, we use a compact yet diverse set of clustering algorithms (k-Means, spectral clustering, and Gaussian mixtures) and anomaly detection  models (Autoencoder, Isolation Forest, and, AE1\_SVM), in order to evaluate candidate splits and clustering-based task  discovery procedures. The purpose of this phase is not to benchmark methods exhaustively, but to obtain sufficiently robust evidence for  task feasibility, heterogeneity, and scenario structure.
Appendix~\ref{appendix:sensitivity} provides an analysis of different hyperparameters in the filtering phase of the framework.
In the second phase, which corresponds to scenario validation, we expand both the evaluation metrics and the set of models. In addition to ROC-AUC, we report normalized PR-AUC in order to better account for class imbalance and provide a complementary view of detection  quality. We also consider a broader family of anomaly detection models, including Variational Autoencoder (VAE), Deep SVDD, DAGMM, and NeutralAD. 
This second phase is intended to  assess whether the retained scenarios are not only feasible, but also non-trivial and capable of exposing meaningful continual learning  dynamics across stronger and more diverse baselines.
More information about data preprocessing, clustering methods, models, evaluation protocol and metrics, and execution metrics can be found in Appendix~\ref{appendix:additional-exp-details},
Finally, the hyperparameters and reproducibility info can be found in Appendix~\ref{appendix:sec:reproducibility}.
}

For comparability, future evaluations of the framework should report results for all six orderings. We recommend reporting ROC-AUC, normalized PR-AUC, FM, and per-task performance matrices, as well as filtering statistics (see Appendix~\ref{app:dataset_progression} and Appendix~\ref{appendix:sec:final-scenarios}).


 \subsection{Experimental validation of the scenarios}
 We organize the discussion around the scenario-validation criteria introduced in Section~\ref{sec:scenario-validation}. In particular, we analyze each scenario in terms of feasibility, non-triviality, and forgetting. 
 It is noteworthy that in the validation phase we are using a different set of models (VAE, Deep SVDD, DAGMM, NeutralAD) than the scenario construction stage (Autoencoder, AE1SVM, IsolationForest). 
 }

}

\begin{table}[h]
\setlength{\tabcolsep}{1pt}
\centering
\small
\caption{ROC\text{-}AUC (mean $\pm$ std across orderings) on \textit{CAD-CICIDS2017}, \textit{CAD-CICIDS2018}, \textit{CAD-CICUNSW}, \textit{MCAD-CIC-3x1}, \textit{MCAD-CIC-3xN}. FM$\downarrow$: lower is better. Best per column in \textbf{bold}}.\label{tab:results_combined_roc_auc}
\begin{tabular}{llcccccccc}
\toprule
\textbf{Dataset} & \textbf{Model} & \multicolumn{2}{c}{\textbf{Naive}} & \multicolumn{2}{c}{\textbf{Replay}} & \multicolumn{2}{c}{\textbf{Cumulative}} & \multicolumn{2}{c}{\textbf{MSTE}} \\
\cmidrule(lr){3-4}\cmidrule(lr){5-6}\cmidrule(lr){7-8}\cmidrule(lr){9-10}
 &  & \textbf{ROC}$\uparrow$ & \textbf{FM}$\downarrow$ & \textbf{ROC}$\uparrow$ & \textbf{FM}$\downarrow$ & \textbf{ROC}$\uparrow$ & \textbf{FM}$\downarrow$ & \textbf{ROC}$\uparrow$ & \textbf{FM}$\downarrow$ \\
\midrule
\multirow{4}{*}{CAD-CICIDS2017} & VAE & 59.3{\scriptsize $\pm$3.4} & 31.0{\scriptsize $\pm$4.9} & 85.2{\scriptsize $\pm$2.5} & 7.4{\scriptsize $\pm$2.5} & 74.5{\scriptsize $\pm$7.8} & 6.4{\scriptsize $\pm$4.5} & \textbf{97.0{\scriptsize $\pm$1.3}} & \textbf{0.0{\scriptsize $\pm$0.0}} \\
 & Deep SVDD & 54.8{\scriptsize $\pm$4.1} & 30.1{\scriptsize $\pm$5.5} & 70.0{\scriptsize $\pm$7.1} & 8.3{\scriptsize $\pm$1.5} & 73.7{\scriptsize $\pm$7.7} & 8.9{\scriptsize $\pm$3.0} & \textbf{91.0{\scriptsize $\pm$2.7}} & \textbf{0.0{\scriptsize $\pm$0.0}} \\
 & DAGMM & 59.4{\scriptsize $\pm$2.2} & 25.1{\scriptsize $\pm$1.3} & 70.7{\scriptsize $\pm$4.6} & 10.6{\scriptsize $\pm$3.6} & 57.6{\scriptsize $\pm$9.2} & 11.6{\scriptsize $\pm$5.3} & \textbf{87.8{\scriptsize $\pm$6.7}} & \textbf{0.0{\scriptsize $\pm$0.0}} \\
 & NeutralAD & 59.8{\scriptsize $\pm$3.4} & 30.2{\scriptsize $\pm$2.6} & 88.3{\scriptsize $\pm$2.2} & 5.8{\scriptsize $\pm$2.4} & 79.0{\scriptsize $\pm$10.3} & 7.4{\scriptsize $\pm$5.2} & \textbf{96.5{\scriptsize $\pm$2.6}} & \textbf{0.0{\scriptsize $\pm$0.0}} \\
\midrule
\multirow{4}{*}{CAD-CICIDS2018} & VAE & 53.4{\scriptsize $\pm$7.8} & 25.1{\scriptsize $\pm$5.3} & 66.6{\scriptsize $\pm$5.1} & 16.7{\scriptsize $\pm$3.5} & 75.6{\scriptsize $\pm$3.9} & 5.4{\scriptsize $\pm$2.0} & \textbf{86.2{\scriptsize $\pm$6.5}} & \textbf{0.0{\scriptsize $\pm$0.0}} \\
 & Deep SVDD & 54.2{\scriptsize $\pm$3.7} & 24.1{\scriptsize $\pm$4.9} & 67.2{\scriptsize $\pm$5.1} & 14.2{\scriptsize $\pm$3.2} & 62.0{\scriptsize $\pm$5.9} & 10.7{\scriptsize $\pm$3.6} & \textbf{83.9{\scriptsize $\pm$1.6}} & \textbf{0.0{\scriptsize $\pm$0.0}} \\
 & DAGMM & 56.8{\scriptsize $\pm$2.7} & 14.8{\scriptsize $\pm$3.4} & \textbf{67.2{\scriptsize $\pm$5.5}} & 8.8{\scriptsize $\pm$4.1} & 60.4{\scriptsize $\pm$3.8} & 8.7{\scriptsize $\pm$3.4} & 65.1{\scriptsize $\pm$7.5} & \textbf{0.0{\scriptsize $\pm$0.0}} \\
 & NeutralAD & 53.6{\scriptsize $\pm$6.2} & 25.9{\scriptsize $\pm$6.3} & 76.2{\scriptsize $\pm$5.8} & 7.4{\scriptsize $\pm$3.5} & 74.3{\scriptsize $\pm$2.9} & 6.9{\scriptsize $\pm$1.7} & \textbf{83.1{\scriptsize $\pm$3.3}} & \textbf{0.0{\scriptsize $\pm$0.0}} \\
\midrule
\multirow{4}{*}{CAD-CICUNSW} & VAE & 54.5{\scriptsize $\pm$11.5} & 34.4{\scriptsize $\pm$9.5} & 83.0{\scriptsize $\pm$3.3} & 9.8{\scriptsize $\pm$2.3} & 84.4{\scriptsize $\pm$3.4} & 3.2{\scriptsize $\pm$1.3} & \textbf{96.3{\scriptsize $\pm$1.4}} & \textbf{0.0{\scriptsize $\pm$0.0}} \\
 & Deep SVDD & 51.1{\scriptsize $\pm$5.1} & 22.5{\scriptsize $\pm$5.3} & 50.5{\scriptsize $\pm$7.1} & 15.3{\scriptsize $\pm$1.2} & 49.2{\scriptsize $\pm$4.4} & 11.3{\scriptsize $\pm$0.9} & \textbf{74.2{\scriptsize $\pm$4.4}} & \textbf{0.0{\scriptsize $\pm$0.0}} \\
 & DAGMM & 55.0{\scriptsize $\pm$4.7} & 26.3{\scriptsize $\pm$7.0} & 60.5{\scriptsize $\pm$6.8} & 14.5{\scriptsize $\pm$5.3} & 61.6{\scriptsize $\pm$7.3} & 11.0{\scriptsize $\pm$6.6} & \textbf{88.4{\scriptsize $\pm$6.2}} & \textbf{0.0{\scriptsize $\pm$0.0}} \\
 & NeutralAD & 56.6{\scriptsize $\pm$3.4} & 29.7{\scriptsize $\pm$3.3} & 82.1{\scriptsize $\pm$4.3} & 9.1{\scriptsize $\pm$2.5} & 68.9{\scriptsize $\pm$9.1} & 10.9{\scriptsize $\pm$3.9} & \textbf{92.8{\scriptsize $\pm$2.2}} & \textbf{0.0{\scriptsize $\pm$0.0}} \\
\midrule
\multirow{4}{*}{MCAD-CIC-3x1} & VAE & 72.2{\scriptsize $\pm$3.3} & 9.2{\scriptsize $\pm$2.4} & 75.4{\scriptsize $\pm$1.8} & 6.1{\scriptsize $\pm$1.5} & 79.4{\scriptsize $\pm$1.7} & 2.0{\scriptsize $\pm$0.7} & \textbf{83.7{\scriptsize $\pm$0.8}} & \textbf{0.0{\scriptsize $\pm$0.0}} \\
 & Deep SVDD & 63.3{\scriptsize $\pm$3.5} & 2.6{\scriptsize $\pm$2.6} & 63.5{\scriptsize $\pm$2.0} & 3.4{\scriptsize $\pm$3.0} & 60.3{\scriptsize $\pm$3.5} & 2.2{\scriptsize $\pm$1.8} & \textbf{65.8{\scriptsize $\pm$2.6}} & \textbf{0.0{\scriptsize $\pm$0.0}} \\
 & DAGMM & \textbf{62.0{\scriptsize $\pm$3.3}} & 4.9{\scriptsize $\pm$1.5} & 61.9{\scriptsize $\pm$5.1} & 2.6{\scriptsize $\pm$3.8} & 60.6{\scriptsize $\pm$2.1} & 3.4{\scriptsize $\pm$3.1} & 59.1{\scriptsize $\pm$7.7} & \textbf{0.0{\scriptsize $\pm$0.0}} \\
 & NeutralAD & 68.4{\scriptsize $\pm$2.6} & 14.0{\scriptsize $\pm$1.6} & 77.1{\scriptsize $\pm$5.8} & 3.9{\scriptsize $\pm$3.2} & 75.4{\scriptsize $\pm$4.9} & 2.9{\scriptsize $\pm$2.2} & \textbf{79.1{\scriptsize $\pm$10.5}} & \textbf{0.0{\scriptsize $\pm$0.0}} \\
\midrule
\multirow{4}{*}{MCAD-CIC-3xN} & VAE & 42.9{\scriptsize $\pm$6.9} & 46.7{\scriptsize $\pm$8.5} & 82.6{\scriptsize $\pm$2.8} & 9.5{\scriptsize $\pm$2.4} & 75.9{\scriptsize $\pm$4.7} & 9.1{\scriptsize $\pm$1.6} & \textbf{95.6{\scriptsize $\pm$2.4}} & \textbf{0.0{\scriptsize $\pm$0.0}} \\
 & Deep SVDD & 46.5{\scriptsize $\pm$4.2} & 34.4{\scriptsize $\pm$6.3} & 67.7{\scriptsize $\pm$5.6} & 14.3{\scriptsize $\pm$1.6} & 67.0{\scriptsize $\pm$7.2} & 11.4{\scriptsize $\pm$2.7} & \textbf{83.6{\scriptsize $\pm$1.7}} & \textbf{0.0{\scriptsize $\pm$0.0}} \\
 & DAGMM & 51.1{\scriptsize $\pm$2.3} & 31.7{\scriptsize $\pm$3.8} & 68.2{\scriptsize $\pm$6.3} & 15.8{\scriptsize $\pm$2.2} & 63.6{\scriptsize $\pm$4.1} & 17.6{\scriptsize $\pm$3.8} & \textbf{82.2{\scriptsize $\pm$4.6}} & \textbf{0.0{\scriptsize $\pm$0.0}} \\
 & NeutralAD & 49.5{\scriptsize $\pm$3.1} & 39.9{\scriptsize $\pm$4.3} & 88.3{\scriptsize $\pm$2.1} & 6.2{\scriptsize $\pm$0.6} & 82.1{\scriptsize $\pm$5.6} & 7.7{\scriptsize $\pm$1.6} & \textbf{94.3{\scriptsize $\pm$2.5}} & \textbf{0.0{\scriptsize $\pm$0.0}} \\
\bottomrule
\end{tabular}
\end{table}

{\color{blue}
Table~\ref{tab:results_combined_roc_auc} supports our intended scenario properties along the three validation axes described in Section~\ref{sec:scenario-validation}, while also showing that the five released scenarios cover different difficulty regimes. 

\textbf{Feasibility:} At least one strong reference strategy reaches a meaningful ROC-AUC ($\geq 0.8$) in every scenario, and MSTE usually provides the best performance. In single-dataset scenarios, MSTE reaches high ROC-AUC for most models. This indicates that retained tasks are not intrinsically degenerate and can be learned when task-specific specialization or stronger retention is available. The lowest best ROC-AUC occurs in MCAD-CIC-3x1 (83.7 with VAE), but this value still presents a significant margin over the random classifier (0.5). 
%
At the same time, the table illustrates that the datasets are intrinsically diverse. 
In MCAD-CIC-3x1, Deep SVDD and DAGMM reach only 65.8 and 59.1 ROC-AUC with MSTE, which
 means that the scenario is not only challenging from a continual-learning perspective, but also difficult from a single-task anomaly detection perspective. 
Conversely, MCAD-CIC-3xN combines severe continual difficulty with strong task-specific learnability: VAE and NeutralAD achieve 95.6 and 94.3 ROC-AUC with MSTE, while Naive falls to 42.9 and 49.5. These contrasting cases are useful because they prevent our scenarios from representing  only one type of difficulty. 
Interestingly, Cumulative achieves subpar performance, which may seem counter-intuitive given the full data availability. Our in-depth analysis revealed that this phenomenon is due to the highly imbalanced tasks in the scenario (see Appendix~\ref{appendix:sec:cumulative-imbalance} for a more detailed discussion).
%


\textbf{Non-triviality:} Across most scenario--model combinations, Naive remains substantially below MSTE and often below Replay or Cumulative. 
This is the desired behavior for CAD validation: the scenarios are learnable, but simple sequential fine-tuning does not already solve them. Replay often improves over Naive, further indicating that the scenarios expose retention and adaptation challenges rather than only detector weakness. 
DAGMM is an exception in two cases, where MSTE performs worse than the other strategies, including Naive on MCAD-CIC-3x1. The low ROC-AUC values indicate that this detector struggles with the underlying tasks already in the single-task setting, rather than only because of continual-learning effects.

\textbf{Forgetting:} All the scenarios exhibit clear forgetting, as Naive incurs large FM values on the single-dataset scenarios and particularly on MCAD-CIC-3xN, where FM reaches 46.7 for VAE and 39.9 for NeutralAD. FM for MSTE is zero or near-zero because MSTE evaluates task-specific experts rather than a single sequentially updated model, while Cumulative and Replay usually reduce forgetting substantially. This separation is important: if Naive matched the stronger references while also showing negligible FM, the scenarios would likely be too weak or redundant. Instead, the combination of low Naive ROC-AUC and high Naive FM confirms that our benchmark scenarios contain genuine continual-learning pressure.

Figure~\ref{fig:validation_roc_auc_smaller_main} provides a qualitative confirmation of the forgetting behavior already summarized in Table~\ref{tab:results_combined_roc_auc}. In particular, the Naive trajectories show a clear degradation on previously observed tasks as training progresses, whereas stronger reference strategies preserve substantially more performance across the scenario. This visual pattern is consistent with the high FM values reported for Naive and supports the interpretation that the scenario induces genuine forgetting rather than uniformly weak task learnability.

Overall, the validation results show that the final scenarios are not interchangeable variants of the same setting. Some are highly learnable but harshly continual, some are intrinsically harder even for single-task experts, and the multi-dataset scenarios introduce additional cross-domain heterogeneity. This diversity is precisely the intended outcome of the framework: a set of principled CAD scenarios that can support future benchmarking without relying on a single narrow notion of difficulty.
%
%
We report more results in terms of both ROC-AUC and Normalized PR-AUC in Appendix~\ref{appendix:scenario-results} and Appendix~\ref{appendix:sec:npr_auc}.

\begin{figure}
    \centering
    \includegraphics[width=\linewidth]{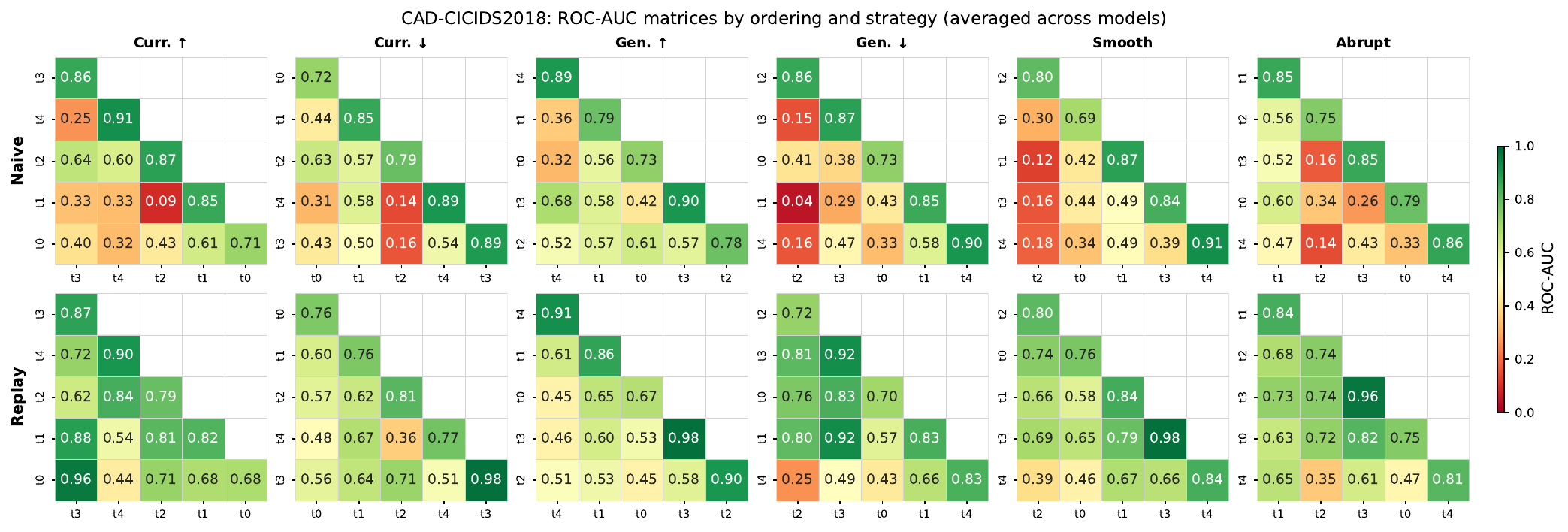}
    \caption{Validation results on CAD-CICIDS2018 across the six orderings. ROC-AUC is reported for the considered continual learning strategies, showing that Naive sequential training degrades substantially more than Replay.}
    \label{fig:validation_roc_auc_smaller_main}
\end{figure}




\FloatBarrier

\section{Conclusions}
{\color{orange}

This paper introduced a framework for constructing continual anomaly detection (CAD) scenarios from tabular anomaly detection datasets not originally designed for continual learning. Instead of treating task sequences as an experimental detail, we make scenario construction explicit, reproducible, and empirically justified. The resulting benchmark suite contains five scenarios, each with six principled ordering families. Validation shows that the retained scenarios are feasible under strong references, non-trivial for naive sequential training, and capable of inducing measurable forgetting.

\textit{Limitations:} The benchmark suite is built on top of only cybersecurity datasets, so its conclusions should not be assumed to transfer directly to other tabular domains. Second, the constructed scenarios depend on design choices in task discovery, filtering, ordering, and validation. Third, the framework is computationally demanding because it requires extensive single-task expert training and cross-task evaluation. These limitations motivate future work on broader domains, additional sensitivity analyses, while positioning the present work as a methodological foundation rather than a definitive, domain-complete CAD benchmark.

\textit{Broader Impact.} More principled CAD scenario construction can improve the reliability of anomaly-detection research by reducing the risk of overstated claims caused by weak or misleading benchmarks. At the same time, deployed anomaly detectors can still cause harms through false alarms, unnecessary investigations, privacy risks, or unfair treatment of benign but unusual behavior. The scenarios introduced here should therefore be treated as evaluation assets, not as evidence of deployment readiness without domain-specific validation and governance.
}

{\color{brown}

}
\begin{ack}
The research presented in this paper was supported by funds from: i) the Polish Ministry of Science and Higher Education allocated to the AGH University of Krakow; ii) 
SOCCER project (grant agreement no. 101128073), with the support of the European Cybersecurity Competence Centre (ECCC), and co-funded by the Polish Ministry of Science and Higher Education under the programme "Co-funded International Projects."

\end{ack}

{
\small



\bibliographystyle{plainnat}
\bibliography{references}
}

\appendix
\FloatBarrier

\section{Notation}
\label{app:notation}

Table~\ref{app:tab:notation} provides a notation reference for our framework.

\begin{table}[h]
\centering
\small
\caption{Notation used throughout the framework.}
\label{app:tab:notation}
\begin{tabular}{@{}c|l|p{0.85\linewidth}@{}}
\toprule
 & \textbf{Symbol} & \textbf{Meaning} \\
\midrule

\multirow{17}{*}{\rotatebox[origin=c]{90}{\emph{Task discovery}}}
 & $\mathcal{D}$ & Input tabular dataset. \\
 & $N$ & Number of observations in $\mathcal{D}$. \\
 & $\mathcal{X}$ & Observation (feature) space. \\
 & $x_i$ & The $i$-th observation (sample) in $\mathcal{D}$, $i \in \{1,\ldots,N\}$. \\
 & $y_i$ & Label of $x_i$; $0$ for normal, $1$ for anomalous. \\
 & $X, Y$ & Random variables for an observation and its label, with realizations $x_i, y_i$. \\
 & $K$ & Number of candidate tasks discovered from $\mathcal{D}$. \\
 & $\mathcal{T}$ & Set of candidate tasks. \\
 & $\tau_k$ & The $k$-th candidate task. \\
 & $\mathcal{D}_k^{\mathrm{norm}}$ & Normal samples assigned to task $k$. \\
 & $\mathcal{D}_k^{\mathrm{anom}}$ & Anomalous samples assigned to task $k$. \\
 & $\mathcal{Z}$ & Set of natural boundaries in $\mathcal{D}$ (e.g., days, users). \\
 & $g$ & Boundary mapping $\mathcal{X}\!\to\!\mathcal{Z}$ used for natural-boundary task discovery. \\
 & $h$ & Mapping $\mathcal{X}\!\to\!\{1,\ldots,K\}$ used for clustering-based task discovery. \\
 & $t$ & A boundary , $t \in \mathcal{Z}$. \\
 & $P_t(X,Y)$ & Joint distribution over observations and labels under boundary $t$. \\
 & $P_t(X\mid Y\!=\!0)$ & Distribution of normal observations under boundary $t$. \\

\midrule
\multirow{7}{*}{\rotatebox[origin=c]{90}{\emph{Task evaluation}}}
 & $F$ & Pool of single-task expert (STE) models used in task evaluation. \\
 & $f$ & A single model, $f \in F$. \\
 & $|F|$ & Number of models in $F$. \\
 & $M^f$ & Cross-task STE performance matrix for model $f$, of size $K \times K$. \\
 & $M^f_{i,j}$ & Performance of $f$ trained on task $i$ and evaluated on task $j$. \\
 & $b_k$ & Random-baseline performance on task $k$ (e.g., $0.5$ for ROC-AUC). \\
 & $S_k^{\star}$ & Best self-performance on task $k$ across models, $\max_{f\in F} M^f_{k,k}$. \\

\midrule
\multirow{8}{*}{\rotatebox[origin=c]{90}{\emph{Task selection}}}
 & $\gamma_l$ & Minimum gain over $b_k$ required for self-learnability. \\
 & $\gamma_t$ & Retained-performance fraction of $S_k^{\star}$ used to count covering tasks. \\
 & $P_c$ & Maximum allowed number of other tasks that cover task $k$ (FC2). \\
 & $\mathcal{C}_k^f$ & Set of tasks whose STE under $f$ covers task $k$ above $\gamma_t S_k^{\star}$. \\
 & $\gamma_d$ & Relative dominance threshold over $S_i^{\star}$. \\
 & $P_d$ & Maximum allowed number of tasks dominated by task $k$. \\
 & $\mathcal{O}_k^f$ & Set of tasks dominated by task $k$ under model $f$. \\
 & $\delta_r$ & Maximum allowed average difference between two tasks profiles. \\

\midrule
\multirow{15}{*}{\rotatebox[origin=c]{90}{\emph{Task ordering}}}
 & $\mathcal{S}$ & Set of scenario candidates produced by task selection. \\
 & $s_i$ & A scenario candidate, $s_i \in \mathcal{S}$. \\
 & $m$ & Number of scenario candidates in $\mathcal{S}$. \\
 & $\Pi$ & Family of ordering strategies considered by the framework. \\
 & $\Pi^f$ & Orderings produced for model $f$ on a given scenario. \\
 & $\pi$ & A single ordering of the $K$ tasks in a scenario. \\
 & $\pi^f_{SD}$ & Smooth-drift ordering for model $f$. \\
 & $\pi^f_{AD}$ & Abrupt-drift ordering for model $f$. \\
 & $\pi^f_{C}$ & Curriculum (easy-to-hard) ordering for model $f$. \\
 & $\pi^f_{RC}$ & Reverse-curriculum (hard-to-easy) ordering for model $f$. \\
 & $\pi^f_{GI}$ & Increasing-generalization ordering for model $f$. \\
 & $\pi^f_{GD}$ & Decreasing-generalization ordering for model $f$. \\
 & $D_{i,j}$ & Cross-task dissimilarity used by drift orderings. \\
 & $d_i$ & Difficulty of task $i$ used by curriculum orderings, $d_i = 1 - M^f_{i,i}$. \\
 & $g_i$ & Generalization score of task $i$ used by generalization orderings. \\

\midrule
\multirow{9}{*}{\rotatebox[origin=c]{90}{\emph{\makecell{Final scenario \\ and ordering selection}}}}
 & $s^{\star}$ & Selected scenario after cross-model concordance, $s^{\star} \in \mathcal{S}$. \\
 & $\Pi^{f,s}$ & Orderings produced for model $f$ on scenario $s$. \\
 & $l$ & Index over the six ordering families ($SD, AD, C, RC, GI, GD$). \\
 & $W_{\pi}$ & Kendall's coefficient of concordance for ordering family $\pi$ across the models in $F$, $W_{\pi} \in [0,1]$. \\
 & $\Pi^{\star}$ & Set of final consensus orderings, one per family. \\
 & $\pi^{\star}_l$ & Final consensus ordering for family $l$, obtained by Borda count on $s^{\star}$. \\
 & $r_{l,f}(t)$ & Rank of task $t$ in the ordering $\pi_l$ produced by model $f$. \\
 & $B_l(t)$ & Borda score of task $t$ in family $l$, $B_l(t) = \sum_{f \in F} r_{l,f}(t)$. \\
 & $P$ & Task-by-task performance matrix; $P_{t,k}$ is the performance on task $k$ after training up to task $t$. \\

\bottomrule
\end{tabular}
\end{table}

\FloatBarrier

\section{Additional descriptions}
\label{appendix:additional-exp-details}

\subsection{Detailed framework overview}
Figure~\ref{fig:methodology-overview-detailed} provides the full version of the framework overview shown compactly in the main paper.

\begin{figure*}[t]
\centering
\resizebox{0.98\textwidth}{!}{%
\begin{tikzpicture}[
    font=\sffamily\footnotesize,
    >=Latex,
    stage/.style={
        draw=#1!75!black,
        fill=#1!5,
        rounded corners=8pt,
        very thick,
        minimum width=5.45cm,
        minimum height=4.75cm,
        text width=4.95cm,
        align=center,
        inner sep=6pt
    },
    stagewide/.style={
        draw=#1!75!black,
        fill=#1!5,
        rounded corners=8pt,
        very thick,
        minimum width=15cm,
        minimum height=2.20cm,
        text width=13cm,
        align=center,
        inner sep=6pt
    },
    card/.style={
        draw=#1!60!black,
        fill=white,
        rounded corners=4pt,
        line width=0.55pt,
        text width=4.80cm,
        align=left,
        inner sep=4pt,
        font=\sffamily\scriptsize
    },
    cardc/.style={
        draw=#1!60!black,
        fill=white,
        rounded corners=4pt,
        line width=0.55pt,
        text width=4.80cm,
        align=center,
        inner sep=4pt,
        font=\sffamily\scriptsize
    },
    mini/.style={
        draw=#1!60!black,
        fill=white,
        rounded corners=4pt,
        line width=0.55pt,
        text width=1.40cm,
        minimum height=0.75cm,
        align=center,
        inner sep=4pt,
        font=\sffamily\scriptsize
    },
    num/.style={
        circle,
        fill=#1!85!black,
        text=white,
        font=\bfseries\sffamily,
        minimum size=0.52cm,
        inner sep=0pt
    },
    badge/.style={
        circle,
        draw=#1!70!black,
        fill=#1!10,
        text=#1!70!black,
        font=\bfseries\sffamily\scriptsize,
        minimum size=0.55cm,
        inner sep=0pt
    },
    title/.style={
        font=\bfseries\sffamily\normalsize,
        text=black!85,
        align=center
    },
    subtitle/.style={
        font=\sffamily\scriptsize,
        text=black!62,
        align=center
    },
    connector/.style={
        ->,
        very thick,
        draw=black!65,
        rounded corners=4pt
    }
]

\definecolor{greenC}{RGB}{45,145,85}
\definecolor{purpleC}{RGB}{111,84,170}
\definecolor{orangeC}{RGB}{219,139,24}
\definecolor{blueC}{RGB}{42,102,180}
\definecolor{tealC}{RGB}{38,145,155}
\definecolor{redC}{RGB}{185,64,82}


\node[stage=greenC] (disc) at (0,0) {};
\node[num=greenC] at ($(disc.north west)+(0.35,-0.35)$) {1};

\node[title, text=greenC!60!black] at ($(disc.north)+(0,-0.38)$)
{Task Discovery};

\node[subtitle] at ($(disc.north)+(0,-0.78)$)
{Generate task candidates};

\node[badge=greenC] at ($(disc.north)+(3.1,-2)$) {$\mathcal{T}$};

\node[card=greenC] at ($(disc.center)+(0,0.9)$)
{\textbf{Natural splits}\\
Time, data source, additional metadata};

\node[card=greenC] at ($(disc.center)+(0,0)$)
{\textbf{Clustering-based discovery}\\
3 methods on top of clustering algorithms};

\node[card=greenC] at ($(disc.center)+(0,-0.9)$)
{\textbf{Multiple datasets}\\
Tasks on top of multiple datasets };

\node[cardc=greenC] at ($(disc.south)+(0,0.50)$)
{\textbf{Output:} multiple candidate task sets $\mathcal{T}=\{\tau_1,\ldots,\tau_K\}$};

\node[stage=purpleC] (eval) at (6.2,0) {};
\node[num=purpleC] at ($(eval.north west)+(0.35,-0.35)$) {2};

\node[title, text=purpleC!65!black] at ($(eval.north)+(0,-0.38)$)
{Task Evaluation};

\node[subtitle] at ($(eval.north)+(0,-0.78)$)
{Learnability and heterogeneity};

\node[badge=purpleC] at ($(eval.north)+(3.1,-2)$) {$M^f$};

\node[card=purpleC] at ($(eval.center)+(0,0.9)$)
{Evaluate $\mathcal{T}$ with multiple AD models creating cross-task performance matrix };

\node[card=purpleC] at ($(eval.center)+(0,-0.3)$)
{
\textbf{Cross-task performance matrix}\\[2pt]
$M^f_{i,j}$ train model $f$ on task $i$ test on task $j$\\[4pt]
\scriptsize Learnability, transfer, overlap, diversity
};

\node[cardc=purpleC] at ($(eval.south)+(0,0.50)$)
{\textbf{Output:} $M^f$ for each candidate tasks set and model $f$};

\node[stage=orangeC] (select) at (12.4,0) {};
\node[num=orangeC] at ($(select.north west)+(0.35,-0.35)$) {3};

\node[title, text=orangeC!70!black] at ($(select.north)+(0,-0.38)$)
{Task Selection};

\node[subtitle] at ($(select.north)+(0,-0.78)$)
{Filter unsuitable tasks};

\node[badge=orangeC] at ($(select.north)+(0.4,-5.12)$) {$S$};

\node[card=orangeC] at ($(select.center)+(0,0.9)$)
{\textbf{FC1. Self-learnability}\\
Reject tasks that no model can learn};

\node[card=orangeC] at ($(select.center)+(0,0)$)
{\textbf{FC2, FC3. Transfer coverage}\\
Limit excessive transfer coverage};

\node[card=orangeC] at ($(select.center)+(0,-0.9)$)
{\textbf{FC4, FC5. Redundancy}\\
Avoid redundant tasks};

\node[cardc=orangeC] at ($(select.south)+(0,0.50)$)
{\textbf{Output:} A set of scenario candidates \\ $S = \{s_1, s_2, \ldots, \}$};


\node[stage=blueC] (construct) at (0,-5.5) {};
\node[num=blueC] at ($(construct.north west)+(0.35,-0.35)$) {4};

\node[title, text=blueC!70!black] at ($(construct.north)+(0,-0.38)$)
{Scenario Orderings};

\node[subtitle] at ($(construct.north)+(0,-0.78)$)
{Define principled orderings};

\node[badge=blueC] at ($(construct.north)+(3.1,-1.9)$) {$\Pi$};

\node[card=blueC] at ($(construct.center)+(0,0.9)$)
{\textbf{Drift-based}\\
Smooth drift; abrupt drift};

\node[card=blueC] at ($(construct.center)+(0,0)$)
{\textbf{Curriculum-based}\\
Easy to hard; Hard to easy};

\node[card=blueC] at ($(construct.center)+(0,-0.9)$)
{\textbf{Generalization}\\
Low-to-high; high-to-low transfer};

\node[cardc=blueC] at ($(construct.south)+(0,0.50)$)
{\textbf{Output:} A set of orderings \\ $\Pi = \{ \pi_{SD},\pi_{AD},\pi_C,\pi_{RC},\pi_{GI},\pi_{GD} \}$};

\node[stage=tealC] (rank) at (6.2,-5.5) {};
\node[num=tealC] at ($(rank.north west)+(0.35,-0.35)$) {5};

\node[title, text=tealC!70!black] at ($(rank.north)+(0,-0.38)$)
{Scenario Selection};

\node[subtitle] at ($(rank.north)+(0,-0.78)$)
{Choose the final scenario and orderings};

\node[badge=tealC] at ($(rank.north)+(3.1,-1.9)$) {$S^{*}$};

\node[card=tealC] at ($(rank.center)+(0,0.9)$)
{Input: A set of scenario candidates $S$ and a set of orderings $\Pi$ per model $f$ and scenario candidate $s \in S$};

\node[card=tealC] at ($(rank.center)+(0,-0.05)$)
{\textbf{Scenario selection}\\
Cross-model concordance via Kendall's $W$};

\node[card=tealC] at ($(rank.center)+(0,-0.9)$)
{\textbf{Ordering consensus across models}\\
Borda count over per-model rankings};

\node[cardc=tealC] at ($(rank.south)+(0,0.50)$)
{\textbf{Output:} Selected scenario $s^{*}$ along with final set of orderings $\Pi$};
\node[stage=redC] (valid) at (12.5,-5.5) {};
\node[num=redC] at ($(valid.north west)+(0.38,-0.35)$) {6};

\node[title, text=redC!70!black] at ($(valid.north)+(0,-0.38)$)
{Scenario Validation};

\node[subtitle] at ($(valid.north)+(0,-0.78)$)
{Check meaningful continual-learning dynamics};


\node[card=redC] at ($(valid.center)+(0,0.9)$)
{
Validation across multiple AD models and learning strategies
};

\node[mini=redC] at ($(valid.center)+(-1.82,-0.32)$)
{\textbf{Feasibility}};

\node[mini=redC] at ($(valid.center)+(0,-0.32)$)
{\textbf{Non-triviality}};

\node[mini=redC] at ($(valid.center)+(1.82,-0.32)$)
{\textbf{Forgetting}};

\node[cardc=redC, text width=4.90cm] at ($(valid.south)+(0,0.50)$)
{\textbf{Output:} Validated scenarios and experimental results};








\node[stagewide=greenC] (out) at (6.2,-9.2) {};
\node[num=greenC] at ($(out.north west)+(0.38,-0.35)$) {7};

\node[title, text=greenC!60!black] at ($(out.north)+(-5.5,-0.38)$)
{Output};

\node[subtitle] at ($(out.north)+(-1.8,-0.38)$)
{Accepted CAD scenario suite ready for benchmarking};


\node[card=greenC, text width=9cm] at ($(out.center)+(-2.5,-0.35)$)
{\textbf{Scenarios}\\
Single-dataset: i) CAD-CICIDS2017; ii) CAD-CICIDS2018; iii) CAD-CICUNSW \\
Multi-dataset: i) MCAD-CIC-3x1; ii) MCAD-CIC-3xN};

\node[card=greenC, text width=4.35cm] at ($(out.center)+(4.8,-0.35)$)
{\textbf{Each scenario}\\
Final set of tasks $T^{*}$ \\
A set of 6 orderings $\Pi$ \\
Statistics and validation results};

\draw[connector] (disc.east) -- (eval.west);
\draw[connector] (eval.east) -- (select.west);

\draw[connector] (select.south) -- ++(0,-0.3) -| ($(construct.north)+(0,0.2)$) -- (construct.north);
\draw[connector] (construct.east) -- (rank.west);
\draw[connector] (rank.east) -- (valid.west);

\draw[connector] ($(valid.south)+(2,0)$) -| ++(0,-1.3) -- (out.east);

\end{tikzpicture}%
}
\caption{Detailed overview of the proposed framework for transforming anomaly detection datasets into validated continual anomaly detection scenarios.}
\label{fig:methodology-overview-detailed}
\end{figure*}
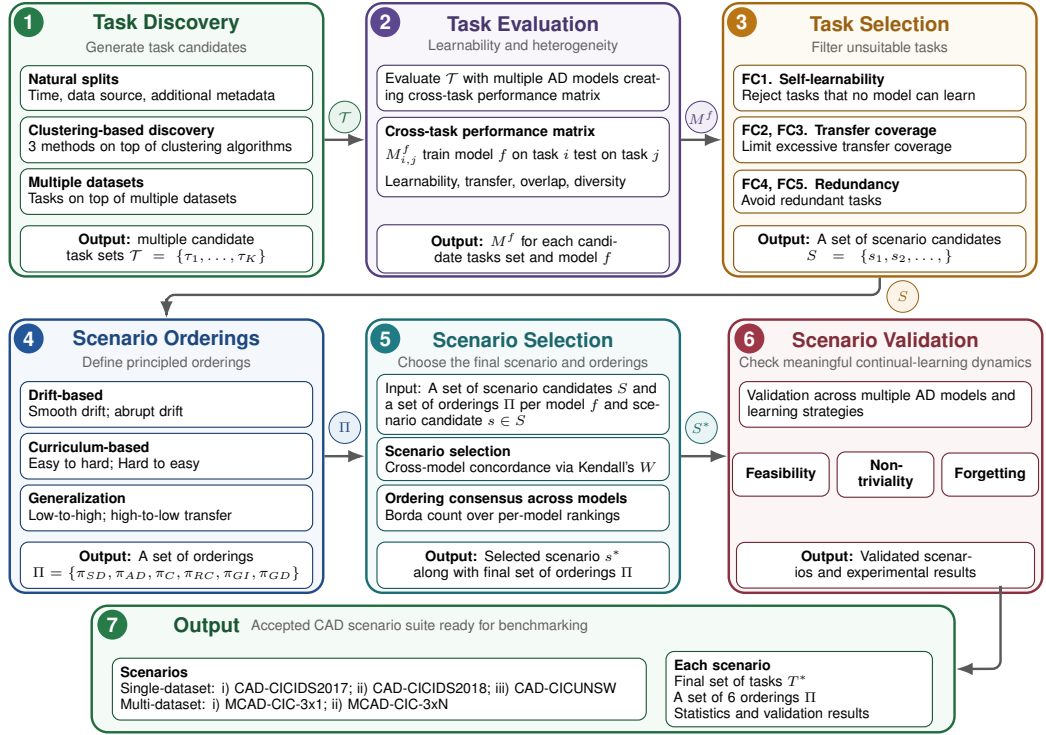

\subsection{Task discovery details}
\label{app:task-discovery-details}
{\color{red}
Task discovery aims to identify candidate tasks from the full dataset. Each candidate task should correspond to a coherent regime of normal behavior together with the anomalies used to evaluate detection performance within that regime.

More formally, let the fully preprocessed dataset be denoted by
$\mathcal{D} = \{(x_i, y_i)\}_{i=1}^N,$
where \(x_i \in \mathcal{X}\) is an observation and \(y_i \in \{0,1\}\) indicates whether the observation is normal (\(y_i=0\)) or anomalous (\(y_i=1\)). The goal of task discovery is to partition or group \(\mathcal{D}\) into a collection of candidate tasks
$
\mathcal{T} = \{\tau_1, \tau_2, \dots, \tau_K\},
$
where each task \(\tau_k\) corresponds to a coherent regime of normality together with the anomalies used to evaluate detection under that regime. More precisely, each task can be written as
$
\tau_k = \bigl(\mathcal{D}_k^{\mathrm{norm}}, \mathcal{D}_k^{\mathrm{anom}}\bigr),
$
with \(\mathcal{D}_k^{\mathrm{norm}} \subseteq \{(x_i,y_i)\in\mathcal{D}: y_i=0\}\) and \(\mathcal{D}_k^{\mathrm{anom}} \subseteq \{(x_i,y_i)\in\mathcal{D}: y_i=1\}\).

The framework supports multiple task discovery mechanisms because the relevant source of task structure is inherently dataset-dependent. In some cases, tasks may be induced by natural boundaries, such as days, users, or operating conditions. If such metadata are available, one may define a mapping
$
g:\mathcal{X}\to\mathcal{Z},
$
where \(\mathcal{Z}\) is a set of boundary labels, and construct tasks from the level sets of \(g\). In other datasets, however, no such explicit structure exists, and tasks must instead be inferred from the data through a grouping rule
$
h:\mathcal{X}\to\{1,\dots,K\},
$
for example via clustering, change-point detection, or other similarity-based procedures.

Importantly, the mere existence of natural boundaries does not imply that they define meaningful continual learning tasks. A boundary-induced partition is only justified if it corresponds to a substantive change in the underlying data-generating distribution. Formally, if \(P_t(X,Y)\) denotes the distribution associated with boundary \(t\), then distinct boundaries \(t\neq t'\) are not automatically informative when
$
P_t(X,Y) \approx P_{t'}(X,Y),
$
or, more specifically in the anomaly detection setting, when
$
P_t(X \mid Y=0) \approx P_{t'}(X \mid Y=0).
$
For example, data collected on different days may still represent essentially the same normal regime.

Accordingly, task discovery seeks to identify a sequence of tasks
$
\tau_1, \tau_2, \dots, \tau_K
$
such that each task is meaningful from a continual learning standpoint, regardless of whether it coincides with the original dataset boundaries. In particular, the objective is to construct tasks that induce sufficiently distinct and non-trivial learning conditions. A newly introduced task \(\tau_k\) should therefore not be reducible to a trivial recombination of previous tasks \(\{\tau_1,\dots,\tau_{k-1}\}\), but should require genuine adaptation because its associated normal regime and evaluation anomalies differ in a material way from those previously encountered.

We note that, in some circumstances, task discovery may be optionally skipped in our framework. For instance, a practitioner may want to extract a multi-dataset scenario with one concept per dataset. In that context, task discovery would not be appropriate since it represents an unnecessary filtering stage.
}
\subsection{Preprocessing}
\label{appendix:sec:preprocessing}
{\color{red}
For tabular data, this includes removing unusable columns, handling missing values, encoding categorical variables, scaling numerical attributes when required by the downstream model, and preserving any metadata that may define a natural task boundary.
Before attempting to create a continual scenario, we also perform a dataset-level feasibility check. The purpose is to avoid constructing elaborate scenarios from a dataset that cannot be solved by the candidate anomaly detectors even as a single task. If no reasonable detector can separate normal and anomalous examples under a non-continual protocol, then failures in the continual setting would not be informative about forgetting or adaptation.
}

\subsection{Clustering algorithms for task discovery}
\label{appendix:sec:clustering}
We use three clustering-based strategies to derive candidate tasks:
\begin{itemize}
    \item \textbf{Both Classes (BC):} the clustering algorithm $c$ is applied to normal and anomalous samples jointly, so that candidate tasks reflect the structure of both classes.
    \item \textbf{Random Anomalies (RA):} the clustering algorithm $c$ is applied only to normal samples, after which anomalous samples are distributed randomly across the resulting tasks.
    \item \textbf{Closest Anomalies (CA):} the clustering algorithm $c$ is applied only to normal samples, after which each anomalous sample is assigned to the task with the nearest normal-cluster centroid.
\end{itemize}

{\color{red}
In our experiments, we leverage three clustering algorithms $c$:
\begin{itemize}
    \item \textbf{Gaussian Mixture Models}, which provide a soft probabilistic partition of the feature space and can capture ellipsoidal clusters with different covariance structures.
    \item \textbf{$k$-means}, which provides a simple centroid-based partition and acts as a scalable baseline for inducing compact normal-behavior regimes.
    \item \textbf{Spectral Clustering}, which uses graph structure to recover non-convex groupings that may not be well represented by centroid- or Gaussian-based assumptions.
\end{itemize}
Combining these algorithms with BC, RA, and CA yields a diverse pool of candidate task decompositions. The subsequent STE-based filtering stage then determines which decompositions produce learnable, non-redundant, and non-dominating tasks.

}

\subsection{Leveraged models}
\label{appendix:sec:models}
{\color{red}
We use different anomaly detectors during scenario construction and final scenario validation. During \textbf{task evaluation}, the goal is not to benchmark anomaly detection methods, but to obtain a robust signal about task learnability and cross-task transfer. We therefore use a compact and heterogeneous pool: an Autoencoder, AE1SVM from PyOD, and Isolation Forest. This combines reconstruction-based, representation-plus-boundary, and tree-ensemble perspectives while keeping the construction phase computationally manageable.

During \textbf{scenario validation}, we evaluate the retained scenarios with a broader set of neural anomaly detectors: DAGMM, Deep SVDD, VAE, and NeutralAD. These models are intentionally distinct from the construction-stage pool, which reduces the risk that the final validation merely confirms biases of the models used to select tasks. Hyperparameters are reported in Appendix~\ref{appendix:sec:reproducibility}.
}

\subsection{Datasets}
\label{appendix:sec:dataset}
{\color{red}
The released scenarios are derived from three public tabular cybersecurity datasets. \textbf{CICIDS2017} and \textbf{CSE-CIC-IDS2018} contain network-flow records collected by the Canadian Institute for Cybersecurity and include benign traffic together with multiple attack categories \citep{sharafaldin2018cicids}. They provide large-scale intrusion-detection settings with heterogeneous attack behavior and substantial sample-size imbalance across candidate regimes. \textbf{CIC-UNSW-NB15}, used in our CAD-CICUNSW scenario, provides an additional network-intrusion dataset with different traffic-generation conditions and attack families \citep{moustafa2015unswnb15,mohammadian2024poisoning}.

These datasets were selected because they are large enough to support multiple retained tasks after filtering, contain meaningful anomaly labels, and differ in size, anomaly prevalence, and distributional structure. The framework uses them as source assets, but the contribution of this paper is the resulting set of validated CAD scenarios, task splits, and orderings derived from them.
}

\subsection{Evaluation protocol}
\label{appendix:sec:evaluation_protocol}

{\color{red}
We evaluate each final scenario by training a model sequentially over the ordered tasks and measuring performance on all tasks after each training step. This produces a task-by-time performance matrix $P$, where $P_{t,k}$ denotes the performance on task $k$ after training has progressed through task $t$. Aggregate performance for metric $\L$ is computed across the tasks in the scenario 
\begin{equation}
L_{\mathrm{AGG}}
=
\frac{2}{K(K+1)}
\sum_{t=1}^{K}
\sum_{j=1}^{t}
P_{t,j}.
    \label{eq:lroc}
\end{equation}
Where $K$ corresponds to the number of tasks in the scenario.

We use ROC-AUC as the primary ranking-based anomaly detection metric. Because all scenarios are imbalanced, we also report normalized PR-AUC, which provides a complementary view of detection quality under skewed anomaly prevalence. Normalized PR-AUC rescales PR-AUC relative to the task-specific anomaly prior, making results more comparable across tasks with different class ratios.
\[
\mathrm{nPR\text{-}AUC}
=
\frac{\mathrm{PR\text{-}AUC}-\pi}{1-\pi},
\]
where \(\pi\) is the anomaly prevalence in the corresponding test set. A value of 0 corresponds to the expected PR-AUC of a random ranking under the task-specific anomaly prior.

For continual-learning behavior, we report the Forgetting Measure (FM). For each task, forgetting is computed as the gap between the best performance achieved on that task after it was learned and the final performance on the same task after subsequent training. FM is then averaged across tasks. Higher FM indicates stronger loss of previously acquired performance, while lower FM indicates better retention. First, we assess $\mathrm{FM'}$ that corresponds to forgetting measured after learning  a single new task $k$:
\begin{equation}
\mathrm{FM'_k}
=
\frac{1}{k-1}
\sum_{j=1}^{k-1}
\left(
\max_{t \in \{j,\ldots,k\}} P_{t,j} - P_{k,j}
\right)
\label{eq:fm}
\end{equation}

And then we compute the final measure $\mathrm{FM}$ as an average in induced forgetting in the scenario:
\begin{equation}
    \mathrm{FM} = \frac{1}{K} \sum_{k=2}^K \mathrm{FM'_k}
\end{equation}

}

\subsection{Continual learning strategies}
\label{appendix:sec:cl_strategy}
{\color{red}
We evaluate four diagnostic training strategies. \textbf{Naive} trains sequentially on the current task without any explicit retention mechanism and therefore serves as the standard lower reference for forgetting. \textbf{Replay} augments sequential training with a fixed-size buffer of samples from previous tasks, providing a practical retention-based continual-learning strategy. \textbf{Cumulative} retrains or updates using all data observed so far, which is not a realistic continual setting but is useful as a strong reference with full past-data access. \textbf{MSTE} uses a pool of multiple single-task experts, one per task, and therefore separates task-specific learnability from forgetting. Together, these strategies allow us to distinguish scenarios that are unlearnable, scenarios that are trivially solved by naive adaptation, and scenarios that create genuine retention pressure.
}

\FloatBarrier
\section{Filtering thresholds sensitivity analysis}
\label{appendix:sensitivity}
{\color{brown}
Figures~\ref{fig:app:sensitivity_cicids2017}--\ref{fig:app:sensitivity_cicunsw} report a one-at-a-time sensitivity analysis of the filtering thresholds used in the task-selection stage. For each candidate split, we vary one threshold while keeping the others fixed at their default values, and we measure how many tasks are removed by the corresponding filtering rule. The red curve shows the mean number of removed tasks across candidate splits, while the faint blue curves show split-specific behavior. The dashed green line marks the default threshold adopted in the main paper.

Across all three datasets, the chosen defaults consistently lie in moderate operating regions rather than at pathological extremes. In particular, the self-learnability threshold $\gamma_l=0.75$ is placed before the sharp rise in removed tasks that appears near the most restrictive values, which avoids discarding large portions of the scenario due to an overly strict learnability requirement. Similarly, the transfer thresholds $\gamma_t=\gamma_d=0.9$ retain enough stringency to eliminate trivial or dominating tasks, but are not so severe as to collapse the candidate split space. The count thresholds $P_t=P_d=2$ also fall near the transition between permissive and highly selective behavior, which is consistent with the intended goal of tolerating limited transfer while rejecting tasks that are broadly redundant or dominant.

The redundancy thresholds $\delta_r=0.1$ for both row-wise and column-wise profiles show the expected monotonic behavior: larger values merge more tasks as increasingly dissimilar profiles are treated as redundant. Here again, the selected default remains in a conservative regime where redundancy filtering is active but not overwhelming. Overall, these plots support the threshold choices used in the main paper: they enforce meaningful filtering across datasets, yet avoid the unstable regions in which small parameter changes would lead to disproportionately large changes in the number of retained tasks.
}

    \begin{figure}
        \centering
        \includegraphics[width=\linewidth]{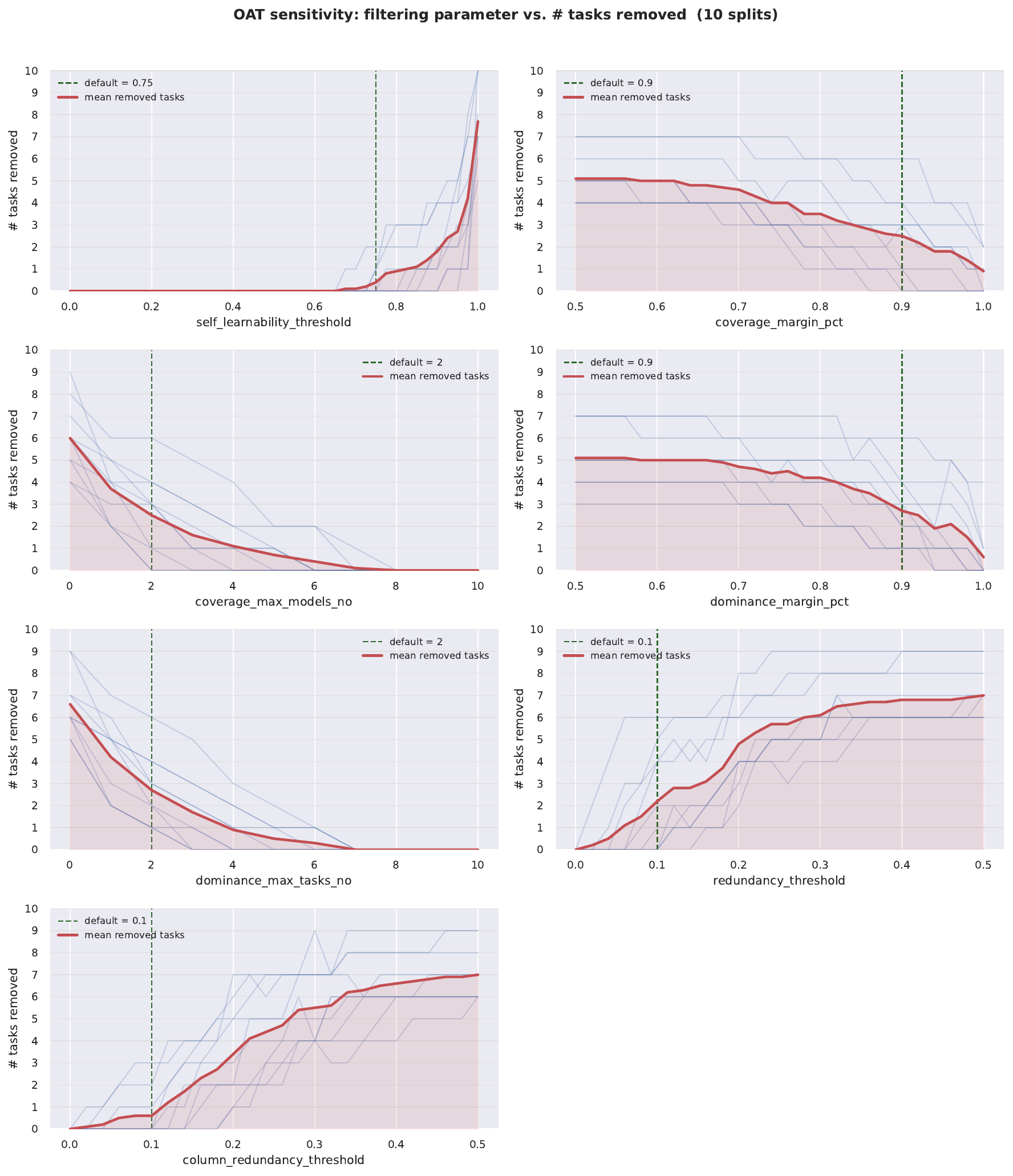}
        \caption{One-at-a-time sensitivity analysis of the filtering thresholds on CICIDS2017. The red curve reports the mean number of removed tasks across candidate splits, the blue curves show split-specific trajectories, and the dashed green line marks the default threshold used in the main paper.}
        \label{fig:app:sensitivity_cicids2017}
    \end{figure}
    
    \begin{figure}
        \centering
        \includegraphics[width=\linewidth]{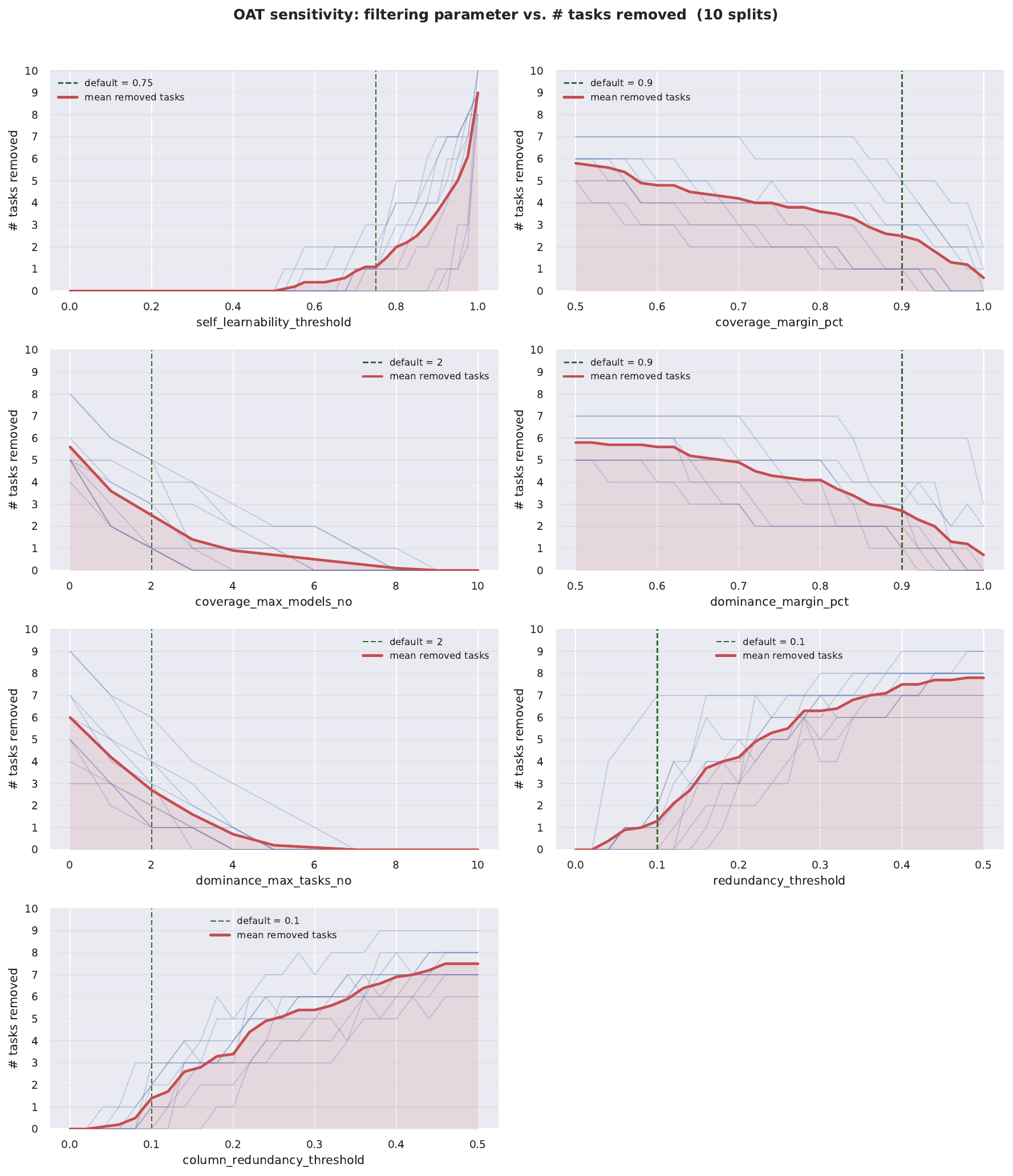}
        \caption{One-at-a-time sensitivity analysis of the filtering thresholds on CICIDS2018. The red curve reports the mean number of removed tasks across candidate splits, the blue curves show split-specific trajectories, and the dashed green line marks the default threshold used in the main paper.}
        \label{fig:app:sensitivity_cicids2018}
    \end{figure}
    
    \begin{figure}
        \centering
        \includegraphics[width=\linewidth]{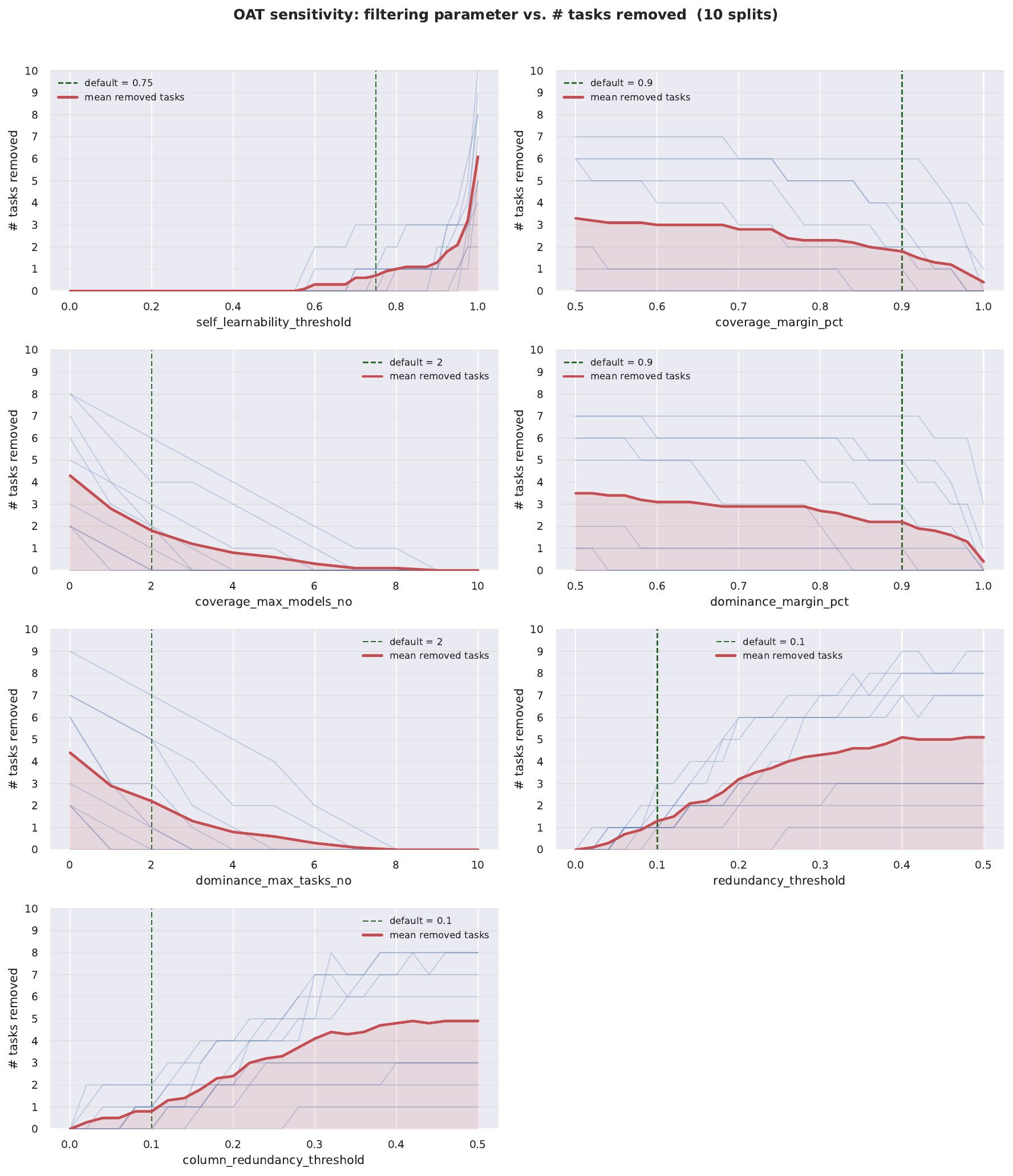}
        \caption{One-at-a-time sensitivity analysis of the filtering thresholds on CICUNSW. The red curve reports the mean number of removed tasks across candidate splits, the blue curves show split-specific trajectories, and the dashed green line marks the default threshold used in the main paper.}
        \label{fig:app:sensitivity_cicunsw}
    \end{figure}

\FloatBarrier

\subsection{Execution times}
\label{appendix:sec:times}

\begin{table}[h]
\centering
\small
\begin{tabular}{lllrrrr}
\toprule
\textbf{Dataset} & \textbf{Model} & \textbf{Time} & \textbf{Naive} & \textbf{Cumulative} & \textbf{MSTE} & \textbf{Replay} \\
\midrule
\multirow{8}{*}{CAD-CICIDS2017} & \multirow{2}{*}{VAE} & Train & 26m\,42s & 101m\,46s & 26m\,47s & 22m\,11s \\
 &  & Eval & 1m\,41s & 1m\,42s & 1m\,2s & 1m\,41s \\
 & \multirow{2}{*}{DAGMM} & Train & 20m\,46s & 72m\,58s & 20m\,36s & 20m\,5s \\
 &  & Eval & 23.0s & 23.1s & 14.5s & 22.9s \\
 & \multirow{2}{*}{DeepSVDD} & Train & 9m\,15s & 32m\,55s & 9m\,19s & 5m\,23s \\
 &  & Eval & 4.8s & 4.8s & 3.2s & 4.8s \\
 & \multirow{2}{*}{NeutralAD} & Train & 21m\,5s & 52m\,13s & 16m\,59s & 19m\,20s \\
 &  & Eval & 58.8s & 58.9s & 35.7s & 59.3s \\
\midrule
\multirow{8}{*}{CAD-CICIDS2018} & \multirow{2}{*}{VAE} & Train & 28m\,45s & 90m\,8s & 28m\,50s & 28m\,37s \\
 &  & Eval & 1m\,44s & 1m\,41s & 1m\,7s & 1m\,57s \\
 & \multirow{2}{*}{DAGMM} & Train & 23m\,14s & 82m\,10s & 23m\,37s & 24m\,9s \\
 &  & Eval & 24.7s & 25.9s & 17.2s & 25.9s \\
 & \multirow{2}{*}{DeepSVDD} & Train & 8m\,10s & 28m\,8s & 8m\,15s & 6m\,47s \\
 &  & Eval & 5.2s & 5.2s & 3.7s & 5.5s \\
 & \multirow{2}{*}{NeutralAD} & Train & 44m\,15s & 62m\,46s & 43m\,57s & 45m\,33s \\
 &  & Eval & 1m\,9s & 1m\,8s & 43.6s & 1m\,9s \\
\midrule
\multirow{8}{*}{CAD-CICUNSW} & \multirow{2}{*}{VAE} & Train & 12m\,45s & 41m\,11s & 12m\,43s & 12m\,15s \\
 &  & Eval & 39.1s & 38.7s & 24.2s & 42.9s \\
 & \multirow{2}{*}{DAGMM} & Train & 10m\,55s & 34m\,50s & 10m\,57s & 11m\,16s \\
 &  & Eval & 10.4s & 9.8s & 6.4s & 10.6s \\
 & \multirow{2}{*}{DeepSVDD} & Train & 5m\,51s & 19m\,1s & 5m\,54s & 2m\,39s \\
 &  & Eval & 2.0s & 2.0s & 1.4s & 1.9s \\
 & \multirow{2}{*}{NeutralAD} & Train & 23m\,16s & 54m\,45s & 23m\,11s & 25m\,11s \\
 &  & Eval & 24.9s & 25.0s & 15.8s & 25.0s \\
\midrule
\multirow{8}{*}{MCAD-CIC-1xN} & \multirow{2}{*}{VAE} & Train & 182m\,5s & 376m\,29s & 183m\,23s & 163m\,17s \\
 &  & Eval & 9m\,46s & 9m\,45s & 6m\,57s & 9m\,46s \\
 & \multirow{2}{*}{DAGMM} & Train & 188m\,14s & 339m\,20s & 176m\,30s & 163m\,13s \\
 &  & Eval & 2m\,40s & 2m\,25s & 1m\,51s & 2m\,38s \\
 & \multirow{2}{*}{DeepSVDD} & Train & 50m\,39s & 130m\,60s & 50m\,54s & 53m\,45s \\
 &  & Eval & 29.3s & 29.7s & 21.6s & 30.0s \\
 & \multirow{2}{*}{NeutralAD} & Train & 44m\,40s & 45m\,5s & 43m\,44s & 45m\,33s \\
 &  & Eval & 6m\,24s & 6m\,33s & 4m\,24s & 6m\,20s \\
\midrule
\multirow{8}{*}{MCAD-CIC-3xN} & \multirow{2}{*}{VAE} & Train & 48m\,25s & 280m\,46s & 49m\,48s & 37m\,40s \\
 &  & Eval & 7m\,21s & 6m\,10s & 4m\,8s & 6m\,13s \\
 & \multirow{2}{*}{DAGMM} & Train & 43m\,16s & 224m\,41s & 40m\,30s & 29m\,53s \\
 &  & Eval & 1m\,53s & 1m\,29s & 59.0s & 1m\,30s \\
 & \multirow{2}{*}{DeepSVDD} & Train & 17m\,9s & 82m\,43s & 13m\,11s & 8m\,27s \\
 &  & Eval & 18.9s & 18.6s & 11.1s & 18.5s \\
 & \multirow{2}{*}{NeutralAD} & Train & 70m\,1s & 160m\,5s & 70m\,25s & 78m\,44s \\
 &  & Eval & 4m\,17s & 4m\,10s & 2m\,17s & 4m\,21s \\
\bottomrule
\end{tabular}
\caption{Execution times (train and eval time separately, mean across orderings) on \textit{CAD-CICIDS2017}, \textit{CAD-CICIDS2018}, \textit{CAD-CICUNSW}, \textit{MCAD-CIC-3x1}, \textit{MCAD-CIC-3xN}. }
\label{tab:execution_times}
\end{table}

It is noteworthy that just our validation phase of the scenarios includes 480 full experimental phases (5 scenarios x 6 orderings x 4 AD models x 4 CL strategies), which corresponds to over 3000 trainings of individual tasks. 
The overall number of experiments is significantly higher, due to the initial phase involving task selection.
    
\FloatBarrier

\section{Reproducibility}
\label{appendix:sec:reproducibility}

{\color{brown}
\subsection{Compute resources}
All experiments were executed on a high-performance computing (HPC) system whose full peak capacity is approximately 36 PFLOPS, with AMD EPYC 9654 and NVIDIA Grace CPUs and about 300~TB of RAM. The experiments reported in this paper used only a subset of these resources. In particular, the GPU-based runs were executed on (in peak) 15 NVIDIA GH200 96GB GPUs, while the full system provides up to 440 nodes. This is reported to clarify both the scale of the underlying platform and the smaller portion of it effectively used for the results presented here.
}

\subsection{Datasets licensing }
{\color{brown}
The datasets used in this work are credited to their original creators and were used under the access conditions stated by their source providers. CICIDS2017 \citep{sharafaldin2018cicids} is made publicly available by the Canadian Institute for Cybersecurity (University of New Brunswick) for research use.
CSE-CIC-IDS2018 \citep{sharafaldin2018cicids} is released by the same source with permission to redistribute, republish, and mirror the dataset.  
For the UNSW-NB15 dataset \citep{moustafa2015unswnb15} (used here through our CICUNSW scenario), UNSW Canberra states that free use for academic research is granted in perpetuity, while commercial use requires agreement from the authors. 
}

\FloatBarrier

\subsection{Hyperparameters}
 \begin{table}[h]                                                                                                                                                                                                                                                                                                
  \centering                                                                                                                                                                                                                                                                                                                                                                                                                                                                                 
  \small                                                                                                                                                                                                                                                                                                            
  \caption{Hyperparameters of the anomaly detection models.}                                                                                                                                                                                                                                                        
  \label{tab:hyperparameters}                                                                                                                                                                                                                                                                                       
  \resizebox{\linewidth}{!}{%
  \begin{tabular}{lccccccc}                                                                                                                                                                                                                                                                                       
  \toprule
  \textbf{Hyperparameter}                                                                                                                                                                                                                                                                                           
    & \textbf{AE}
    & \textbf{VAE}                                                                                                                                                                                                                                                                                                  
    & \textbf{Deep SVDD}                                                                                                                                                                                                                                                                                          
    & \textbf{DAGMM}                                                                                                                                                                                                                                                                                                
    & \textbf{NeuTraLAD}
    & \textbf{AE1SVM}                                                                                                                                                                                                                                                                                               
    & \textbf{IForest} \\                                                                                                                                                                                                                                                                                         
  \midrule
  \multicolumn{8}{l}{\textit{Training}} \\
  \midrule                                                                                                                                                                                                                                                                                                          
  Epochs              & 20  & 20    & 20    & 20    & 20    & 20    & --- \\
  Batch size          & 128           & 128   & 128   & 128   & 128   & 128   & --- \\                                                                                                                                                                                                                              
  Optimizer           & Adam          & Adam  & Adam  & Adam  & Adam  & Adam  & --- \\                                                                                                                                                                                                                              
  Learning rate       & 1e-3          & 1e-3  & 1e-3  & 1e-3  & 1e-3  & 1e-3  & --- \\                                                                                                                                                                                                                              
  Weight decay        & ---           & 1e-5  & 0.1   & ---   & ---   & 1e-5  & --- \\                                                                                                                                                                                                                              
  \midrule                                                                                                                                                                                                                                                                                                          
  \multicolumn{8}{l}{\textit{Architecture}} \\                                                                                                                                                                                                                                                                      
  \midrule                                                                                                                                                                                                                                                                                                          
  Encoder hidden       & {[}64, 16{]}       & {[}128, 64, 32{]}  & {[}64, 32{]}  & auto  & {[}100, 50{]}  & {[}64, 32{]}  & --- \\                                                                                                                                                                     
  Decoder hidden       & {[}16, 64{]}       & {[}32, 64, 128{]}  & ---           & auto  & ---            & {[}32, 64{]}  & --- \\                                                                                                                                                                       
  Latent dim           & 16                 & 2                  & ---           & $5+\lfloor d/20\rfloor$  & 128  & 32  & --- \\                                                                                                                                                                                   
  Activation           & ReLU               & ReLU               & ReLU          & Tanh             & LeakyReLU      & ReLU          & --- \\                                                                                                                                                                       
  Batch norm           & ---                & ---                & ---           & ---              & ---            & \checkmark    & --- \\                                                                                                                                                                       
  Dropout              & ---                & 0.2                & 0.2           & ---              & ---            & 0.2           & --- \\                                                                                                                                                                       
  \midrule                                                                                                                                                                                                                                                                                                          
  \multicolumn{8}{l}{\textit{Model-specific}} \\                                                                                                                                                                                                                                                                    
  \midrule                                                                                                                                                                                                                                                                                                          
  $\beta$ (KL weight)                 & ---  & 1.0   & ---   & ---   & ---       & ---    & --- \\                                                                                                                                                                                                                
  GMM components $K$                  & ---  & ---   & ---   & 3     & ---       & ---    & --- \\                                                                                                                                                                                                                  
  $\lambda_\text{energy}$             & ---  & ---   & ---   & 0.1   & ---       & ---    & --- \\                                                                                                                                                                                                                  
  $\lambda_\text{cov}$                & ---  & ---   & ---   & 0.005 & ---       & ---    & --- \\                                                                                                                                                                                                                  
  Transformations $K$                 & ---  & ---   & ---   & ---   & 11        & ---    & --- \\                                                                                                                                                                                                                  
  Transform type                      & ---  & ---   & ---   & ---   & residual  & ---    & --- \\                                                                                                                                                                                                                  
  Temperature $\tau$                  & ---  & ---   & ---   & ---   & 0.1       & ---    & --- \\                                                                                                                                                                                                                  
  RFF features                        & ---  & ---   & ---   & ---   & ---       & 512    & --- \\                                                                                                                                                                                                                  
  $\sigma$ (RBF kernel)               & ---  & ---   & ---   & ---   & ---       & 1.0    & --- \\                                                                                                                                                                                                                  
  $\nu$ (one-class SVM)               & ---  & ---   & ---   & ---   & ---       & 0.1    & --- \\                                                                                                                                                                                                                  
  Estimators                          & ---  & ---   & ---   & ---   & ---       & ---    & 100 \\                                                                                                                                                                                                                  
  Max samples                         & ---  & ---   & ---   & ---   & ---       & ---    & auto \\                                                                                                                                                                                                                 
  \bottomrule                                                                                                                                                                                                                                                                                                       
  \end{tabular}}                                                                                                           
  \end{table}

\begin{table}[h]
\centering
\small
\caption{Clustering algorithm hyperparameters. $K$ is the number of clusters, set per dataset.}
\label{tab:clustering_hyperparams}
\begin{tabular}{lll}
\toprule
\textbf{Algorithm} & \textbf{Hyperparameter} & \textbf{Value} \\
\midrule
\multirow{1}{*}{KMeans}
 & \texttt{n\_clusters} & 10 \\
\midrule
\multirow{3}{*}{Gaussian Mixture}
 & \texttt{n\_components} & 10 \\
 & \texttt{covariance\_type} & \texttt{diag} \\
 & \texttt{reg\_covar} & $10^{-3}$ \\
\midrule
\multirow{2}{*}{Spectral Clustering}
 & \texttt{n\_clusters} & 10 \\
 & \texttt{affinity} & \texttt{nearest\_neighbors} \\
\bottomrule
\end{tabular}
\end{table}

 \begin{table}[h]                                                                                                                                                                                                                                                                                                  
  \centering                                                                                                                                                                                              \caption{Hyperparameters of the STE filtering criteria. Each criterion iteratively removes                                                                                                                                                                                                                        
  concepts violating its condition; parameters define the decision boundaries.}                                                                                                                                                                                                                                     
  \label{tab:filtering_params}                                                                                                            
  \small                                                                                                                                                                                                                                                                                                            
  \begin{tabular}{llcl}                                                                                                                                                                                                                                                                                             
  \toprule                                                                                                                                                                                                                                                                                                          
  \textbf{Criterion} & \textbf{Parameter} & \textbf{Value} \\
  \midrule                                                                                                                                                                                                                                                                                                          
  FC1. Self-learnability
    & {$\gamma_l$} & 0.75                                                                                                                                                                                                                                                    \\                                                                                                                                                                                                                                       
  \midrule                                                                                                                                                                                                                                                                                                          
  \multirow{2}{*}{FC2. Limited incoming transfer coverage}                                                                                                                                                                                                                                                                                         
    & $\gamma_t$      & 0.90                                                                                                                                                                                                                                   \\
    & {$P_t$}  & 2                                                                                                                                                                                                                                                          \\
\midrule                                                                                                                                                                                                    \multirow{2}{*}{FC3. Limited outgoing transfer coverage}                                                                                                                                                                                                                                                                                         
    & $\gamma_d$      & 0.90                                                                                                                                                                                                                                   \\
    & {$P_d$}  & 2                                                                                                                                                                                                                                                          \\
  \midrule                                                                                                                                                                                                                                                                                          
  \multirow{1}{*}{FC4. Source profile redundancy}
    & $\delta_r$       & 0.10                                                                                                                                                     \\
      \multirow{1}{*}{FC5. Target profile redundancy}
    & $ \delta_r$       & 0.10                                                                                                                                                     \\  
                                                                                                                                                                                                              
  \bottomrule                                                                                                                                                                                                                                                                                                       
  \end{tabular}

  \end{table}

Additional hyperparameters across the framework execution:
\begin{itemize}
    \item Minimal number of samples in clusters created during task discovery: i) normal = 1500; ii) anomalous = 1000
    \item Minimal number of tasks candidates in the split: i) CAD-CICIDS2017 = 5 tasks; ii) CAD-CICIDS2018 = 5 tasks; iii) CAD-CICUNSW = 3 tasks; iv) MCAD-CIC-3x1 = 3 tasks; v) MCAD-CIC-3xN = 5 tasks.
    \item  Replay uses a fixed global buffer of 25,000 samples. At each training stage, the buffer budget is distributed uniformly over previously observed tasks. Samples are selected uniformly at random from each observed task.
\end{itemize}

\FloatBarrier

\section{Scenario selection details}
\label{app:scenario-selection-details}
{\color{red}

\subsection{Scenario selection}
For scenario selection, we start from a set of scenario candidates $\mathcal{S} = \{s_1, \ldots, s_m\}$ and the orderings induced by each model $f$ on each scenario candidate $s_i$: $\Pi^{f, s_i} = \{ \pi_{SD}, \pi_{AD}, \pi_{C}, \pi_{RC}, \pi_{GI}, \pi_{GD} \}$. The rationale is that if multiple models with distinct inductive biases independently produce similar concept orderings, this agreement suggests the ordering reflects genuine structural properties of the concept space rather than model-specific behaviors.

We quantify cross-model concordance using Kendall's $W$ \citep{kendall1938new}, a coefficient of concordance for $|F|$ rankers (models $f \in F$). For each task $t$, $W$ compares the sum of ranks assigned by all models $f$ against what would be expected under independent random rankings. It is normalized to $[0, 1]$, where $W = 1$ indicates perfect agreement and $\mathbb{E}[W] = \frac{1}{|F|}$ under random rankings. Crucially, the normalization ensures $W$ is comparable across scenarios with different numbers of tasks.

Kendall's $W$ only measures whether models agree with each other, not what direction they agree in. If all models consistently rank task $t$ early in ascending order, they also consistently rank it late in descending order. Due to this, we use only four ordering families for split selection: abrupt drift, smooth drift, curriculum ascending, and generalization ascending, omitting curriculum descending and generalization descending.

Then, we select the scenario that maximizes the mean concordance:
\begin{equation}
    s^* = \underset{s \,\in\, \mathcal{S}}{\arg\max}\;
          \frac{W_{\pi_{SD}} + W_{\pi_{AD}} + W_{\pi_{c}} + W_{\pi_{G}}}{4}.
    \label{eq:split_selection}
\end{equation}

\subsection{Final ordering for selected scenario}
For the final consensus ordering, given the selected scenario $s^*$, we aggregate per-model orderings $\Pi^{f, s*}$ into a single consensus $\Pi^{*}$ using Borda count \citep{rothe2019borda}, by summing the rank assigned to each task across all models and sorting concepts by their total rank. Borda count naturally complements Kendall's $W$: while $W$ identifies splits with high cross-model agreement, Borda extracts the ordering that best reflects this agreement. By integrating rankings across models, it mitigates the influence of any single model and produces a stable consensus under moderate variability. Concepts consistently ranked early receive low aggregate scores and appear first, yielding an ordering that captures shared structure while preserving the interpretability of each ordering family.

The final ordering $\pi_l^* \  \forall \pi_l \ \in \Pi^*$ (where $\Pi^*$ consist of all orderings defined in the ,,Scenario orderings'' step) is achieved by sorting concepts by ascending Borda score:
\begin{equation}
B_{\ell}(t)=\sum_{f\in F} r_{\ell,f}(t),
\qquad
\pi^{\star}_{\ell}=\operatorname{argsort}_{t\in T_{s^\star}} B_{\ell}(t).
\end{equation}
where \(r_{\ell,f}(t)\) denotes the rank assigned to task \(t\) by model \(f\) for ordering family \(\ell\).

\subsection{Kendall W}
\label{app:kendall}

Let \(m = |F|\) be the number of construction models and let \(K\) be the number of tasks in scenario candidate \(s\). For ordering family \(\ell\), let \(r_{\ell,f}(t)\) be the rank assigned to task \(t\) by model \(f\). The rank sum is
\[
R_{\ell}(t)=\sum_{f \in F} r_{\ell,f}(t),
\qquad
\bar{R}=\frac{m(K+1)}{2}.
\]
Kendall's coefficient of concordance is
\[
W_{\ell}(s)
=
\frac{12\sum_{t\in T_s}(R_{\ell}(t)-\bar{R})^2}
{m^2(K^3-K)}.
\]

Under independent random rankings, the expected value is approximately \(1/m\), where \(m\) is the number of rankers. Values substantially exceeding this baseline indicate concordance beyond chance. The denominator \(m^2(K^3-K)\) ensures that \(W_\ell(s)\in[0,1]\), making the coefficient comparable across scenario candidates with different numbers of tasks.
}

\section{Detailed Specifications of Final Benchmark Scenarios}
\label{appendix:sec:final-scenarios}

This appendix provides detailed information about the final benchmark scenarios produced by our framework. For each scenario, we report the resulting sequence of tasks together with per-task statistics, including the number of normal and anomalous samples. We also provide the final task orderings used in the experiments. These details complement the main paper by making the constructed scenarios fully transparent and reproducible, and by allowing readers to inspect the scale, composition, and progression of each benchmark beyond the aggregate results reported in the main text.

\subsection{CAD-CICIDS2017}

\begin{table}[h]
\centering
\small
\caption{Task orderings for CAD-CICIDS2017}
\label{tab:orderingsCAD-CICIDS2017}
\begin{tabular}{ll}
\toprule
\textbf{Ordering} & \textbf{Task sequence} \\
\midrule
Curriculum (asc.) & t5 $\to$ t2 $\to$ t0 $\to$ t3 $\to$ t4 $\to$ t1 \\
Curriculum (desc.) & t1 $\to$ t4 $\to$ t3 $\to$ t0 $\to$ t2 $\to$ t5 \\
Generalization (desc.) & t4 $\to$ t3 $\to$ t0 $\to$ t2 $\to$ t5 $\to$ t1 \\
Generalization (asc.) & t1 $\to$ t5 $\to$ t2 $\to$ t0 $\to$ t3 $\to$ t4 \\
Smooth drift & t5 $\to$ t1 $\to$ t4 $\to$ t0 $\to$ t2 $\to$ t3 \\
Abrupt drift & t4 $\to$ t5 $\to$ t3 $\to$ t1 $\to$ t2 $\to$ t0 \\
\bottomrule
\end{tabular}
\end{table}
\begin{table}[ht]
\centering
\caption{Per-task statistics for CAD-CICIDS2017}
\label{tab:ds_stats_CAD-CICIDS2017}
\begin{tabular}{llrrr}
\toprule
Task ID & Task Name & Train Samples & Test Samples & Test Anomalies \\
\midrule
0 & cicids2017\_0 & 77,385 & 38,694 & 19,347 \\
1 & cicids2017\_1 & 1,434,082 & 411,738 & 53,217 \\
2 & cicids2017\_2 & 19,006 & 9,504 & 4,752 \\
3 & cicids2017\_3 & 8,392 & 4,196 & 2,098 \\
4 & cicids2017\_4 & 47,520 & 23,760 & 11,880 \\
5 & cicids2017\_5 & 1,713 & 858 & 429 \\
\bottomrule
\end{tabular}
\end{table}

{\color{brown}
Tables~\ref{tab:orderingsCAD-CICIDS2017} and \ref{tab:ds_stats_CAD-CICIDS2017} specify the retained CAD-CICIDS2017 scenario. The task statistics show a strongly imbalanced task-size distribution, with one dominant concept and several much smaller retained concepts. This imbalance is useful for stress-testing cumulative and replay-based strategies, because good scenario-level performance requires preserving smaller tasks rather than only fitting the dominant regime. The six orderings expose the same task set under curriculum, generalization, and drift assumptions, allowing the scenario to be reused without tying conclusions to one arbitrary sequence.
}

\FloatBarrier



\FloatBarrier
\subsection{CAD-CICIDS2018}
\begin{table}[h]
\centering
\small
\caption{Task orderings for CAD-CICIDS2018}
\label{tab:orderingsCAD-CICIDS2018}
\begin{tabular}{ll}
\toprule
\textbf{Ordering} & \textbf{Task sequence} \\
\midrule
Curriculum (asc.) & t3 $\to$ t4 $\to$ t2 $\to$ t1 $\to$ t0 \\
Curriculum (desc.) & t0 $\to$ t1 $\to$ t2 $\to$ t4 $\to$ t3 \\
Generalization (desc.) & t2 $\to$ t3 $\to$ t0 $\to$ t1 $\to$ t4 \\
Generalization (asc.) & t4 $\to$ t1 $\to$ t0 $\to$ t3 $\to$ t2 \\
Smooth drift & t2 $\to$ t0 $\to$ t1 $\to$ t3 $\to$ t4 \\
Abrupt drift & t1 $\to$ t2 $\to$ t3 $\to$ t0 $\to$ t4 \\
\bottomrule
\end{tabular}
\end{table}
\begin{table}[ht]
\centering
\caption{Per-task statistics for CAD-CICIDS2018}
\label{tab:ds_stats_CAD-CICIDS2018}
\begin{tabular}{llrrr}
\toprule
Task ID & Task Name & Train Samples & Test Samples & Test Anomalies \\
\midrule
0 & cicids2018\_0 & 251,873 & 67,807 & 4,855 \\
1 & cicids2018\_1 & 652,959 & 176,660 & 13,421 \\
2 & cicids2018\_2 & 642,521 & 316,606 & 158,303 \\
3 & cicids2018\_3 & 211,855 & 62,074 & 9,161 \\
4 & cicids2018\_4 & 166,025 & 42,391 & 907 \\
\bottomrule
\end{tabular}
\end{table}

{\color{brown}
Tables~\ref{tab:orderingsCAD-CICIDS2018} and \ref{tab:ds_stats_CAD-CICIDS2018} summarize the final CAD-CICIDS2018 scenario. Compared with CAD-CICIDS2017, the retained tasks are larger and more balanced in sample size, but the anomaly ratios still vary substantially across tasks. This makes the scenario complementary: it tests CAD behavior under a different balance of scale, anomaly prevalence, and ordering sensitivity while preserving enough task heterogeneity to avoid a trivial sequence.
}

\FloatBarrier
\subsection{CAD-CICUNSW}
\begin{table}[h]
\centering
\small
\caption{Task orderings for CAD-CICUNSW}
\label{tab:orderingsCAD-CICUNSW}
\begin{tabular}{ll}
\toprule
\textbf{Ordering} & \textbf{Task sequence} \\
\midrule
Curriculum (asc.) & t4 $\to$ t1 $\to$ t3 $\to$ t2 $\to$ t0 \\
Curriculum (desc.) & t0 $\to$ t2 $\to$ t3 $\to$ t1 $\to$ t4 \\
Generalization (desc.) & t4 $\to$ t1 $\to$ t2 $\to$ t3 $\to$ t0 \\
Generalization (asc.) & t0 $\to$ t3 $\to$ t2 $\to$ t1 $\to$ t4 \\
Smooth drift & t4 $\to$ t2 $\to$ t1 $\to$ t3 $\to$ t0 \\
Abrupt drift & t1 $\to$ t2 $\to$ t3 $\to$ t0 $\to$ t4 \\
\bottomrule
\end{tabular}
\end{table}
\begin{table}[ht]
\centering
\caption{Per-task statistics for CAD-CICUNSW}
\label{tab:ds_stats_CAD-CICUNSW}
\begin{tabular}{llrrr}
\toprule
Task ID & Task Name & Train Samples & Test Samples & Test Anomalies \\
\midrule
0 & cicunsw\_0 & 180,512 & 54,078 & 8,950 \\
1 & cicunsw\_1 & 406,320 & 110,531 & 8,950 \\
2 & cicunsw\_2 & 240,550 & 69,088 & 8,950 \\
3 & cicunsw\_3 & 3,503 & 1,752 & 876 \\
4 & cicunsw\_4 & 12,396 & 6,198 & 3,099 \\
\bottomrule
\end{tabular}
\end{table}

{\color{brown}
Tables~\ref{tab:orderingsCAD-CICUNSW} and \ref{tab:ds_stats_CAD-CICUNSW} describe the CAD-CICUNSW scenario. The retained split combines several large tasks with smaller high-anomaly tasks, yielding a setting where both task imbalance and distributional heterogeneity matter. This supports the role of CAD-CICUNSW as a complementary cybersecurity scenario rather than a simple variant of the CICIDS-derived tasks.
}

\FloatBarrier

\subsection{MCAD-CIC-3x1}
\begin{table}[h]
\centering
\small
\caption{Task orderings for MCAD-CIC-3x1}
\label{tab:orderingsMCAD-CIC-3x1}
\begin{tabular}{ll}
\toprule
\textbf{Ordering} & \textbf{Task sequence} \\
\midrule
Curriculum (asc.) & cicunsw $\to$ cicids2017 $\to$ cicids2018 \\
Curriculum (desc.) & cicids2018 $\to$ cicids2017 $\to$ cicunsw \\
Generalization (desc.) & cicids2018 $\to$ cicunsw $\to$ cicids2017 \\
Generalization (asc.) & cicids2017 $\to$ cicunsw $\to$ cicids2018 \\
Smooth drift & cicids2018 $\to$ cicids2017 $\to$ cicunsw \\
Abrupt drift & cicids2017 $\to$ cicids2018 $\to$ cicunsw \\
\bottomrule
\end{tabular}
\end{table}
\begin{table}[ht]
\centering
\caption{Per-task statistics for MCAD-CIC-3x1}
\label{tab:ds_stats_MCAD-CIC-3x1}
\begin{tabular}{llrrr}
\toprule
Task ID & Task Name & Train Samples & Test Samples & Test Anomalies \\
\midrule
0 & cicids2017 & 1,571,292 & 949,506 & 425,741 \\
1 & cicids2018 & 7,923,578 & 3,992,560 & 1,351,367 \\
2 & cicunsw & 2,541,843 & 936,790 & 89,508 \\
\bottomrule
\end{tabular}
\end{table}

{\color{brown}
Tables~\ref{tab:orderingsMCAD-CIC-3x1} and \ref{tab:ds_stats_MCAD-CIC-3x1} define the coarse multi-dataset scenario in which each source dataset contributes one task. This setting is intentionally simple in number of tasks, but not necessarily easy: each task corresponds to a different dataset-level distribution, so the scenario emphasizes cross-dataset transfer and dataset-level domain shift.
}

\FloatBarrier
\subsection{MCAD-CIC-3xN}
\begin{table}[h]
\centering
\small
\caption{Task orderings for MCAD-CIC-3xN: \textit{c17} corresponds to tasks from CICIDS2017, \textit{c18} corresponds to tasks from CICIDS2018, \textit{cu} corresponds to tasks from CICUNSW}
\label{tab:orderingsMCAD-CIC-3xN}
\begin{tabular}{ll}
\toprule
\textbf{Ordering} & \textbf{Task sequence} \\
\midrule
Curriculum (asc.) & cu\textsubscript{3} $\to$ c17\textsubscript{1} $\to$ c17\textsubscript{4} $\to$ c17\textsubscript{0} $\to$ cu\textsubscript{1} $\to$ c17\textsubscript{2} $\to$ c18\textsubscript{2} $\to$ c17\textsubscript{3} $\to$ cu\textsubscript{0} $\to$ cu\textsubscript{2} $\to$ c18\textsubscript{3} $\to$ c18\textsubscript{1} $\to$ c18\textsubscript{0} \\
Curriculum (desc.) & c18\textsubscript{0} $\to$ c18\textsubscript{1} $\to$ c18\textsubscript{3} $\to$ cu\textsubscript{2} $\to$ cu\textsubscript{0} $\to$ c17\textsubscript{3} $\to$ c18\textsubscript{2} $\to$ c17\textsubscript{2} $\to$ cu\textsubscript{1} $\to$ c17\textsubscript{0} $\to$ c17\textsubscript{4} $\to$ c17\textsubscript{1} $\to$ cu\textsubscript{3} \\
Generalization (desc.) & cu\textsubscript{3} $\to$ cu\textsubscript{1} $\to$ cu\textsubscript{0} $\to$ c17\textsubscript{4} $\to$ cu\textsubscript{2} $\to$ c18\textsubscript{0} $\to$ c18\textsubscript{2} $\to$ c17\textsubscript{1} $\to$ c18\textsubscript{1} $\to$ c17\textsubscript{3} $\to$ c17\textsubscript{2} $\to$ c17\textsubscript{0} $\to$ c18\textsubscript{3} \\
Generalization (asc.) & c18\textsubscript{3} $\to$ c17\textsubscript{0} $\to$ c17\textsubscript{2} $\to$ c17\textsubscript{3} $\to$ c18\textsubscript{1} $\to$ c17\textsubscript{1} $\to$ c18\textsubscript{2} $\to$ c18\textsubscript{0} $\to$ cu\textsubscript{2} $\to$ c17\textsubscript{4} $\to$ cu\textsubscript{0} $\to$ cu\textsubscript{1} $\to$ cu\textsubscript{3} \\
Smooth drift & c18\textsubscript{2} $\to$ c18\textsubscript{3} $\to$ cu\textsubscript{0} $\to$ c18\textsubscript{1} $\to$ c17\textsubscript{4} $\to$ cu\textsubscript{1} $\to$ cu\textsubscript{3} $\to$ cu\textsubscript{2} $\to$ c18\textsubscript{0} $\to$ c17\textsubscript{0} $\to$ c17\textsubscript{3} $\to$ c17\textsubscript{1} $\to$ c17\textsubscript{2} \\
Abrupt drift & cu\textsubscript{0} $\to$ c17\textsubscript{2} $\to$ cu\textsubscript{3} $\to$ c17\textsubscript{1} $\to$ cu\textsubscript{2} $\to$ c18\textsubscript{2} $\to$ c17\textsubscript{4} $\to$ c17\textsubscript{0} $\to$ cu\textsubscript{1} $\to$ c17\textsubscript{3} $\to$ c18\textsubscript{1} $\to$ c18\textsubscript{3} $\to$ c18\textsubscript{0} \\
\bottomrule
\end{tabular}
\end{table}
\begin{table}[ht]
\centering
\caption{Per-task statistics for MCAD-CIC-3xN}
\label{tab:ds_stats_MCAD-CIC-3xN}
\begin{tabular}{llrrr}
\toprule
Task ID & Task Name & Train Samples & Test Samples & Test Anomalies \\
\midrule
0 & cicids2017\_0 & 77,385 & 38,694 & 19,347 \\
1 & cicids2017\_1 & 19,006 & 9,504 & 4,752 \\
2 & cicids2017\_2 & 8,392 & 4,196 & 2,098 \\
3 & cicids2017\_3 & 47,520 & 23,760 & 11,880 \\
4 & cicids2017\_4 & 1,713 & 858 & 429 \\
5 & cicids2018\_0 & 652,959 & 176,660 & 13,421 \\
6 & cicids2018\_1 & 642,521 & 316,606 & 158,303 \\
7 & cicids2018\_2 & 211,855 & 62,074 & 9,161 \\
8 & cicids2018\_3 & 166,025 & 42,391 & 907 \\
9 & cicunsw\_0 & 180,512 & 54,078 & 8,950 \\
10 & cicunsw\_1 & 406,320 & 110,531 & 8,950 \\
11 & cicunsw\_2 & 240,550 & 69,088 & 8,950 \\
12 & cicunsw\_3 & 12,396 & 6,198 & 3,099 \\
\bottomrule
\end{tabular}
\end{table}

{\color{brown}
Tables~\ref{tab:orderingsMCAD-CIC-3xN} and \ref{tab:ds_stats_MCAD-CIC-3xN} specify the fine-grained multi-dataset scenario. Unlike MCAD-CIC-3x1, this scenario retains multiple concepts from each source dataset and interleaves them through principled orderings. It is therefore the most heterogeneous scenario in the suite, combining within-dataset task variation with cross-dataset distribution shift.
}

\FloatBarrier
\section{Validation results in terms of Normalized PR-AUC}
\label{appendix:sec:npr_auc}
\begin{table}[h]
\setlength{\tabcolsep}{3pt}
\centering
\small
\caption{PR\text{-}AUC (mean $\pm$ std across orderings) on \textit{CAD-CICIDS2017}, \textit{CAD-CICIDS2018}, \textit{CAD-CICUNSW}, \textit{MCAD-CIC-3x1}, \textit{MCAD-CIC-3xN}. FM$\downarrow$: lower is better. Best per column in \textbf{bold}.}
\label{tab:results_combined_pr_auc}
\begin{tabular}{llcccccccc}
\toprule
\textbf{Dataset} & \textbf{Model} & \multicolumn{2}{c}{\textbf{Naive}} & \multicolumn{2}{c}{\textbf{Replay}} & \multicolumn{2}{c}{\textbf{Cumulative}} & \multicolumn{2}{c}{\textbf{MSTE}} \\
\cmidrule(lr){3-4}\cmidrule(lr){5-6}\cmidrule(lr){7-8}\cmidrule(lr){9-10}
 &  & \textbf{PR}$\uparrow$ & \textbf{FM}$\downarrow$ & \textbf{PR}$\uparrow$ & \textbf{FM}$\downarrow$ & \textbf{PR}$\uparrow$ & \textbf{FM}$\downarrow$ & \textbf{PR}$\uparrow$ & \textbf{FM}$\downarrow$ \\
\midrule
\multirow{4}{*}{CAD-CICIDS2017} & VAE & 64.7{\scriptsize $\pm$3.3} & 34.6{\scriptsize $\pm$2.9} & 83.1{\scriptsize $\pm$2.9} & 10.0{\scriptsize $\pm$2.5} & 79.1{\scriptsize $\pm$6.9} & 6.0{\scriptsize $\pm$4.1} & \textbf{94.7{\scriptsize $\pm$2.8}} & \textbf{0.0{\scriptsize $\pm$0.0}} \\
 & Deep SVDD & 61.0{\scriptsize $\pm$5.3} & 34.5{\scriptsize $\pm$5.0} & 68.7{\scriptsize $\pm$9.2} & 12.7{\scriptsize $\pm$2.9} & 72.9{\scriptsize $\pm$9.2} & 11.1{\scriptsize $\pm$3.6} & \textbf{89.9{\scriptsize $\pm$3.0}} & \textbf{0.0{\scriptsize $\pm$0.0}} \\
 & DAGMM & 59.9{\scriptsize $\pm$5.1} & 30.6{\scriptsize $\pm$3.6} & 71.2{\scriptsize $\pm$7.1} & 11.1{\scriptsize $\pm$4.2} & 60.0{\scriptsize $\pm$11.9} & 13.4{\scriptsize $\pm$4.6} & \textbf{83.3{\scriptsize $\pm$9.0}} & \textbf{0.0{\scriptsize $\pm$0.0}} \\
 & NeutralAD & 62.8{\scriptsize $\pm$5.4} & 32.2{\scriptsize $\pm$5.2} & 86.4{\scriptsize $\pm$2.8} & 6.3{\scriptsize $\pm$3.3} & 78.3{\scriptsize $\pm$11.2} & 7.5{\scriptsize $\pm$5.8} & \textbf{91.1{\scriptsize $\pm$5.9}} & \textbf{0.0{\scriptsize $\pm$0.0}} \\
\midrule
\multirow{4}{*}{CAD-CICIDS2018} & VAE & 35.2{\scriptsize $\pm$5.4} & 34.6{\scriptsize $\pm$10.1} & 43.0{\scriptsize $\pm$10.6} & 26.3{\scriptsize $\pm$11.3} & 46.3{\scriptsize $\pm$9.6} & 13.4{\scriptsize $\pm$9.3} & \textbf{64.8{\scriptsize $\pm$8.5}} & \textbf{0.0{\scriptsize $\pm$0.0}} \\
 & Deep SVDD & 32.5{\scriptsize $\pm$4.6} & 34.8{\scriptsize $\pm$6.3} & 38.3{\scriptsize $\pm$8.4} & 25.4{\scriptsize $\pm$6.3} & 34.6{\scriptsize $\pm$10.9} & 16.4{\scriptsize $\pm$9.4} & \textbf{60.1{\scriptsize $\pm$6.9}} & \textbf{0.0{\scriptsize $\pm$0.0}} \\
 & DAGMM & 32.1{\scriptsize $\pm$7.4} & 19.9{\scriptsize $\pm$8.4} & \textbf{39.0{\scriptsize $\pm$8.0}} & 13.5{\scriptsize $\pm$10.9} & 35.3{\scriptsize $\pm$6.9} & 13.2{\scriptsize $\pm$6.6} & 36.2{\scriptsize $\pm$3.4} & \textbf{0.0{\scriptsize $\pm$0.0}} \\
 & NeutralAD & 35.4{\scriptsize $\pm$3.3} & 40.7{\scriptsize $\pm$3.8} & 53.9{\scriptsize $\pm$12.9} & 16.9{\scriptsize $\pm$11.1} & 50.3{\scriptsize $\pm$7.6} & 15.2{\scriptsize $\pm$2.8} & \textbf{65.9{\scriptsize $\pm$3.4}} & \textbf{0.0{\scriptsize $\pm$0.0}} \\
\midrule
\multirow{4}{*}{CAD-CICUNSW} & VAE & 49.5{\scriptsize $\pm$3.9} & 45.1{\scriptsize $\pm$8.0} & 64.3{\scriptsize $\pm$2.0} & 23.0{\scriptsize $\pm$3.9} & 68.0{\scriptsize $\pm$7.4} & 7.9{\scriptsize $\pm$2.0} & \textbf{86.8{\scriptsize $\pm$4.1}} & \textbf{0.0{\scriptsize $\pm$0.0}} \\
 & Deep SVDD & 43.6{\scriptsize $\pm$2.7} & 28.1{\scriptsize $\pm$4.7} & 41.4{\scriptsize $\pm$9.8} & 18.0{\scriptsize $\pm$4.4} & 39.3{\scriptsize $\pm$5.6} & 15.2{\scriptsize $\pm$3.3} & \textbf{62.8{\scriptsize $\pm$5.6}} & \textbf{0.0{\scriptsize $\pm$0.0}} \\
 & DAGMM & 40.3{\scriptsize $\pm$5.9} & 32.7{\scriptsize $\pm$9.4} & 43.4{\scriptsize $\pm$8.3} & 23.6{\scriptsize $\pm$7.4} & 44.1{\scriptsize $\pm$10.1} & 17.5{\scriptsize $\pm$7.4} & \textbf{72.5{\scriptsize $\pm$11.8}} & \textbf{0.0{\scriptsize $\pm$0.0}} \\
 & NeutralAD & 48.9{\scriptsize $\pm$5.3} & 41.7{\scriptsize $\pm$7.0} & 66.4{\scriptsize $\pm$3.5} & 17.7{\scriptsize $\pm$4.6} & 56.2{\scriptsize $\pm$9.9} & 17.3{\scriptsize $\pm$4.2} & \textbf{84.6{\scriptsize $\pm$3.3}} & \textbf{0.0{\scriptsize $\pm$0.0}} \\
\midrule
\multirow{4}{*}{MCAD-CIC-3x1} & VAE & 52.9{\scriptsize $\pm$5.2} & 16.3{\scriptsize $\pm$2.7} & 56.5{\scriptsize $\pm$6.0} & 10.3{\scriptsize $\pm$2.4} & 57.9{\scriptsize $\pm$3.4} & 3.7{\scriptsize $\pm$1.9} & \textbf{64.8{\scriptsize $\pm$3.9}} & \textbf{0.0{\scriptsize $\pm$0.0}} \\
 & Deep SVDD & 47.8{\scriptsize $\pm$7.4} & 1.9{\scriptsize $\pm$2.1} & 47.5{\scriptsize $\pm$5.0} & 3.8{\scriptsize $\pm$1.3} & 47.3{\scriptsize $\pm$4.4} & 2.2{\scriptsize $\pm$0.7} & \textbf{47.9{\scriptsize $\pm$4.7}} & \textbf{0.0{\scriptsize $\pm$0.0}} \\
 & DAGMM & 45.5{\scriptsize $\pm$8.0} & 6.6{\scriptsize $\pm$2.5} & 46.1{\scriptsize $\pm$5.3} & 5.7{\scriptsize $\pm$5.9} & \textbf{48.4{\scriptsize $\pm$2.8}} & 4.9{\scriptsize $\pm$4.4} & 44.0{\scriptsize $\pm$5.6} & \textbf{0.0{\scriptsize $\pm$0.0}} \\
 & NeutralAD & 56.1{\scriptsize $\pm$4.1} & 27.6{\scriptsize $\pm$4.3} & 61.8{\scriptsize $\pm$9.1} & 11.2{\scriptsize $\pm$6.3} & 62.8{\scriptsize $\pm$6.8} & 6.3{\scriptsize $\pm$4.9} & \textbf{68.3{\scriptsize $\pm$9.6}} & \textbf{0.0{\scriptsize $\pm$0.0}} \\
\midrule
\multirow{4}{*}{MCAD-CIC-3xN} & VAE & 42.6{\scriptsize $\pm$4.4} & 50.2{\scriptsize $\pm$4.0} & 67.1{\scriptsize $\pm$7.9} & 18.8{\scriptsize $\pm$5.3} & 63.5{\scriptsize $\pm$11.8} & 13.1{\scriptsize $\pm$4.3} & \textbf{88.3{\scriptsize $\pm$4.4}} & \textbf{0.0{\scriptsize $\pm$0.0}} \\
 & Deep SVDD & 43.0{\scriptsize $\pm$6.4} & 35.6{\scriptsize $\pm$4.6} & 56.8{\scriptsize $\pm$10.1} & 18.6{\scriptsize $\pm$4.2} & 56.4{\scriptsize $\pm$11.1} & 14.0{\scriptsize $\pm$3.2} & \textbf{73.6{\scriptsize $\pm$4.7}} & \textbf{0.0{\scriptsize $\pm$0.0}} \\
 & DAGMM & 38.5{\scriptsize $\pm$5.5} & 38.7{\scriptsize $\pm$3.7} & 57.3{\scriptsize $\pm$10.9} & 18.1{\scriptsize $\pm$3.3} & 54.7{\scriptsize $\pm$8.5} & 20.0{\scriptsize $\pm$5.5} & \textbf{68.8{\scriptsize $\pm$7.9}} & \textbf{0.0{\scriptsize $\pm$0.0}} \\
 & NeutralAD & 45.7{\scriptsize $\pm$5.7} & 45.5{\scriptsize $\pm$1.2} & 74.9{\scriptsize $\pm$7.5} & 11.7{\scriptsize $\pm$3.3} & 70.0{\scriptsize $\pm$10.5} & 11.8{\scriptsize $\pm$4.7} & \textbf{86.4{\scriptsize $\pm$4.3}} & \textbf{0.0{\scriptsize $\pm$0.0}} \\
\bottomrule
\end{tabular}
\end{table}

\FloatBarrier
\section{Per-Dataset Validation Results for Final Scenarios}
\label{appendix:scenario-results}

This appendix provides additional validation results for the final benchmark scenarios constructed by our framework. For each dataset, we report the performance obtained under the considered task orderings using ROC-AUC and normalized PR-AUC, enabling a more detailed comparison of how ordering affects anomaly detection performance. We further include heatmaps for each ordering and continual learning strategy, reported for both evaluation metrics. These results complement the aggregate analysis in the main paper by exposing the full per-dataset validation evidence, including task-level transfer patterns and ordering-specific performance differences.

    \subsection{CICIDS2017}

\begin{table}[h]
\centering
\small
\caption{ROC\text{-}AUC (mean $\pm$ std across models) per ordering on \textit{CAD-CICIDS2017}. FM$\downarrow$: lower is better. Best strategy per ordering in \textbf{bold}.}
\label{tab:ordering_results_cad_cicids2017_roc_auc}
\begin{tabular}{llcccc}
\toprule
\textbf{Ordering} & \textbf{Metric} & \textbf{Naive} & \textbf{Cumulative} & \textbf{MSTE} & \textbf{Replay} \\
\midrule
\multirow{2}{*}{Curriculum (asc.)} & ROC\text{-}AUC$\uparrow$ & 54.7{\scriptsize $\pm$5.5} & 82.3{\scriptsize $\pm$9.9} & \textbf{96.1{\scriptsize $\pm$3.7}} & 81.5{\scriptsize $\pm$6.4} \\
 & FM$\downarrow$ & 51.4{\scriptsize $\pm$11.2} & 12.1{\scriptsize $\pm$7.7} & \textbf{0.0{\scriptsize $\pm$0.0}} & 12.4{\scriptsize $\pm$5.3} \\
\midrule
\multirow{2}{*}{Curriculum (desc.)} & ROC\text{-}AUC$\uparrow$ & 61.5{\scriptsize $\pm$2.1} & 65.2{\scriptsize $\pm$10.5} & \textbf{89.3{\scriptsize $\pm$8.1}} & 76.3{\scriptsize $\pm$15.0} \\
 & FM$\downarrow$ & 36.2{\scriptsize $\pm$4.5} & 5.6{\scriptsize $\pm$5.4} & \textbf{0.0{\scriptsize $\pm$0.0}} & 7.8{\scriptsize $\pm$3.2} \\
\midrule
\multirow{2}{*}{Generalization (asc.)} & ROC\text{-}AUC$\uparrow$ & 56.7{\scriptsize $\pm$4.5} & 67.7{\scriptsize $\pm$9.0} & \textbf{95.3{\scriptsize $\pm$2.3}} & 83.7{\scriptsize $\pm$9.7} \\
 & FM$\downarrow$ & 40.4{\scriptsize $\pm$6.5} & 7.2{\scriptsize $\pm$4.6} & \textbf{0.0{\scriptsize $\pm$0.0}} & 6.5{\scriptsize $\pm$2.8} \\
\midrule
\multirow{2}{*}{Generalization (desc.)} & ROC\text{-}AUC$\uparrow$ & 58.7{\scriptsize $\pm$1.1} & 83.5{\scriptsize $\pm$8.0} & \textbf{95.0{\scriptsize $\pm$4.5}} & 79.9{\scriptsize $\pm$6.8} \\
 & FM$\downarrow$ & 37.2{\scriptsize $\pm$3.1} & 6.9{\scriptsize $\pm$3.2} & \textbf{0.0{\scriptsize $\pm$0.0}} & 10.7{\scriptsize $\pm$1.4} \\
\midrule
\multirow{2}{*}{Smooth drift} & ROC\text{-}AUC$\uparrow$ & 56.0{\scriptsize $\pm$2.5} & 66.8{\scriptsize $\pm$8.2} & \textbf{93.4{\scriptsize $\pm$5.7}} & 81.9{\scriptsize $\pm$9.9} \\
 & FM$\downarrow$ & 44.8{\scriptsize $\pm$5.2} & 18.9{\scriptsize $\pm$3.1} & \textbf{0.0{\scriptsize $\pm$0.0}} & 14.7{\scriptsize $\pm$6.9} \\
\midrule
\multirow{2}{*}{Abrupt drift} & ROC\text{-}AUC$\uparrow$ & 59.1{\scriptsize $\pm$3.3} & 67.9{\scriptsize $\pm$6.9} & \textbf{95.0{\scriptsize $\pm$4.0}} & 79.0{\scriptsize $\pm$9.4} \\
 & FM$\downarrow$ & 43.7{\scriptsize $\pm$4.9} & 17.2{\scriptsize $\pm$3.6} & \textbf{0.0{\scriptsize $\pm$0.0}} & 13.0{\scriptsize $\pm$2.8} \\
\bottomrule
\end{tabular}
\end{table}

\begin{table}[h]
\centering
\small
\caption{Normalized\text{-}PR\text{-}AUC (mean $\pm$ std across models) per ordering on \textit{CAD-CICIDS2017}. FM$\downarrow$: lower is better. Best strategy per ordering in \textbf{bold}.}
\label{tab:ordering_results_cad_cicids2017_normalized_pr_auc}
\begin{tabular}{llcccc}
\toprule
\textbf{Ordering} & \textbf{Metric} & \textbf{Naive} & \textbf{Cumulative} & \textbf{MSTE} & \textbf{Replay} \\
\midrule
\multirow{2}{*}{Curriculum (asc.)} & Normalized\text{-}PR\text{-}AUC$\uparrow$ & 35.1{\scriptsize $\pm$5.5} & 72.8{\scriptsize $\pm$14.6} & \textbf{93.3{\scriptsize $\pm$5.7}} & 70.7{\scriptsize $\pm$9.3} \\
 & FM$\downarrow$ & 70.9{\scriptsize $\pm$13.5} & 16.7{\scriptsize $\pm$10.0} & \textbf{0.0{\scriptsize $\pm$0.0}} & 18.1{\scriptsize $\pm$6.1} \\
\midrule
\multirow{2}{*}{Curriculum (desc.)} & Normalized\text{-}PR\text{-}AUC$\uparrow$ & 35.8{\scriptsize $\pm$4.6} & 41.1{\scriptsize $\pm$19.1} & \textbf{79.2{\scriptsize $\pm$12.3}} & 54.9{\scriptsize $\pm$24.6} \\
 & FM$\downarrow$ & 55.5{\scriptsize $\pm$3.2} & 9.7{\scriptsize $\pm$10.3} & \textbf{0.0{\scriptsize $\pm$0.0}} & 14.5{\scriptsize $\pm$6.2} \\
\midrule
\multirow{2}{*}{Generalization (asc.)} & Normalized\text{-}PR\text{-}AUC$\uparrow$ & 33.8{\scriptsize $\pm$7.5} & 47.3{\scriptsize $\pm$16.8} & \textbf{87.1{\scriptsize $\pm$4.8}} & 67.3{\scriptsize $\pm$17.0} \\
 & FM$\downarrow$ & 54.5{\scriptsize $\pm$10.5} & 11.2{\scriptsize $\pm$7.4} & \textbf{0.0{\scriptsize $\pm$0.0}} & 11.1{\scriptsize $\pm$4.9} \\
\midrule
\multirow{2}{*}{Generalization (desc.)} & Normalized\text{-}PR\text{-}AUC$\uparrow$ & 35.3{\scriptsize $\pm$4.3} & 73.5{\scriptsize $\pm$13.0} & \textbf{90.8{\scriptsize $\pm$7.4}} & 66.4{\scriptsize $\pm$11.3} \\
 & FM$\downarrow$ & 57.9{\scriptsize $\pm$3.2} & 11.1{\scriptsize $\pm$5.6} & \textbf{0.0{\scriptsize $\pm$0.0}} & 17.6{\scriptsize $\pm$3.2} \\
\midrule
\multirow{2}{*}{Smooth drift} & Normalized\text{-}PR\text{-}AUC$\uparrow$ & 32.1{\scriptsize $\pm$3.7} & 46.0{\scriptsize $\pm$18.0} & \textbf{85.4{\scriptsize $\pm$10.5}} & 65.4{\scriptsize $\pm$16.5} \\
 & FM$\downarrow$ & 64.2{\scriptsize $\pm$4.8} & 27.1{\scriptsize $\pm$7.9} & \textbf{0.0{\scriptsize $\pm$0.0}} & 22.5{\scriptsize $\pm$8.9} \\
\midrule
\multirow{2}{*}{Abrupt drift} & Normalized\text{-}PR\text{-}AUC$\uparrow$ & 35.5{\scriptsize $\pm$3.9} & 48.1{\scriptsize $\pm$15.1} & \textbf{89.3{\scriptsize $\pm$6.6}} & 62.0{\scriptsize $\pm$17.5} \\
 & FM$\downarrow$ & 65.0{\scriptsize $\pm$5.4} & 26.7{\scriptsize $\pm$8.4} & \textbf{0.0{\scriptsize $\pm$0.0}} & 21.6{\scriptsize $\pm$6.3} \\
\bottomrule
\end{tabular}
\end{table}

{\color{brown}
The ordering-specific results for CAD-CICIDS2017 confirm the aggregate findings from the main paper. Across the six retained orderings, Naive remains consistently below Replay, Cumulative, and MSTE, while forgetting remains clearly visible for sequential fine-tuning alone. This indicates that the continual-learning difficulty of the scenario is not tied to a single ordering, but persists across curriculum-, generalization-, and drift-based sequences. Figures~\ref{fig:app:val_cicids2017_roc_auc} and \ref{fig:app:val_cicids2017_pr_auc} visualize this pattern under ROC-AUC and normalized PR-AUC.
}

    \begin{figure}
        \centering
        \includegraphics[width=\linewidth]{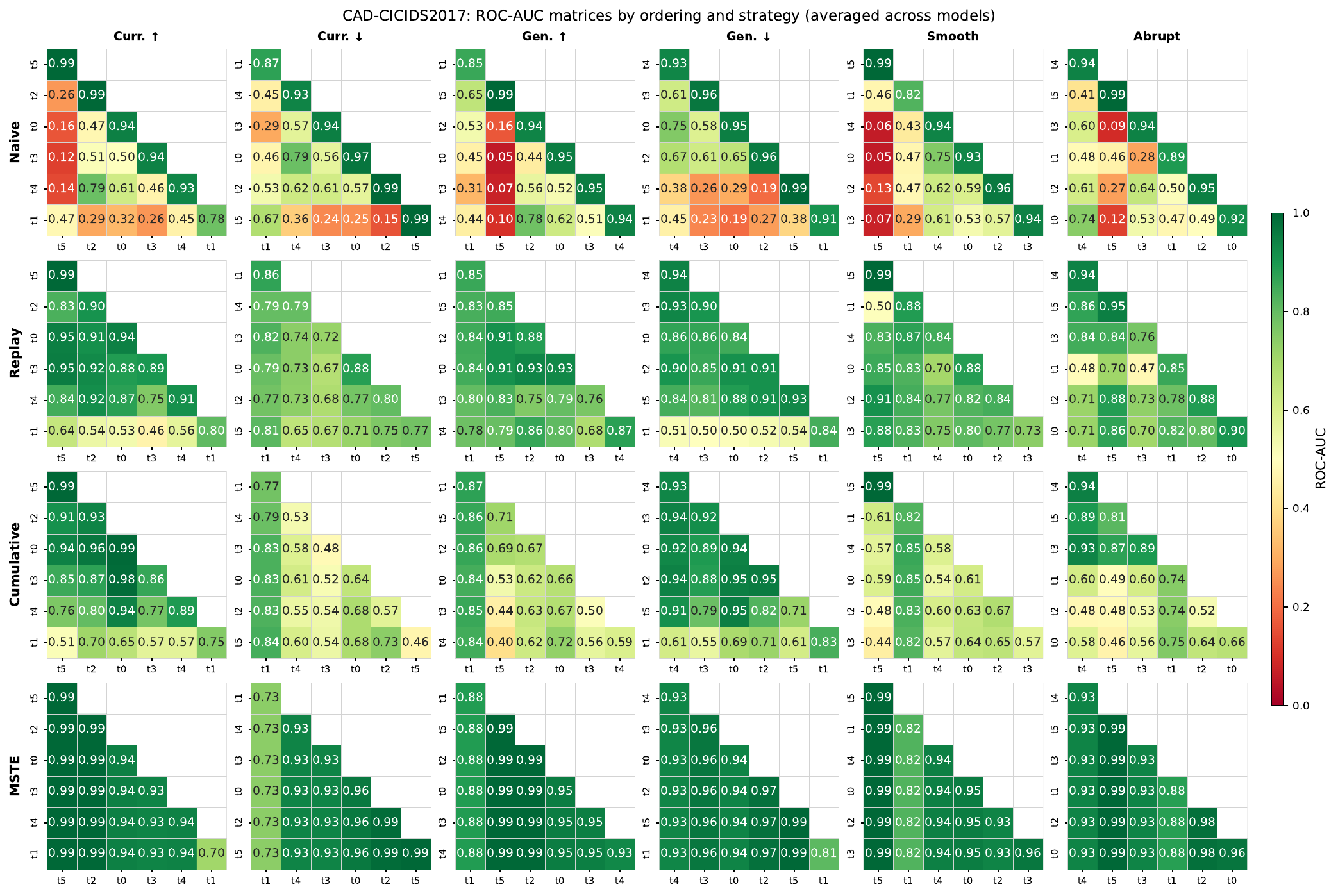}
        \caption{Ordering-specific validation results on CAD-CICIDS2017 measured with ROC-AUC. The heatmaps compare Naive, Replay, Cumulative, and MSTE across the six retained orderings, highlighting the persistent performance gap between Naive and the stronger reference strategies.}
        \label{fig:app:val_cicids2017_roc_auc}
    \end{figure}
    
    \begin{figure}
        \centering
        \includegraphics[width=\linewidth]{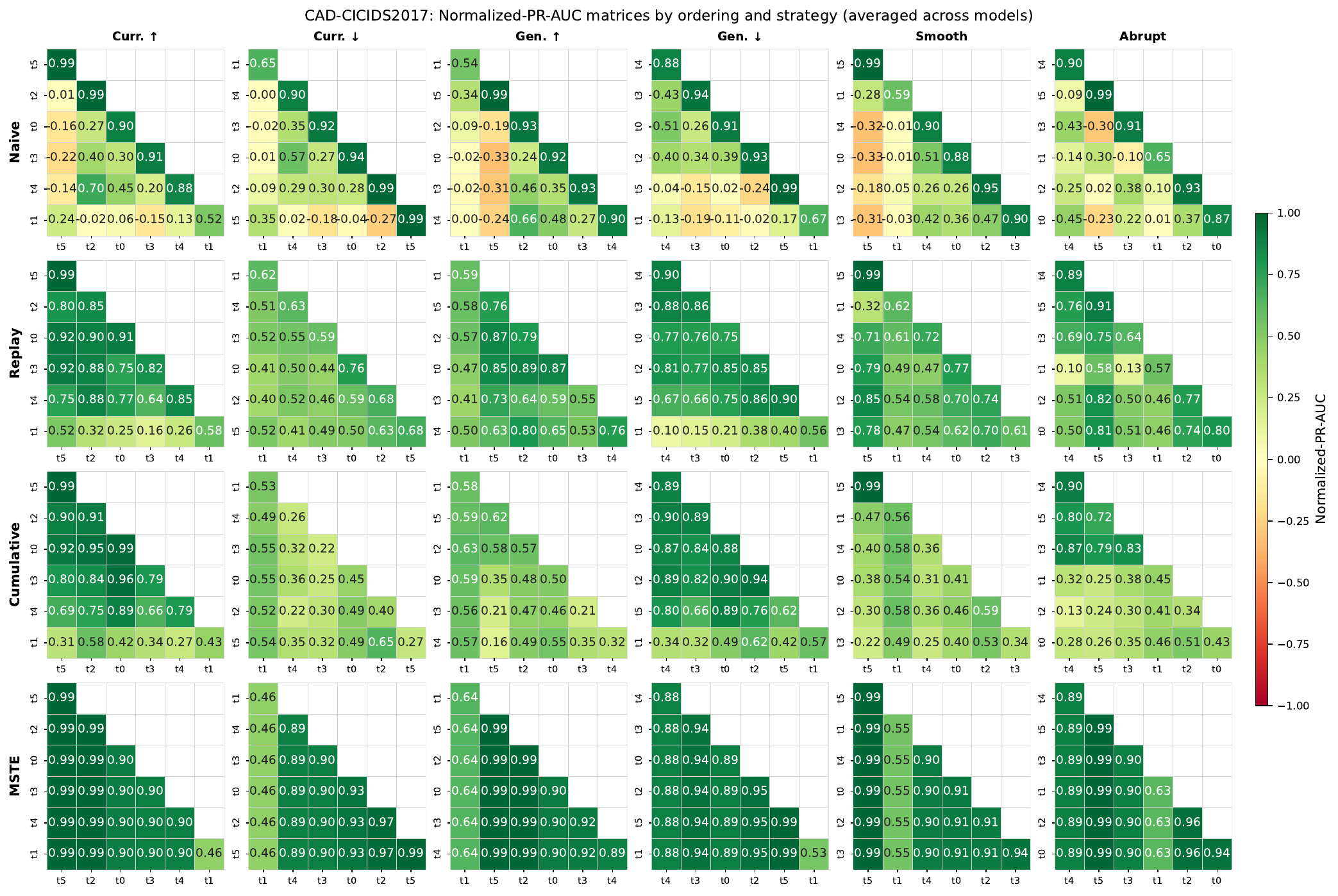}
        \caption{Ordering-specific validation results on CAD-CICIDS2017 measured with normalized PR-AUC. The qualitative pattern matches ROC-AUC and confirms that the scenario remains non-trivial under an imbalance-aware metric.}
        \label{fig:app:val_cicids2017_pr_auc}
    \end{figure}

 \FloatBarrier   
    \subsection{CICIDS2018}

\begin{table}[h]
\centering
\small
\caption{ROC\text{-}AUC (mean $\pm$ std across models) per ordering on \textit{CAD-CICIDS2018}. FM$\downarrow$: lower is better. Best strategy per ordering in \textbf{bold}.}
\label{tab:ordering_results_cad_cicids2018_roc_auc}
\begin{tabular}{llcccc}
\toprule
\textbf{Ordering} & \textbf{Metric} & \textbf{Naive} & \textbf{Cumulative} & \textbf{MSTE} & \textbf{Replay} \\
\midrule
\multirow{2}{*}{Curriculum (asc.)} & ROC\text{-}AUC$\uparrow$ & 55.2{\scriptsize $\pm$4.0} & 71.8{\scriptsize $\pm$12.3} & \textbf{84.2{\scriptsize $\pm$9.5}} & 75.5{\scriptsize $\pm$6.4} \\
 & FM$\downarrow$ & 40.0{\scriptsize $\pm$9.8} & 12.3{\scriptsize $\pm$7.7} & \textbf{0.3{\scriptsize $\pm$0.7}} & 15.1{\scriptsize $\pm$6.7} \\
\midrule
\multirow{2}{*}{Curriculum (desc.)} & ROC\text{-}AUC$\uparrow$ & 56.6{\scriptsize $\pm$2.1} & 66.2{\scriptsize $\pm$7.3} & \textbf{82.8{\scriptsize $\pm$8.3}} & 65.9{\scriptsize $\pm$4.4} \\
 & FM$\downarrow$ & 27.8{\scriptsize $\pm$4.3} & 9.7{\scriptsize $\pm$2.4} & \textbf{0.0{\scriptsize $\pm$0.0}} & 18.1{\scriptsize $\pm$6.0} \\
\midrule
\multirow{2}{*}{Generalization (asc.)} & ROC\text{-}AUC$\uparrow$ & 62.2{\scriptsize $\pm$4.9} & 68.2{\scriptsize $\pm$5.9} & \textbf{85.1{\scriptsize $\pm$5.8}} & 65.0{\scriptsize $\pm$6.0} \\
 & FM$\downarrow$ & 29.5{\scriptsize $\pm$3.6} & 13.5{\scriptsize $\pm$2.1} & \textbf{0.0{\scriptsize $\pm$0.0}} & 25.7{\scriptsize $\pm$4.9} \\
\midrule
\multirow{2}{*}{Generalization (desc.)} & ROC\text{-}AUC$\uparrow$ & 50.0{\scriptsize $\pm$3.6} & 72.8{\scriptsize $\pm$9.2} & \textbf{79.3{\scriptsize $\pm$13.7}} & 70.6{\scriptsize $\pm$2.3} \\
 & FM$\downarrow$ & 45.1{\scriptsize $\pm$13.1} & 7.4{\scriptsize $\pm$2.3} & \textbf{0.1{\scriptsize $\pm$0.3}} & 13.2{\scriptsize $\pm$5.4} \\
\midrule
\multirow{2}{*}{Smooth drift} & ROC\text{-}AUC$\uparrow$ & 50.1{\scriptsize $\pm$5.0} & 70.1{\scriptsize $\pm$7.3} & \textbf{77.7{\scriptsize $\pm$10.5}} & 70.6{\scriptsize $\pm$5.8} \\
 & FM$\downarrow$ & 37.8{\scriptsize $\pm$14.2} & 11.6{\scriptsize $\pm$5.6} & \textbf{0.0{\scriptsize $\pm$0.0}} & 16.0{\scriptsize $\pm$2.1} \\
\midrule
\multirow{2}{*}{Abrupt drift} & ROC\text{-}AUC$\uparrow$ & 53.3{\scriptsize $\pm$3.7} & 71.4{\scriptsize $\pm$10.7} & \textbf{81.3{\scriptsize $\pm$13.2}} & 70.4{\scriptsize $\pm$6.8} \\
 & FM$\downarrow$ & 34.0{\scriptsize $\pm$6.5} & 10.4{\scriptsize $\pm$6.2} & \textbf{0.6{\scriptsize $\pm$1.4}} & 16.2{\scriptsize $\pm$9.7} \\
\bottomrule
\end{tabular}
\end{table}

\begin{table}[h]
\centering
\small
\caption{Normalized\text{-}PR\text{-}AUC (mean $\pm$ std across models) per ordering on \textit{CAD-CICIDS2018}. FM$\downarrow$: lower is better. Best strategy per ordering in \textbf{bold}.}
\label{tab:ordering_results_cad_cicids2018_normalized_pr_auc}
\begin{tabular}{llcccc}
\toprule
\textbf{Ordering} & \textbf{Metric} & \textbf{Naive} & \textbf{Cumulative} & \textbf{MSTE} & \textbf{Replay} \\
\midrule
\multirow{2}{*}{Curriculum (asc.)} & Normalized\text{-}PR\text{-}AUC$\uparrow$ & 24.5{\scriptsize $\pm$5.4} & 39.3{\scriptsize $\pm$17.3} & \textbf{57.3{\scriptsize $\pm$20.0}} & 47.1{\scriptsize $\pm$14.2} \\
 & FM$\downarrow$ & 49.3{\scriptsize $\pm$13.8} & 17.4{\scriptsize $\pm$9.8} & \textbf{0.5{\scriptsize $\pm$1.0}} & 23.7{\scriptsize $\pm$12.2} \\
\midrule
\multirow{2}{*}{Curriculum (desc.)} & Normalized\text{-}PR\text{-}AUC$\uparrow$ & 15.4{\scriptsize $\pm$4.6} & 22.9{\scriptsize $\pm$10.8} & \textbf{50.7{\scriptsize $\pm$13.5}} & 22.4{\scriptsize $\pm$5.4} \\
 & FM$\downarrow$ & 36.0{\scriptsize $\pm$8.7} & 11.2{\scriptsize $\pm$5.3} & \textbf{0.0{\scriptsize $\pm$0.0}} & 26.0{\scriptsize $\pm$8.1} \\
\midrule
\multirow{2}{*}{Generalization (asc.)} & Normalized\text{-}PR\text{-}AUC$\uparrow$ & 23.3{\scriptsize $\pm$6.8} & 28.5{\scriptsize $\pm$10.9} & \textbf{52.8{\scriptsize $\pm$15.0}} & 27.2{\scriptsize $\pm$5.9} \\
 & FM$\downarrow$ & 43.3{\scriptsize $\pm$8.7} & 23.0{\scriptsize $\pm$6.1} & \textbf{0.0{\scriptsize $\pm$0.0}} & 39.4{\scriptsize $\pm$4.9} \\
\midrule
\multirow{2}{*}{Generalization (desc.)} & Normalized\text{-}PR\text{-}AUC$\uparrow$ & 15.7{\scriptsize $\pm$3.7} & 39.6{\scriptsize $\pm$12.4} & \textbf{49.1{\scriptsize $\pm$23.1}} & 36.5{\scriptsize $\pm$5.8} \\
 & FM$\downarrow$ & 57.1{\scriptsize $\pm$17.8} & 13.7{\scriptsize $\pm$6.0} & \textbf{0.0{\scriptsize $\pm$0.0}} & 20.1{\scriptsize $\pm$10.9} \\
\midrule
\multirow{2}{*}{Smooth drift} & Normalized\text{-}PR\text{-}AUC$\uparrow$ & 12.7{\scriptsize $\pm$3.1} & 31.2{\scriptsize $\pm$9.4} & \textbf{42.4{\scriptsize $\pm$19.5}} & 29.9{\scriptsize $\pm$8.9} \\
 & FM$\downarrow$ & 48.9{\scriptsize $\pm$22.4} & 18.8{\scriptsize $\pm$9.6} & \textbf{0.0{\scriptsize $\pm$0.0}} & 24.1{\scriptsize $\pm$6.2} \\
\midrule
\multirow{2}{*}{Abrupt drift} & Normalized\text{-}PR\text{-}AUC$\uparrow$ & 15.6{\scriptsize $\pm$2.3} & 35.2{\scriptsize $\pm$11.3} & \textbf{53.4{\scriptsize $\pm$24.2}} & 35.1{\scriptsize $\pm$11.2} \\
 & FM$\downarrow$ & 47.6{\scriptsize $\pm$16.0} & 23.6{\scriptsize $\pm$6.9} & \textbf{0.9{\scriptsize $\pm$2.0}} & 28.4{\scriptsize $\pm$15.1} \\
\bottomrule
\end{tabular}
\end{table}

{\color{brown}
CAD-CICIDS2018 exhibits a similar but somewhat more variable picture. Stronger strategies still dominate Naive overall, but the margins depend more visibly on the detector and on the ordering, which makes this dataset a useful complement to CICIDS2017. This reinforces the motivation for evaluating multiple principled sequences rather than relying on a single task order. Figures~\ref{fig:app:val_cicids2018_roc_auc} and \ref{fig:app:val_cicids2018_pr_auc} summarize these effects under both metrics.
}

    \begin{figure}
    \centering
    \includegraphics[width=\linewidth]{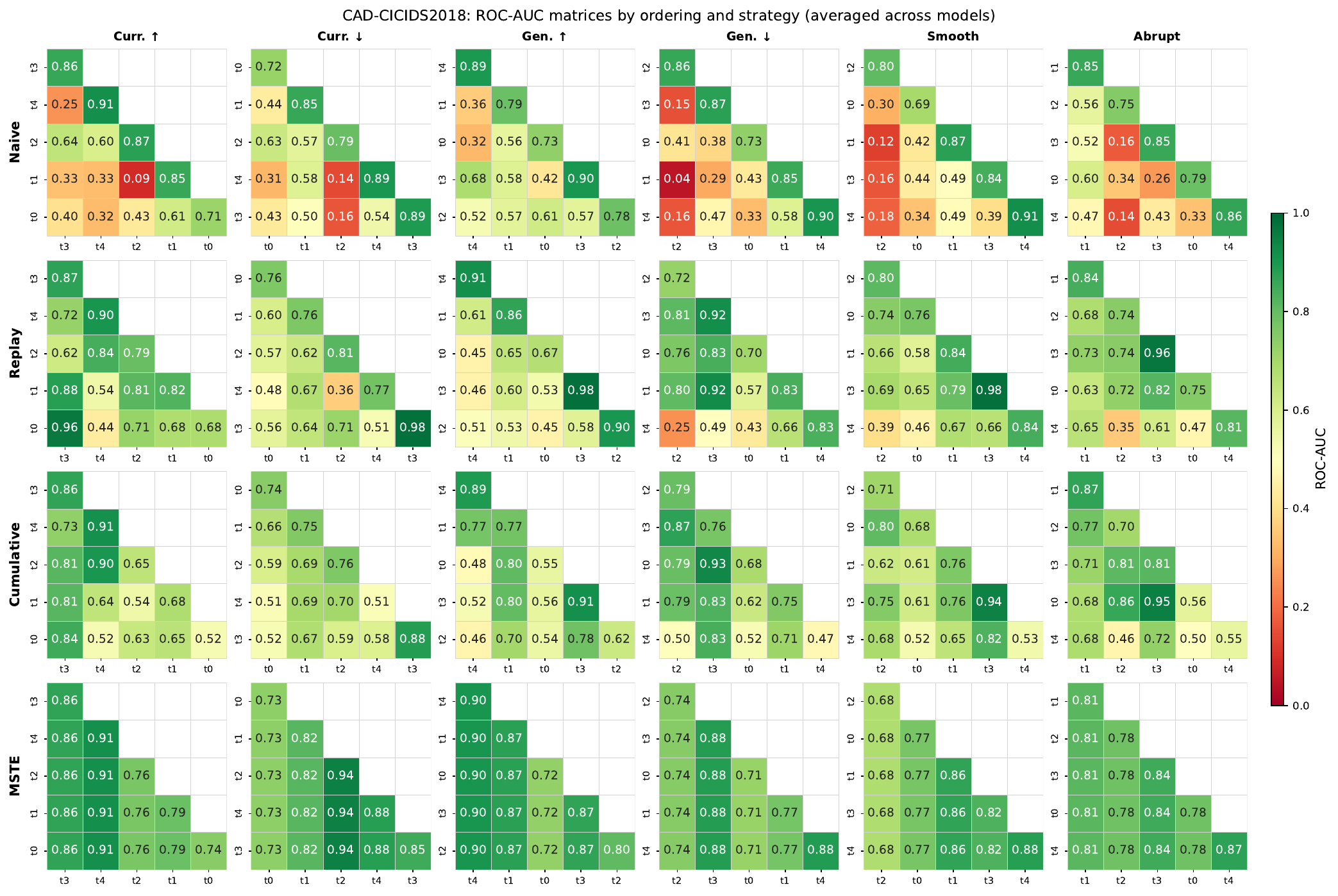}
    \caption{Ordering-specific validation results on CAD-CICIDS2018 measured with ROC-AUC. Compared with CAD-CICIDS2017, the results show somewhat stronger ordering and model sensitivity while preserving the overall advantage of stronger continual references over Naive.}
    \label{fig:app:val_cicids2018_roc_auc}
\end{figure}

\begin{figure}
    \centering
    \includegraphics[width=\linewidth]{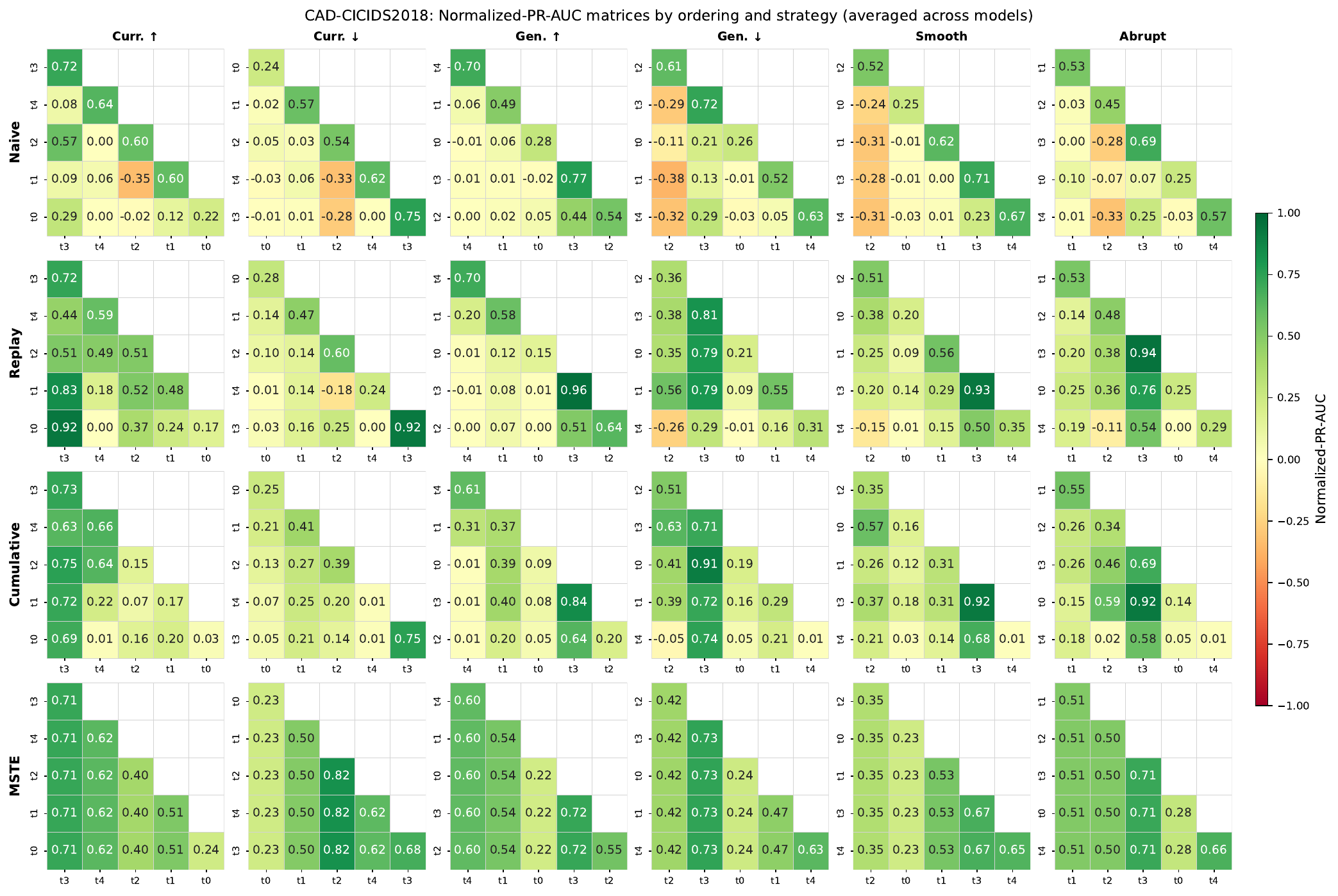}
    \caption{Ordering-specific validation results on CAD-CICIDS2018 measured with normalized PR-AUC. The figure confirms that the scenario remains challenging under imbalance-aware evaluation and that ordering effects remain visible at the task-sequence level.}
    \label{fig:app:val_cicids2018_pr_auc}
\end{figure}

\FloatBarrier
    \subsection{CICUNSW}
        
\begin{table}[h]
\centering
\small
\caption{ROC\text{-}AUC (mean $\pm$ std across models) per ordering on \textit{CAD-CICUNSW}. FM$\downarrow$: lower is better. Best strategy per ordering in \textbf{bold}.}
\label{tab:ordering_results_cad_cicunsw_roc_auc}
\begin{tabular}{llcccc}
\toprule
\textbf{Ordering} & \textbf{Metric} & \textbf{Naive} & \textbf{Cumulative} & \textbf{MSTE} & \textbf{Replay} \\
\midrule
\multirow{2}{*}{Curriculum (asc.)} & ROC\text{-}AUC$\uparrow$ & 46.4{\scriptsize $\pm$6.2} & 74.4{\scriptsize $\pm$16.7} & \textbf{86.8{\scriptsize $\pm$16.2}} & 70.9{\scriptsize $\pm$8.0} \\
 & FM$\downarrow$ & 48.2{\scriptsize $\pm$16.1} & 14.7{\scriptsize $\pm$7.9} & 7.5{\scriptsize $\pm$18.1} & 20.1{\scriptsize $\pm$3.7} \\
\midrule
\multirow{2}{*}{Curriculum (desc.)} & ROC\text{-}AUC$\uparrow$ & 62.8{\scriptsize $\pm$4.3} & 64.5{\scriptsize $\pm$15.4} & \textbf{87.5{\scriptsize $\pm$10.8}} & 72.8{\scriptsize $\pm$19.5} \\
 & FM$\downarrow$ & 32.2{\scriptsize $\pm$5.8} & 10.2{\scriptsize $\pm$5.9} & \textbf{0.5{\scriptsize $\pm$1.2}} & 14.1{\scriptsize $\pm$7.0} \\
\midrule
\multirow{2}{*}{Generalization (asc.)} & ROC\text{-}AUC$\uparrow$ & 61.1{\scriptsize $\pm$4.7} & 67.3{\scriptsize $\pm$12.9} & \textbf{88.1{\scriptsize $\pm$9.7}} & 68.4{\scriptsize $\pm$20.1} \\
 & FM$\downarrow$ & 34.2{\scriptsize $\pm$5.2} & 11.7{\scriptsize $\pm$4.7} & \textbf{0.3{\scriptsize $\pm$0.7}} & 15.5{\scriptsize $\pm$4.1} \\
\midrule
\multirow{2}{*}{Generalization (desc.)} & ROC\text{-}AUC$\uparrow$ & 49.5{\scriptsize $\pm$3.5} & 72.3{\scriptsize $\pm$14.7} & \textbf{91.0{\scriptsize $\pm$9.4}} & 73.3{\scriptsize $\pm$10.8} \\
 & FM$\downarrow$ & 50.7{\scriptsize $\pm$7.5} & 14.7{\scriptsize $\pm$11.1} & \textbf{0.2{\scriptsize $\pm$0.5}} & 19.4{\scriptsize $\pm$5.6} \\
\midrule
\multirow{2}{*}{Smooth drift} & ROC\text{-}AUC$\uparrow$ & 49.3{\scriptsize $\pm$4.7} & 70.8{\scriptsize $\pm$18.5} & \textbf{91.6{\scriptsize $\pm$8.7}} & 71.9{\scriptsize $\pm$15.3} \\
 & FM$\downarrow$ & 50.3{\scriptsize $\pm$5.5} & 19.5{\scriptsize $\pm$15.9} & \textbf{0.3{\scriptsize $\pm$0.6}} & 20.9{\scriptsize $\pm$12.1} \\
\midrule
\multirow{2}{*}{Abrupt drift} & ROC\text{-}AUC$\uparrow$ & 55.1{\scriptsize $\pm$2.3} & 68.3{\scriptsize $\pm$16.2} & \textbf{87.5{\scriptsize $\pm$11.9}} & 72.5{\scriptsize $\pm$18.2} \\
 & FM$\downarrow$ & 43.5{\scriptsize $\pm$13.0} & 8.2{\scriptsize $\pm$5.1} & \textbf{0.4{\scriptsize $\pm$0.9}} & 16.3{\scriptsize $\pm$2.5} \\
\bottomrule
\end{tabular}
\end{table}

\begin{table}[h]
\centering
\small
\caption{Normalized\text{-}PR\text{-}AUC (mean $\pm$ std across models) per ordering on \textit{CAD-CICUNSW}. FM$\downarrow$: lower is better. Best strategy per ordering in \textbf{bold}.}
\label{tab:ordering_results_cad_cicunsw_normalized_pr_auc}
\begin{tabular}{llcccc}
\toprule
\textbf{Ordering} & \textbf{Metric} & \textbf{Naive} & \textbf{Cumulative} & \textbf{MSTE} & \textbf{Replay} \\
\midrule
\multirow{2}{*}{Curriculum (asc.)} & Normalized\text{-}PR\text{-}AUC$\uparrow$ & 21.9{\scriptsize $\pm$9.8} & 51.2{\scriptsize $\pm$20.0} & \textbf{75.8{\scriptsize $\pm$20.9}} & 47.6{\scriptsize $\pm$10.4} \\
 & FM$\downarrow$ & 63.8{\scriptsize $\pm$25.9} & 24.9{\scriptsize $\pm$8.7} & 9.8{\scriptsize $\pm$24.0} & 31.8{\scriptsize $\pm$5.4} \\
\midrule
\multirow{2}{*}{Curriculum (desc.)} & Normalized\text{-}PR\text{-}AUC$\uparrow$ & 30.0{\scriptsize $\pm$9.1} & 28.3{\scriptsize $\pm$14.7} & \textbf{68.0{\scriptsize $\pm$15.8}} & 39.5{\scriptsize $\pm$20.7} \\
 & FM$\downarrow$ & 45.5{\scriptsize $\pm$10.1} & 14.6{\scriptsize $\pm$6.8} & \textbf{0.0{\scriptsize $\pm$0.0}} & 24.2{\scriptsize $\pm$4.7} \\
\midrule
\multirow{2}{*}{Generalization (asc.)} & Normalized\text{-}PR\text{-}AUC$\uparrow$ & 29.1{\scriptsize $\pm$6.9} & 32.3{\scriptsize $\pm$11.6} & \textbf{71.2{\scriptsize $\pm$13.9}} & 35.4{\scriptsize $\pm$22.7} \\
 & FM$\downarrow$ & 50.4{\scriptsize $\pm$10.1} & 20.2{\scriptsize $\pm$7.4} & \textbf{0.0{\scriptsize $\pm$0.0}} & 27.8{\scriptsize $\pm$5.5} \\
\midrule
\multirow{2}{*}{Generalization (desc.)} & Normalized\text{-}PR\text{-}AUC$\uparrow$ & 26.1{\scriptsize $\pm$6.1} & 44.2{\scriptsize $\pm$17.8} & \textbf{80.3{\scriptsize $\pm$12.0}} & 48.0{\scriptsize $\pm$12.5} \\
 & FM$\downarrow$ & 66.2{\scriptsize $\pm$10.6} & 25.1{\scriptsize $\pm$13.3} & \textbf{0.0{\scriptsize $\pm$0.0}} & 31.4{\scriptsize $\pm$6.4} \\
\midrule
\multirow{2}{*}{Smooth drift} & Normalized\text{-}PR\text{-}AUC$\uparrow$ & 24.8{\scriptsize $\pm$3.5} & 41.7{\scriptsize $\pm$22.9} & \textbf{77.7{\scriptsize $\pm$11.6}} & 44.3{\scriptsize $\pm$20.4} \\
 & FM$\downarrow$ & 63.7{\scriptsize $\pm$9.0} & 26.4{\scriptsize $\pm$16.5} & \textbf{0.0{\scriptsize $\pm$0.0}} & 31.5{\scriptsize $\pm$10.7} \\
\midrule
\multirow{2}{*}{Abrupt drift} & Normalized\text{-}PR\text{-}AUC$\uparrow$ & 28.8{\scriptsize $\pm$7.7} & 32.9{\scriptsize $\pm$18.6} & \textbf{70.9{\scriptsize $\pm$16.2}} & 39.8{\scriptsize $\pm$21.7} \\
 & FM$\downarrow$ & 49.9{\scriptsize $\pm$18.0} & 16.0{\scriptsize $\pm$5.0} & \textbf{0.0{\scriptsize $\pm$0.0}} & 30.9{\scriptsize $\pm$9.9} \\
\bottomrule
\end{tabular}
\end{table}

{\color{brown}
The CICUNSW scenario shows one of the clearest separations between naive sequential adaptation and stronger continual references. Across orderings, Naive exhibits both lower final performance and larger forgetting, while MSTE and, in many cases, Replay or Cumulative preserve substantially better performance. This makes CICUNSW particularly useful for stressing the framework under harder transfer and retention conditions. Figures~\ref{fig:app:val_cicunsw_roc_auc} and \ref{fig:app:val_cicunsw_pr_auc} provide the corresponding ordering-wise summaries.
}

\begin{figure}
    \centering
    \includegraphics[width=\linewidth]{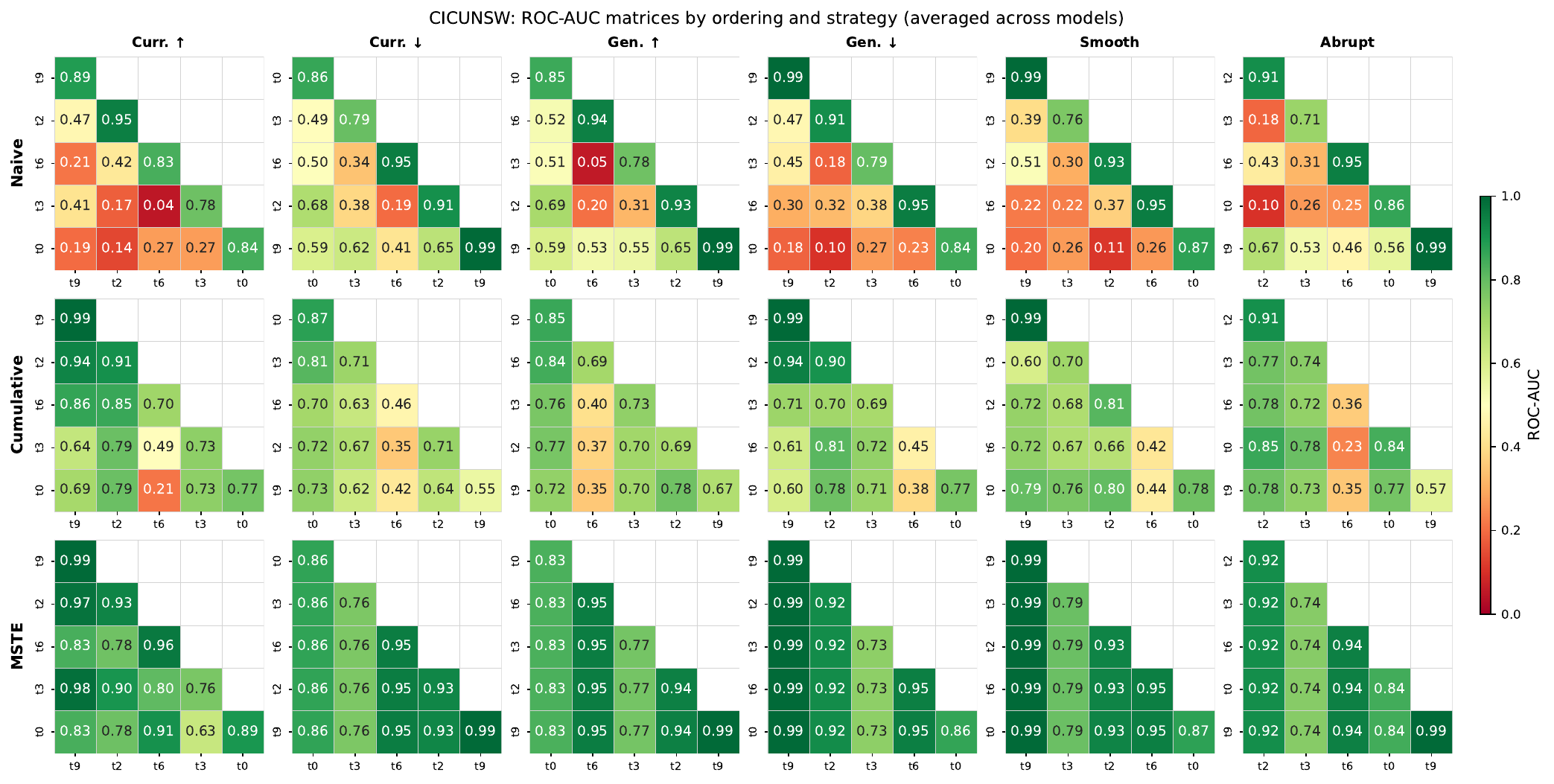}
    \caption{Ordering-specific validation results on CAD-CICUNSW measured with ROC-AUC. The figure highlights the strong performance gap between Naive and stronger reference strategies across the retained orderings.}
    \label{fig:app:val_cicunsw_roc_auc}
\end{figure}

\begin{figure}
    \centering
    \includegraphics[width=\linewidth]{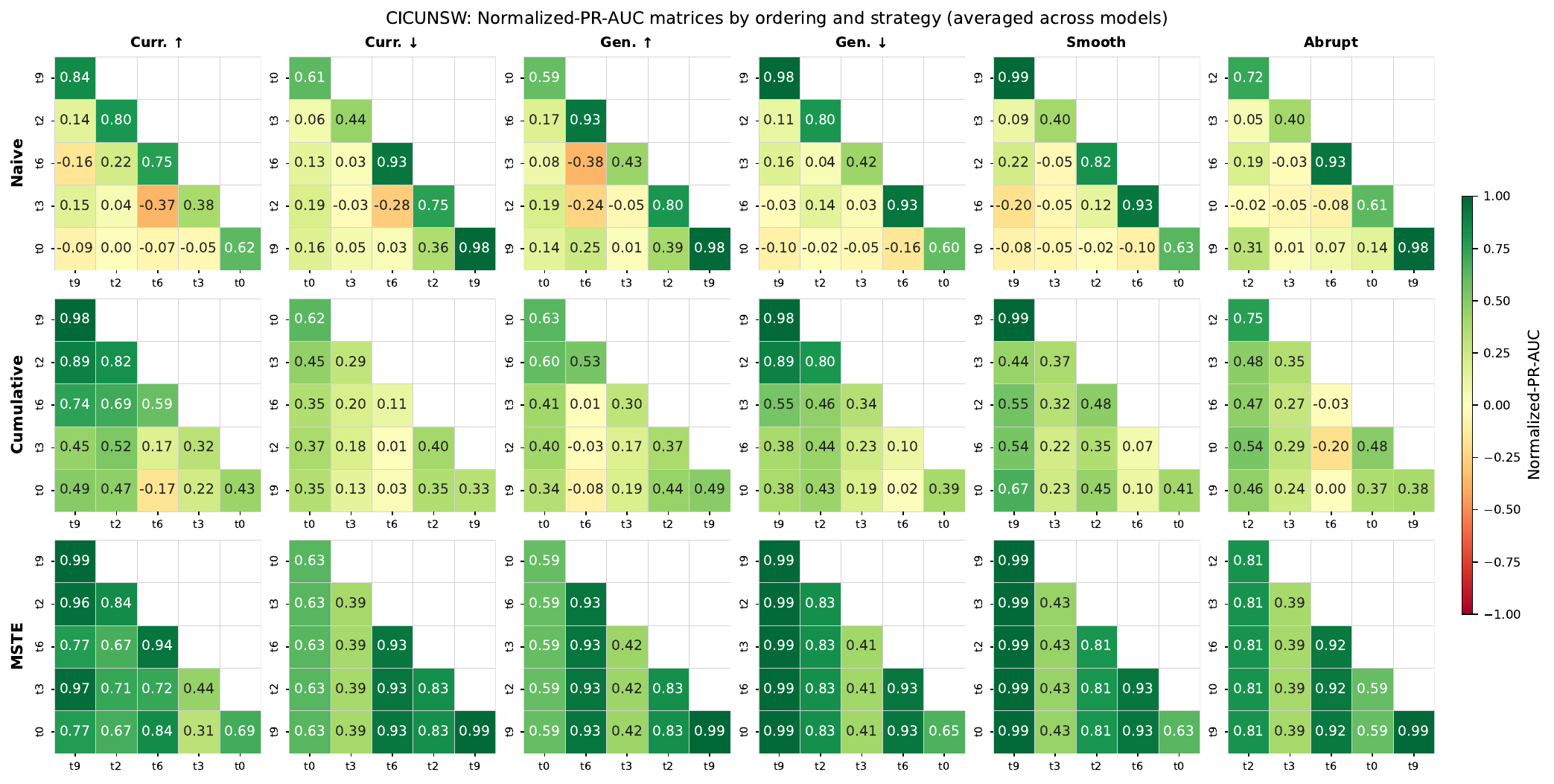}
    \caption{Ordering-specific validation results on CAD-CICUNSW measured with normalized PR-AUC. The same general pattern as in ROC-AUC is preserved, confirming that the scenario remains challenging under imbalance-aware evaluation.}
    \label{fig:app:val_cicunsw_pr_auc}
\end{figure}

    \FloatBarrier
    \subsection{MCAD-CIC-3x1}
    \begin{table}[h]
\centering
\small
\caption{ROC\text{-}AUC (mean $\pm$ std across models) per ordering on \textit{MCAD-CIC-3x1}. FM$\downarrow$: lower is better. Best strategy per ordering in \textbf{bold}.}
\label{tab:ordering_results_mcad_cic_3x1_roc_auc}
\begin{tabular}{llcccc}
\toprule
\textbf{Ordering} & \textbf{Metric} & \textbf{Naive} & \textbf{Cumulative} & \textbf{MSTE} & \textbf{Replay} \\
\midrule
\multirow{2}{*}{Curriculum (asc.)} & ROC\text{-}AUC$\uparrow$ & 64.3{\scriptsize $\pm$6.5} & 71.5{\scriptsize $\pm$9.7} & \textbf{79.1{\scriptsize $\pm$9.6}} & 69.4{\scriptsize $\pm$9.4} \\
 & FM$\downarrow$ & 17.8{\scriptsize $\pm$7.6} & 7.4{\scriptsize $\pm$4.2} & \textbf{0.0{\scriptsize $\pm$0.0}} & 12.8{\scriptsize $\pm$4.4} \\
\midrule
\multirow{2}{*}{Curriculum (desc.)} & ROC\text{-}AUC$\uparrow$ & 66.1{\scriptsize $\pm$5.3} & 68.9{\scriptsize $\pm$13.0} & \textbf{70.7{\scriptsize $\pm$13.8}} & 70.1{\scriptsize $\pm$8.3} \\
 & FM$\downarrow$ & 9.4{\scriptsize $\pm$7.5} & 5.0{\scriptsize $\pm$1.7} & \textbf{0.0{\scriptsize $\pm$0.0}} & 3.2{\scriptsize $\pm$4.3} \\
\midrule
\multirow{2}{*}{Generalization (asc.)} & ROC\text{-}AUC$\uparrow$ & 66.7{\scriptsize $\pm$5.2} & 69.4{\scriptsize $\pm$8.0} & 68.9{\scriptsize $\pm$11.0} & \textbf{74.4{\scriptsize $\pm$9.4}} \\
 & FM$\downarrow$ & 15.4{\scriptsize $\pm$9.4} & 4.7{\scriptsize $\pm$3.9} & \textbf{0.1{\scriptsize $\pm$0.2}} & 5.3{\scriptsize $\pm$2.7} \\
\midrule
\multirow{2}{*}{Generalization (desc.)} & ROC\text{-}AUC$\uparrow$ & 64.8{\scriptsize $\pm$5.3} & 68.5{\scriptsize $\pm$11.0} & \textbf{74.1{\scriptsize $\pm$16.4}} & 67.5{\scriptsize $\pm$5.1} \\
 & FM$\downarrow$ & 13.7{\scriptsize $\pm$9.6} & 4.7{\scriptsize $\pm$5.7} & \textbf{0.0{\scriptsize $\pm$0.0}} & 7.7{\scriptsize $\pm$6.3} \\
\midrule
\multirow{2}{*}{Smooth drift} & ROC\text{-}AUC$\uparrow$ & 67.4{\scriptsize $\pm$5.5} & 67.0{\scriptsize $\pm$11.0} & \textbf{72.0{\scriptsize $\pm$11.0}} & 66.4{\scriptsize $\pm$9.8} \\
 & FM$\downarrow$ & 11.6{\scriptsize $\pm$9.3} & 3.0{\scriptsize $\pm$1.1} & \textbf{0.0{\scriptsize $\pm$0.0}} & 7.2{\scriptsize $\pm$6.5} \\
\midrule
\multirow{2}{*}{Abrupt drift} & ROC\text{-}AUC$\uparrow$ & \textbf{69.5{\scriptsize $\pm$4.6}} & 64.5{\scriptsize $\pm$6.0} & 66.8{\scriptsize $\pm$12.7} & 69.0{\scriptsize $\pm$7.9} \\
 & FM$\downarrow$ & 9.9{\scriptsize $\pm$9.2} & 4.2{\scriptsize $\pm$3.9} & \textbf{0.7{\scriptsize $\pm$1.5}} & 5.4{\scriptsize $\pm$3.6} \\
\bottomrule
\end{tabular}
\end{table}

\begin{table}[h]
\centering
\small
\caption{Normalized\text{-}PR\text{-}AUC (mean $\pm$ std across models) per ordering on \textit{MCAD-CIC-3x1}. FM$\downarrow$: lower is better. Best strategy per ordering in \textbf{bold}.}
\label{tab:ordering_results_mcad_cic_3x1_normalized_pr_auc}
\begin{tabular}{llcccc}
\toprule
\textbf{Ordering} & \textbf{Metric} & \textbf{Naive} & \textbf{Cumulative} & \textbf{MSTE} & \textbf{Replay} \\
\midrule
\multirow{2}{*}{Curriculum (asc.)} & Normalized\text{-}PR\text{-}AUC$\uparrow$ & 28.4{\scriptsize $\pm$12.2} & 37.7{\scriptsize $\pm$10.1} & \textbf{41.7{\scriptsize $\pm$15.1}} & 34.9{\scriptsize $\pm$12.0} \\
 & FM$\downarrow$ & 17.7{\scriptsize $\pm$12.0} & 9.0{\scriptsize $\pm$5.5} & \textbf{0.0{\scriptsize $\pm$0.0}} & 13.7{\scriptsize $\pm$4.9} \\
\midrule
\multirow{2}{*}{Curriculum (desc.)} & Normalized\text{-}PR\text{-}AUC$\uparrow$ & 27.4{\scriptsize $\pm$4.8} & 37.3{\scriptsize $\pm$18.6} & \textbf{39.2{\scriptsize $\pm$23.6}} & 34.7{\scriptsize $\pm$12.1} \\
 & FM$\downarrow$ & 19.9{\scriptsize $\pm$16.4} & 6.8{\scriptsize $\pm$3.8} & \textbf{0.0{\scriptsize $\pm$0.0}} & 9.8{\scriptsize $\pm$3.7} \\
\midrule
\multirow{2}{*}{Generalization (asc.)} & Normalized\text{-}PR\text{-}AUC$\uparrow$ & 33.1{\scriptsize $\pm$4.3} & 35.2{\scriptsize $\pm$8.9} & 34.5{\scriptsize $\pm$14.2} & \textbf{42.3{\scriptsize $\pm$14.1}} \\
 & FM$\downarrow$ & 23.2{\scriptsize $\pm$20.9} & 7.1{\scriptsize $\pm$7.2} & \textbf{0.0{\scriptsize $\pm$0.0}} & 9.0{\scriptsize $\pm$5.3} \\
\midrule
\multirow{2}{*}{Generalization (desc.)} & Normalized\text{-}PR\text{-}AUC$\uparrow$ & 24.7{\scriptsize $\pm$8.1} & 33.3{\scriptsize $\pm$9.0} & \textbf{37.7{\scriptsize $\pm$26.1}} & 27.5{\scriptsize $\pm$2.5} \\
 & FM$\downarrow$ & 18.1{\scriptsize $\pm$20.4} & 6.6{\scriptsize $\pm$5.5} & \textbf{0.0{\scriptsize $\pm$0.0}} & 11.6{\scriptsize $\pm$8.0} \\
\midrule
\multirow{2}{*}{Smooth drift} & Normalized\text{-}PR\text{-}AUC$\uparrow$ & 28.3{\scriptsize $\pm$6.9} & 35.7{\scriptsize $\pm$14.2} & \textbf{42.6{\scriptsize $\pm$14.7}} & 27.1{\scriptsize $\pm$14.2} \\
 & FM$\downarrow$ & 21.3{\scriptsize $\pm$19.1} & 4.0{\scriptsize $\pm$3.1} & \textbf{0.0{\scriptsize $\pm$0.0}} & 15.0{\scriptsize $\pm$14.7} \\
\midrule
\multirow{2}{*}{Abrupt drift} & Normalized\text{-}PR\text{-}AUC$\uparrow$ & \textbf{37.3{\scriptsize $\pm$4.3}} & 34.9{\scriptsize $\pm$7.0} & 34.6{\scriptsize $\pm$16.6} & 36.7{\scriptsize $\pm$10.8} \\
 & FM$\downarrow$ & 17.6{\scriptsize $\pm$14.3} & 3.5{\scriptsize $\pm$1.6} & \textbf{0.6{\scriptsize $\pm$1.2}} & 8.1{\scriptsize $\pm$4.7} \\
\bottomrule
\end{tabular}
\end{table}

    \FloatBarrier
    \subsection{MCAD-CIC-3xN}
\begin{table}[h]
\centering
\small
\caption{ROC\text{-}AUC (mean $\pm$ std across models) per ordering on \textit{MCAD-CIC-3xN}. FM$\downarrow$: lower is better. Best strategy per ordering in \textbf{bold}.}
\label{tab:ordering_results_mcad_cic_3xn_roc_auc}
\begin{tabular}{llcccc}
\toprule
\textbf{Ordering} & \textbf{Metric} & \textbf{Naive} & \textbf{Cumulative} & \textbf{MSTE} & \textbf{Replay} \\
\midrule
\multirow{2}{*}{Curriculum (asc.)} & ROC\text{-}AUC$\uparrow$ & 46.1{\scriptsize $\pm$6.1} & 78.3{\scriptsize $\pm$9.4} & \textbf{92.0{\scriptsize $\pm$6.5}} & 80.7{\scriptsize $\pm$8.0} \\
 & FM$\downarrow$ & 56.8{\scriptsize $\pm$9.9} & 12.6{\scriptsize $\pm$7.5} & \textbf{0.0{\scriptsize $\pm$0.1}} & 12.2{\scriptsize $\pm$5.6} \\
\midrule
\multirow{2}{*}{Curriculum (desc.)} & ROC\text{-}AUC$\uparrow$ & 50.6{\scriptsize $\pm$1.1} & 66.5{\scriptsize $\pm$8.4} & \textbf{87.3{\scriptsize $\pm$6.9}} & 70.7{\scriptsize $\pm$13.0} \\
 & FM$\downarrow$ & 38.9{\scriptsize $\pm$5.6} & 16.1{\scriptsize $\pm$3.9} & \textbf{0.5{\scriptsize $\pm$1.0}} & 16.9{\scriptsize $\pm$6.3} \\
\midrule
\multirow{2}{*}{Generalization (asc.)} & ROC\text{-}AUC$\uparrow$ & 53.4{\scriptsize $\pm$1.4} & 70.7{\scriptsize $\pm$6.6} & \textbf{88.9{\scriptsize $\pm$5.2}} & 78.2{\scriptsize $\pm$7.7} \\
 & FM$\downarrow$ & 41.9{\scriptsize $\pm$6.7} & 12.7{\scriptsize $\pm$4.1} & \textbf{0.3{\scriptsize $\pm$0.7}} & 12.3{\scriptsize $\pm$3.6} \\
\midrule
\multirow{2}{*}{Generalization (desc.)} & ROC\text{-}AUC$\uparrow$ & 42.9{\scriptsize $\pm$5.9} & 71.7{\scriptsize $\pm$12.9} & \textbf{88.9{\scriptsize $\pm$8.0}} & 75.9{\scriptsize $\pm$12.0} \\
 & FM$\downarrow$ & 54.1{\scriptsize $\pm$9.9} & 18.4{\scriptsize $\pm$8.7} & \textbf{0.2{\scriptsize $\pm$0.3}} & 16.2{\scriptsize $\pm$8.0} \\
\midrule
\multirow{2}{*}{Smooth drift} & ROC\text{-}AUC$\uparrow$ & 46.7{\scriptsize $\pm$3.4} & 68.4{\scriptsize $\pm$11.0} & \textbf{86.0{\scriptsize $\pm$9.1}} & 74.8{\scriptsize $\pm$13.3} \\
 & FM$\downarrow$ & 43.3{\scriptsize $\pm$7.1} & 17.4{\scriptsize $\pm$7.2} & \textbf{1.1{\scriptsize $\pm$2.1}} & 14.2{\scriptsize $\pm$7.6} \\
\midrule
\multirow{2}{*}{Abrupt drift} & ROC\text{-}AUC$\uparrow$ & 45.2{\scriptsize $\pm$5.0} & 73.7{\scriptsize $\pm$11.1} & \textbf{90.5{\scriptsize $\pm$7.7}} & 79.9{\scriptsize $\pm$8.9} \\
 & FM$\downarrow$ & 49.4{\scriptsize $\pm$10.6} & 13.1{\scriptsize $\pm$5.4} & \textbf{0.2{\scriptsize $\pm$0.3}} & 12.4{\scriptsize $\pm$5.0} \\
\bottomrule
\end{tabular}
\end{table}

\begin{table}[h]
\centering
\small
\caption{Normalized\text{-}PR\text{-}AUC (mean $\pm$ std across models) per ordering on \textit{MCAD-CIC-3xN}. FM$\downarrow$: lower is better. Best strategy per ordering in \textbf{bold}.}
\label{tab:ordering_results_mcad_cic_3xn_normalized_pr_auc}
\begin{tabular}{llcccc}
\toprule
\textbf{Ordering} & \textbf{Metric} & \textbf{Naive} & \textbf{Cumulative} & \textbf{MSTE} & \textbf{Replay} \\
\midrule
\multirow{2}{*}{Curriculum (asc.)} & Normalized\text{-}PR\text{-}AUC$\uparrow$ & 19.4{\scriptsize $\pm$4.8} & 61.1{\scriptsize $\pm$13.1} & \textbf{83.4{\scriptsize $\pm$10.8}} & 64.3{\scriptsize $\pm$10.7} \\
 & FM$\downarrow$ & 78.2{\scriptsize $\pm$9.6} & 18.0{\scriptsize $\pm$9.8} & \textbf{0.0{\scriptsize $\pm$0.1}} & 18.5{\scriptsize $\pm$6.7} \\
\midrule
\multirow{2}{*}{Curriculum (desc.)} & Normalized\text{-}PR\text{-}AUC$\uparrow$ & 15.6{\scriptsize $\pm$3.8} & 31.9{\scriptsize $\pm$8.1} & \textbf{67.8{\scriptsize $\pm$14.5}} & 36.1{\scriptsize $\pm$14.9} \\
 & FM$\downarrow$ & 54.2{\scriptsize $\pm$8.2} & 24.2{\scriptsize $\pm$3.2} & \textbf{0.8{\scriptsize $\pm$1.5}} & 27.9{\scriptsize $\pm$6.7} \\
\midrule
\multirow{2}{*}{Generalization (asc.)} & Normalized\text{-}PR\text{-}AUC$\uparrow$ & 19.2{\scriptsize $\pm$6.6} & 45.5{\scriptsize $\pm$8.2} & \textbf{73.7{\scriptsize $\pm$11.2}} & 54.7{\scriptsize $\pm$10.0} \\
 & FM$\downarrow$ & 61.9{\scriptsize $\pm$10.3} & 18.8{\scriptsize $\pm$4.0} & \textbf{0.4{\scriptsize $\pm$0.9}} & 22.6{\scriptsize $\pm$4.5} \\
\midrule
\multirow{2}{*}{Generalization (desc.)} & Normalized\text{-}PR\text{-}AUC$\uparrow$ & 14.7{\scriptsize $\pm$7.0} & 46.8{\scriptsize $\pm$16.9} & \textbf{73.1{\scriptsize $\pm$13.4}} & 51.1{\scriptsize $\pm$14.7} \\
 & FM$\downarrow$ & 70.5{\scriptsize $\pm$8.2} & 24.6{\scriptsize $\pm$11.6} & \textbf{0.0{\scriptsize $\pm$0.0}} & 25.9{\scriptsize $\pm$5.5} \\
\midrule
\multirow{2}{*}{Smooth drift} & Normalized\text{-}PR\text{-}AUC$\uparrow$ & 16.1{\scriptsize $\pm$3.2} & 39.3{\scriptsize $\pm$13.1} & \textbf{67.4{\scriptsize $\pm$15.4}} & 48.7{\scriptsize $\pm$15.7} \\
 & FM$\downarrow$ & 58.1{\scriptsize $\pm$6.9} & 24.3{\scriptsize $\pm$7.5} & \textbf{0.9{\scriptsize $\pm$1.7}} & 23.0{\scriptsize $\pm$6.4} \\
\midrule
\multirow{2}{*}{Abrupt drift} & Normalized\text{-}PR\text{-}AUC$\uparrow$ & 16.5{\scriptsize $\pm$5.5} & 52.6{\scriptsize $\pm$14.8} & \textbf{78.2{\scriptsize $\pm$12.3}} & 60.8{\scriptsize $\pm$11.2} \\
 & FM$\downarrow$ & 67.1{\scriptsize $\pm$13.3} & 17.8{\scriptsize $\pm$5.5} & \textbf{0.0{\scriptsize $\pm$0.1}} & 19.0{\scriptsize $\pm$4.6} \\
\bottomrule
\end{tabular}
\end{table}

{\color{brown}
The multi-dataset CAD-CIC-3xN scenario is substantially harder than the single-dataset cases. The ordering-wise results show that Naive degrades sharply and that even stronger continual baselines must operate under more severe distributional heterogeneity. This is consistent with the intended role of multi-dataset scenarios as a stress test for continual anomaly detection. Figures~\ref{fig:app:val_3xn_roc_auc} and \ref{fig:app:val_3xn_pr_auc} illustrate this effect.
}

\begin{figure}
    \centering
    \includegraphics[width=\linewidth]{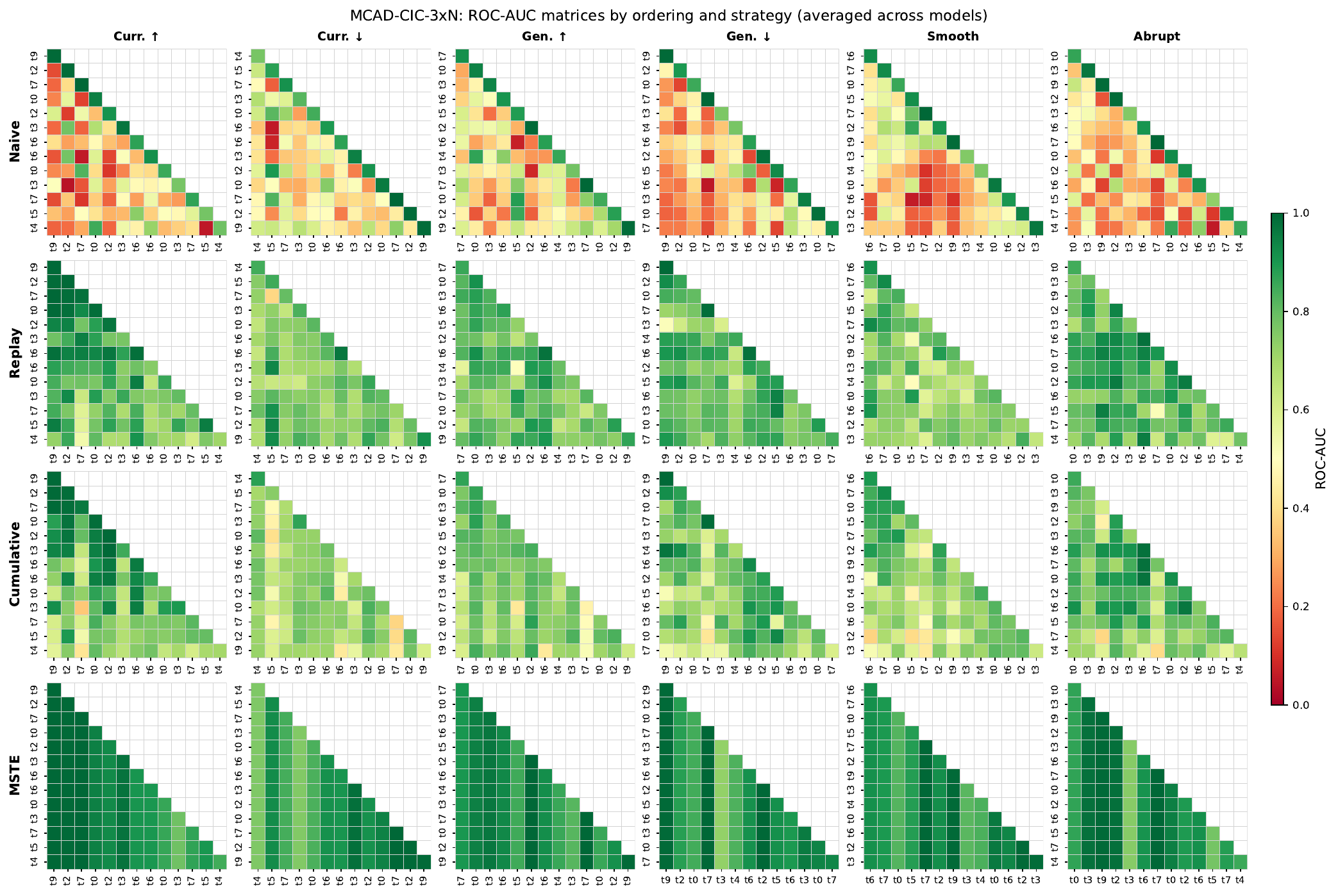}
    \caption{Ordering-specific validation results on CAD-CIC-3xN measured with ROC-AUC. Compared with the single-dataset scenarios, the multi-dataset setting induces substantially harsher continual-learning conditions.}
    \label{fig:app:val_3xn_roc_auc}
\end{figure}

\begin{figure}
    \centering
    \includegraphics[width=\linewidth]{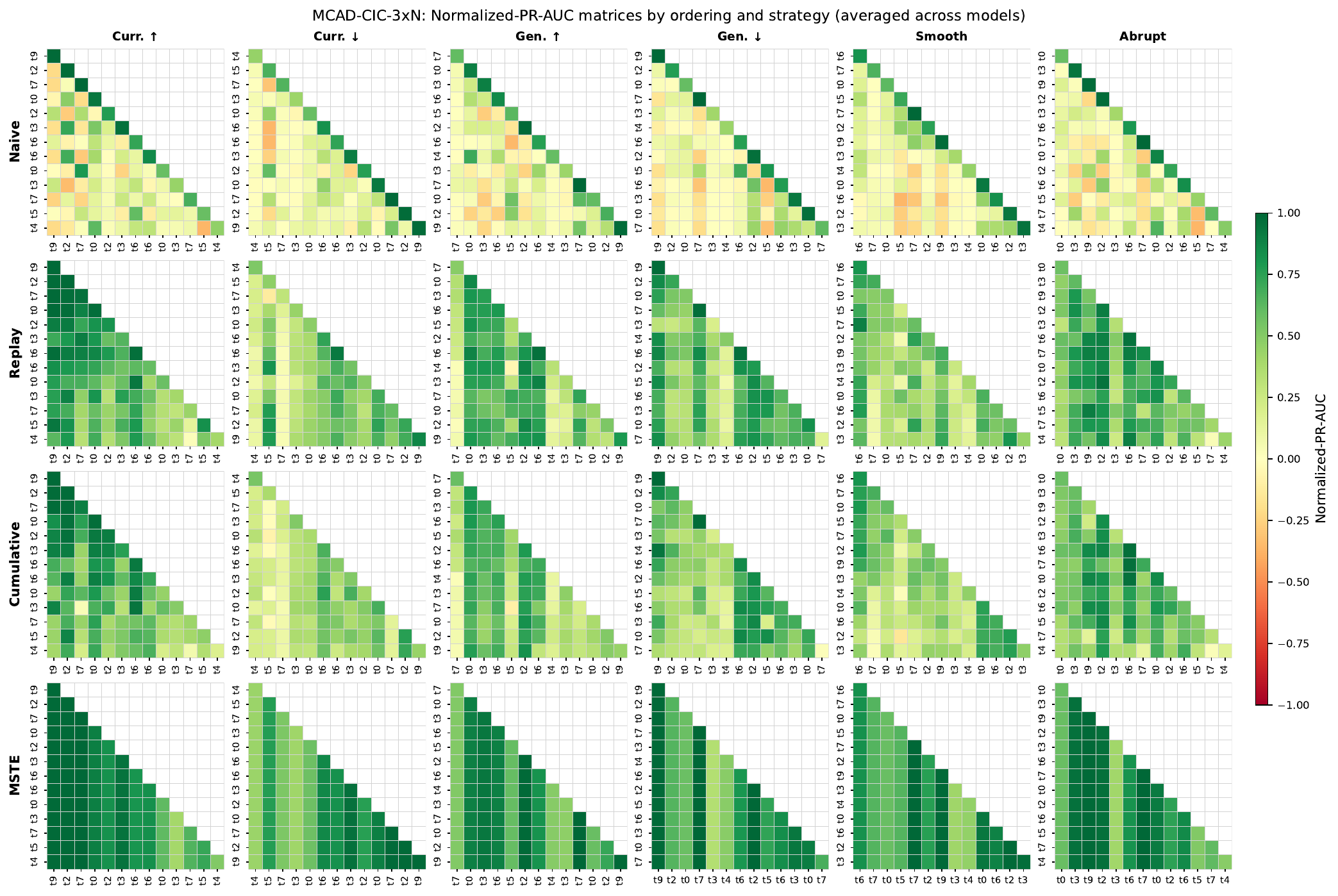}
    \caption{Ordering-specific validation results on CAD-CIC-3xN measured with normalized PR-AUC. The stronger degradation of Naive persists under imbalance-aware evaluation, reinforcing the difficulty of the multi-dataset setting.}
    \label{fig:app:val_3xn_pr_auc}
\end{figure}

\FloatBarrier
\section{Additional results discussion}
{\color{brown}
\subsection{Why can Cumulative underperform?}
\label{appendix:sec:cumulative-imbalance}

    The CICIDS2017 scenario provides a concrete example of why full access to past data does not necessarily make Cumulative the strongest continual reference. The cumulative strategy retrains on the union of all data observed so far. While this removes forgetting in principle, it also means that the effective optimization objective is heavily shaped by the empirical distribution of the retained tasks. If that distribution is strongly imbalanced, the model can become disproportionately tuned to a small number of dominant concepts, at the expense of smaller ones.
    
    The per-concept statistics (see Table~\ref{tab:ds_stats_CAD-CICIDS2017}) show that this imbalance is substantial in the final CICIDS2017 scenario. In particular, concept \texttt{cicids2017\_1} contains 1,434,082 training samples and 411,738 test samples, whereas the smallest concepts contain only 1,713 or 8,392 training samples. Aggregated over the final task set, \texttt{cicids2017\_1} alone contributes roughly 88\% of the training data and about 81\% of the test data. As a result, cumulative retraining without explicit task balancing or reweighting is dominated by the largest concept, while the optimization signal associated with smaller concepts becomes comparatively weak.
    Similar observations can be drawn for other datasets, as shown in Tables~\ref{tab:ds_stats_CAD-CICIDS2018} and \ref{tab:ds_stats_CAD-CICUNSW}.

    This matters because our evaluation is scenario-oriented rather than distribution-oriented: we care about performance across all tasks, not only about performance on the dominant regime. A detector that improves strongly on the largest concept can still yield disappointing average task-level performance if it underfits smaller concepts or fails to preserve their specific decision boundaries. In this sense, more data is not automatically better when that data is concentrated in a single regime. The additional samples can bias the learned representation toward the dominant concept instead of improving balanced generalization across the scenario.
    
    This observation also clarifies why MSTE and Replay can behave more favorably. MSTE is unaffected by cross-task imbalance because each expert is trained independently on a single task. Replay, while still imperfect, partially limits domination by the largest concept because the replay buffer caps the amount of data that any one task can contribute during training. Cumulative, by contrast, inherits the raw imbalance of the scenario directly.
    
    The broader implication is methodological. In CAD, cumulative should not always be interpreted as a clean upper bound solely because it has access to all past data. Its behavior also depends on how the scenario distributes samples across tasks. This is precisely why explicit scenario characterization is necessary: baseline performance is shaped not only by the continual strategy itself, but also by structural properties such as task size imbalance.

    }

\FloatBarrier
\section{Per-Dataset Scenario Construction Reports}
\label{app:dataset_progression}

This appendix provides a dataset-level trace of how each final benchmark scenario was constructed using our framework. For each dataset, we report the intermediate outputs produced at successive stages of the pipeline, including the generation of candidate tasks, the results of the filtering criteria, and the scenario task evaluation (STE) scores obtained across different candidate task configurations. These reports complement the aggregate description in the main paper by making the scenario-construction process transparent, auditable, and reproducible for each dataset individually.

For clarity, task identifiers used during framework execution may differ from the human-readable identifiers reported in the final scenario tables, as we rename them to keep the consistent incremental task ids ($0, 1, 2)$

    \subsection{CICIDS2017}
    \begin{table}[h]
\centering
\small
\setlength{\tabcolsep}{4pt}
\caption{Scenario CAD-CICIDS2017: Task filtering results per task discovery configuration. BC~=~BothClasses, CA~=~Closest Anomalies, RA~=~Random Anomalies, GM~=~Gaussian Mixture, KM~=~K-Means, SC~=~Spectral Clustering.By Day~=~Split by day/part of the day}
\label{tab:filtering_results_CAD-CICIDS2017}
\begin{tabular}{lclllllccl}
\toprule
\textbf{Config} & \textbf{IT.} & \textbf{\makecell{FC1}} & \textbf{\makecell{FC2}} & \textbf{\makecell{FC3}} & \textbf{\makecell{FC4}} & \textbf{\makecell{FC5}} & \textbf{FT} & \textbf{MT} & \textbf{Final tasks.} \\
\midrule
By Day & 7 & t1 & t1, t2, t3, t4 & t1, t3, t4 & t0, t1, t2, t3, t4, t5 & t0, t2, t3, t4 & 1 & \xmark & --- \\
BC / GM & 7 & --- & t4 & --- & --- & --- & 6 & \cmark & t0, t1, t2, t3, t5, t6 \\
BC / KM & 7 & --- & t1, t5 & t1, t5 & --- & --- & 5 & \cmark & t0, t2, t3, t4, t6 \\
BC / SC & 10 & --- & t1, t4, t6, t7 & t0, t7 & t1, t6, t7 & --- & 5 & \cmark & t2, t3, t5, t8, t9 \\
CA / GM & 8 & t2, t6 & t0, t2, t5, t6 & t5 & --- & --- & 4 & \cmark & t1, t3, t4, t7 \\
CA / KM & 8 & --- & t0, t2, t4 & t4 & t0, t4 & --- & 5 & \cmark & t1, t3, t5, t6, t7 \\
CA / SC & 10 & t2 & t0, t1, t2, t3, t8 & t0, t1, t2, t3, t6 & t1, t6 & --- & 4 & \cmark & t4, t5, t7, t9 \\
RA / GM & 9 & --- & --- & t4 & t7 & --- & 7 & \cmark & t0, t1, t2, t3, t5, t6, t8 \\
RA / KM & 8 & --- & --- & t5 & t4 & t4 & 6 & \cmark & t0, t1, t2, t3, t6, t7 \\
RA / SC & 10 & --- & t1 & t6 & t1, t4, t5, t6, t8, t9 & t1, t5, t6 & 4 & \cmark & t0, t2, t3, t7 \\
\bottomrule
\end{tabular}
\end{table}
{\color{brown}
The CICIDS2017 construction report shows that the final split is not the outcome of a single heuristic, but of sequential filtering based on learnability, transfer coverage, and redundancy. The filtering table identifies which candidate task configurations are rejected and why, while Figures~\ref{fig:app:cicids2017_ste_roc_auc} and \ref{fig:app:cicids2017_ste_pr_auc} visualize the corresponding STE transfer structure. The retained split exhibits the intended pattern of strong self-performance together with non-trivial off-diagonal variation, indicating that tasks are both learnable and sufficiently heterogeneous.
}
        
        \begin{figure}[h]
            \centering
            \includegraphics[width=\linewidth]{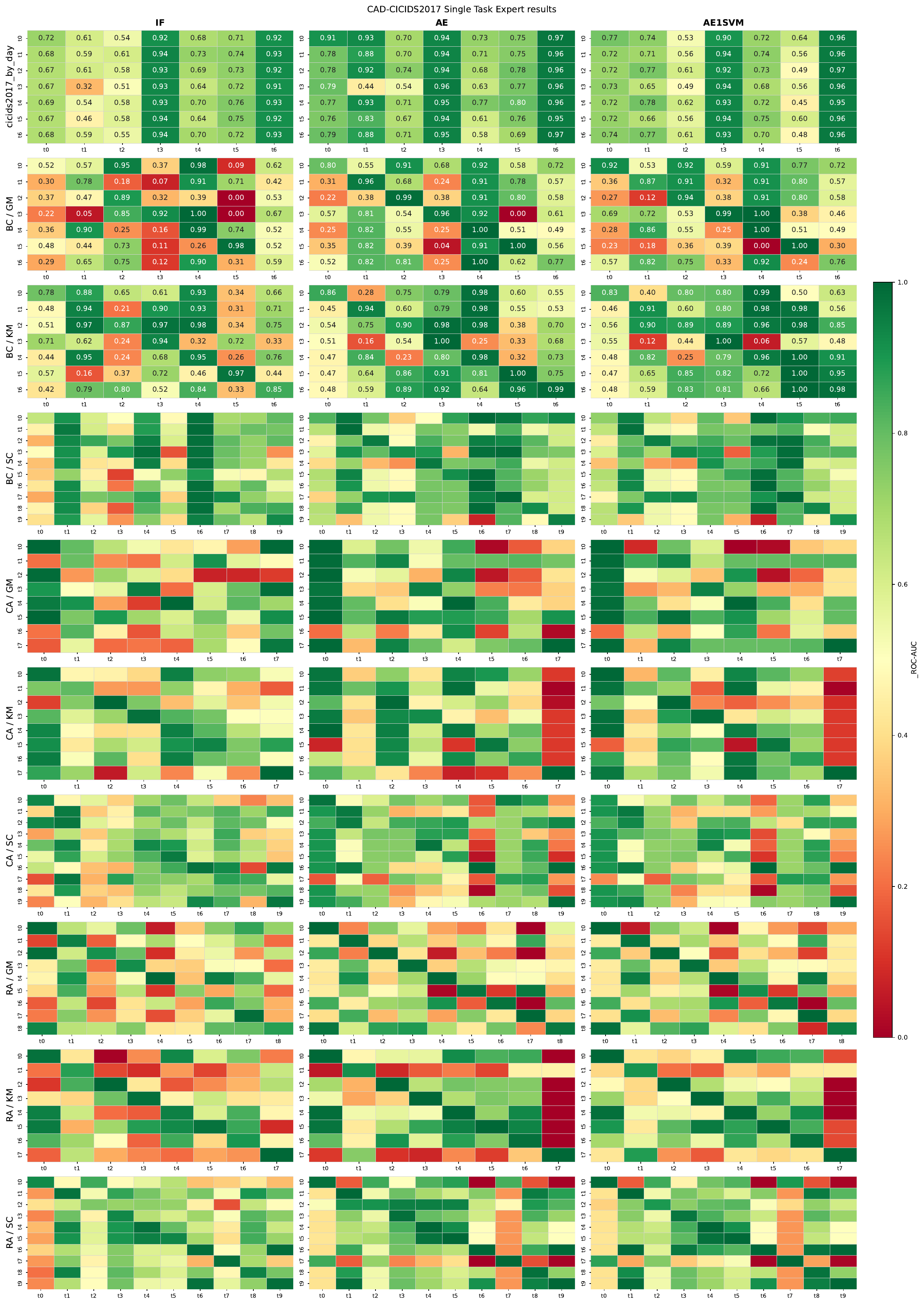}
            \caption{STE evaluation heatmaps for the CICIDS2017 candidate splits measured with ROC-AUC. Strong diagonal values indicate self-learnability, while the off-diagonal structure reveals transfer, dominance, and redundancy patterns used during filtering.}
            \label{fig:app:cicids2017_ste_roc_auc}
        \end{figure}
        
        \begin{figure}[h]
            \centering
            \includegraphics[width=\linewidth]{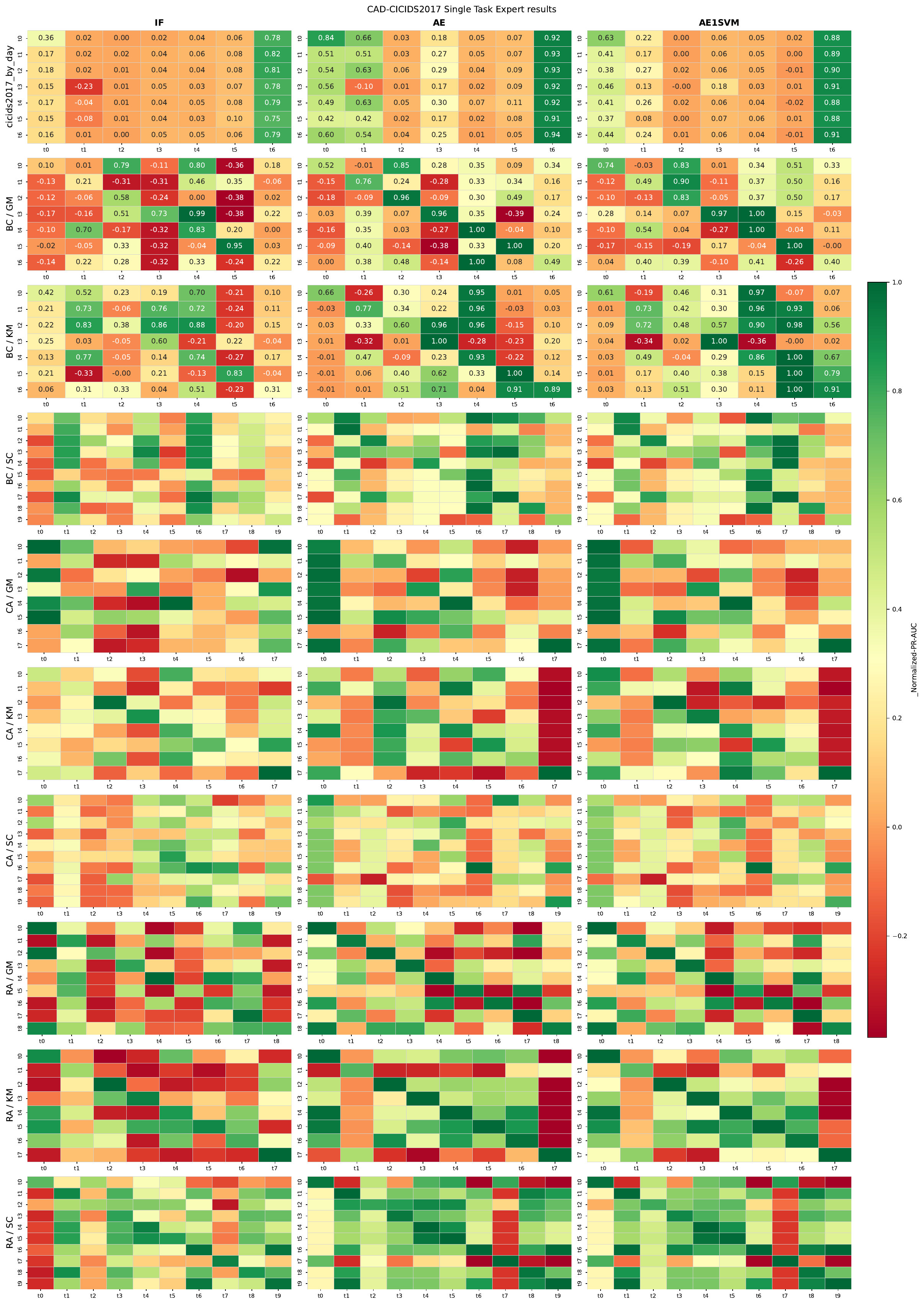}
            \caption{STE evaluation heatmaps for the CICIDS2017 candidate splits measured with normalized PR-AUC. The same split-level structure remains visible under an imbalance-aware metric, supporting the robustness of the filtering decisions.}
            \label{fig:app:cicids2017_ste_pr_auc}
        \end{figure}
    \FloatBarrier
    
    \subsection{CICIDS2018}
    \begin{table}[h]
\centering
\small
\setlength{\tabcolsep}{4pt}
\caption{Scenario CAD-CICIDS2018: Task filtering results per task discovery configuration. BC~=~BothClasses, CA~=~Closest Anomalies, RA~=~Random Anomalies, GM~=~Gaussian Mixture, KM~=~K-Means, SC~=~Spectral Clustering.By Day~=~Split by day/part of the day}
\label{tab:filtering_results_CAD-CICIDS2018}
\begin{tabular}{lclllllccl}
\toprule
\textbf{Config} & \textbf{IT.} & \textbf{\makecell{FC1}} & \textbf{\makecell{FC2}} & \textbf{\makecell{FC3}} & \textbf{\makecell{FC4}} & \textbf{\makecell{FC5}} & \textbf{FT} & \textbf{MT} & \textbf{Final tasks.} \\
\midrule
By Day & 10 & t4, t9 & t2, t4, t5, t6, t9 & t0, t2, t4, t5, t6 & t0, t2, t4, t5, t6, t7, t9 & t2, t4, t5 & 3 & \cmark & t1, t3, t8 \\
BC / GM & 9 & t6 & t8 & t6 & --- & --- & 7 & \cmark & t0, t1, t2, t3, t4, t5, t7 \\
BC / KM & 9 & t3, t6 & t2, t3, t5, t6 & t2, t3 & t3 & t3 & 5 & \cmark & t0, t1, t4, t7, t8 \\
BC / SC & 8 & t2 & t2 & t2, t3 & --- & --- & 6 & \cmark & t0, t1, t4, t5, t6, t7 \\
CA / GM & 9 & t0 & t2, t3, t8 & t2, t3 & --- & --- & 5 & \cmark & t1, t4, t5, t6, t7 \\
CA / KM & 9 & t3 & t2, t3, t4, t7 & t3, t4 & t3 & t3, t7 & 5 & \cmark & t0, t1, t5, t6, t8 \\
CA / SC & 10 & t0, t1, t7 & t0, t1, t9 & t0, t4, t9 & t4, t9 & t1 & 5 & \cmark & t2, t3, t5, t6, t8 \\
RA / GM & 9 & --- & t3 & t2 & --- & t3 & 7 & \cmark & t0, t1, t4, t5, t6, t7, t8 \\
RA / KM & 9 & --- & t3 & t3 & --- & t3 & 8 & \cmark & t0, t1, t2, t4, t5, t6, t7, t8 \\
RA / SC & 10 & --- & t4 & t4, t8 & t4, t8 & t4, t8 & 8 & \cmark & t0, t1, t2, t3, t5, t6, t7, t9 \\
\bottomrule
\end{tabular}
\end{table}
{\color{brown}
For CICIDS2018, the STE reports again show that not all candidate decompositions are equally suitable for CAD. Some splits exhibit stronger cross-task coverage or weaker self-learnability, whereas the selected configuration preserves a better balance between task solvability and heterogeneity. Figures~\ref{fig:app:cicids2018_ste_roc_auc} and \ref{fig:app:cicids2018_ste_pr_auc} therefore provide direct evidence that the final scenario is grounded in observed task behavior rather than in arbitrary partitioning.
}
        
        \begin{figure}[h]
            \centering
            \includegraphics[width=\linewidth]{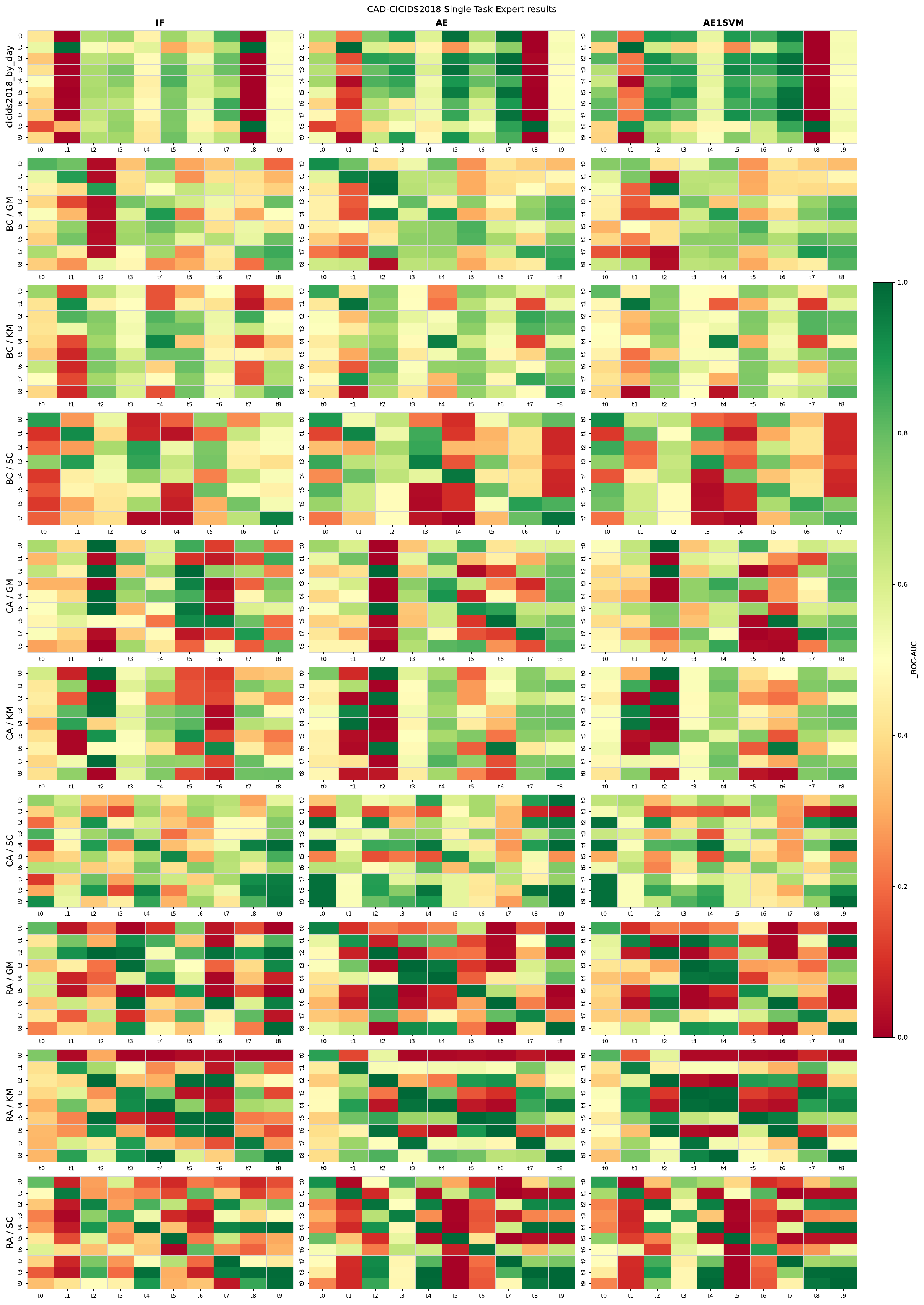}
            \caption{STE evaluation heatmaps for the CICIDS2018 candidate splits measured with ROC-AUC. The selected split retains clear diagonal dominance without collapsing into uniformly high transfer across tasks.}
            \label{fig:app:cicids2018_ste_roc_auc}
        \end{figure}
        
        \begin{figure}[h]
            \centering
            \includegraphics[width=\linewidth]{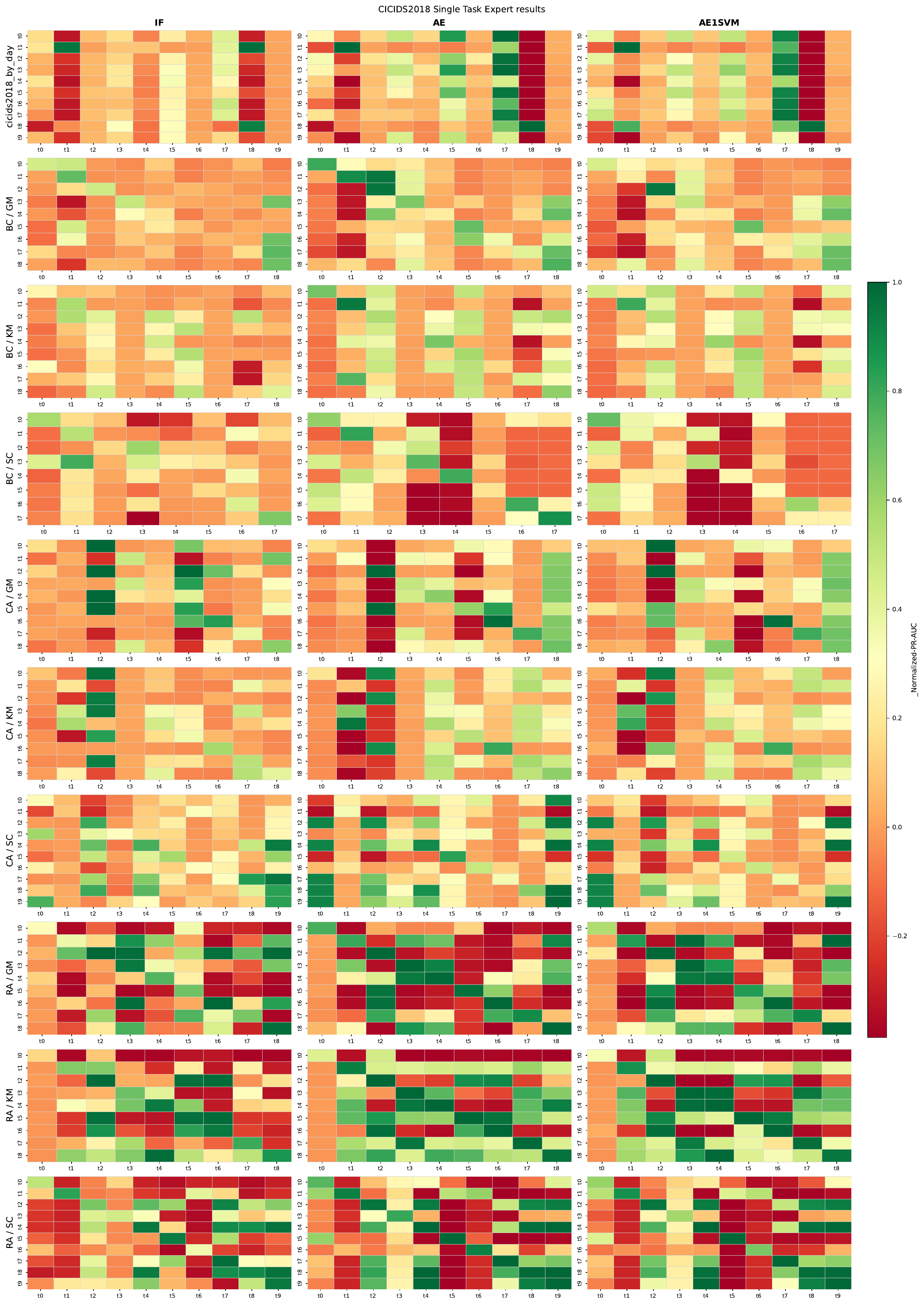}
            \caption{STE evaluation heatmaps for the CICIDS2018 candidate splits measured with normalized PR-AUC. The retained split remains differentiated under imbalance-aware evaluation, which supports the consistency of the filtering stage.}
            \label{fig:app:cicids2018_ste_pr_auc}
        \end{figure}
    \FloatBarrier
    
    \subsection{CICUNSW}
    \begin{table}[h]
\centering
\small
\setlength{\tabcolsep}{4pt}
\caption{Scenario CAD-CICUNSW: Task filtering results per task discovery configuration. BC~=~BothClasses, CA~=~Closest Anomalies, RA~=~Random Anomalies, GM~=~Gaussian Mixture, KM~=~K-Means, SC~=~Spectral Clustering.By Day~=~Split by day/part of the day}
\label{tab:filtering_results_CAD-CICUNSW}
\begin{tabular}{lclllllccl}
\toprule
\textbf{Config} & \textbf{IT.} & \textbf{\makecell{FC1}} & \textbf{\makecell{FC2}} & \textbf{\makecell{FC3}} & \textbf{\makecell{FC4}} & \textbf{\makecell{FC5}} & \textbf{FT} & \textbf{MT} & \textbf{Final tasks.} \\
\midrule
By Day & 3 & --- & --- & --- & t0, t1 & t0, t1 & 1 & \xmark & --- \\
BC / GM & 4 & t0, t1, t2 & t1 & --- & t1 & t1 & 1 & \xmark & --- \\
BC / SC & 8 & t7 & t1, t4 & t1 & t1, t4 & t4 & 5 & \cmark & t0, t2, t3, t5, t6 \\
CA / GM & 9 & t6 & t2, t5, t6 & t2, t6 & t6 & --- & 6 & \cmark & t0, t1, t3, t4, t7, t8 \\
CA / KM & 5 & --- & --- & --- & t3 & --- & 4 & \cmark & t0, t1, t2, t4 \\
CA / SC & 10 & t6 & t2, t4, t5, t6, t7, t8 & t2, t4, t5, t6, t7 & t5 & t5 & 4 & \cmark & t0, t1, t3, t9 \\
RA / GM & 9 & --- & t5, t7 & t5 & t7 & --- & 7 & \cmark & t0, t1, t2, t3, t4, t6, t8 \\
RA / KM & 5 & --- & --- & --- & t3 & t3 & 4 & \cmark & t0, t1, t2, t4 \\
RA / SC & 10 & --- & t4, t5, t7, t8 & t1, t4 & t4, t5 & t4, t5, t7 & 5 & \cmark & t0, t2, t3, t6, t9 \\
\bottomrule
\end{tabular}
\end{table}
{\color{brown}
The CICUNSW construction report highlights a more demanding transfer structure. Compared with the CICIDS scenarios, several candidate tasks are more weakly connected, which increases the importance of excluding dead or overly dominant tasks before ordering. Figures~\ref{fig:app:cicunsw_ste_roc_auc} and \ref{fig:app:cicunsw_ste_pr_auc} show that the retained split preserves learnability while still exhibiting the heterogeneity needed for meaningful continual evaluation.
}
            
        \begin{figure}[h]
            \centering
            \includegraphics[width=\linewidth]{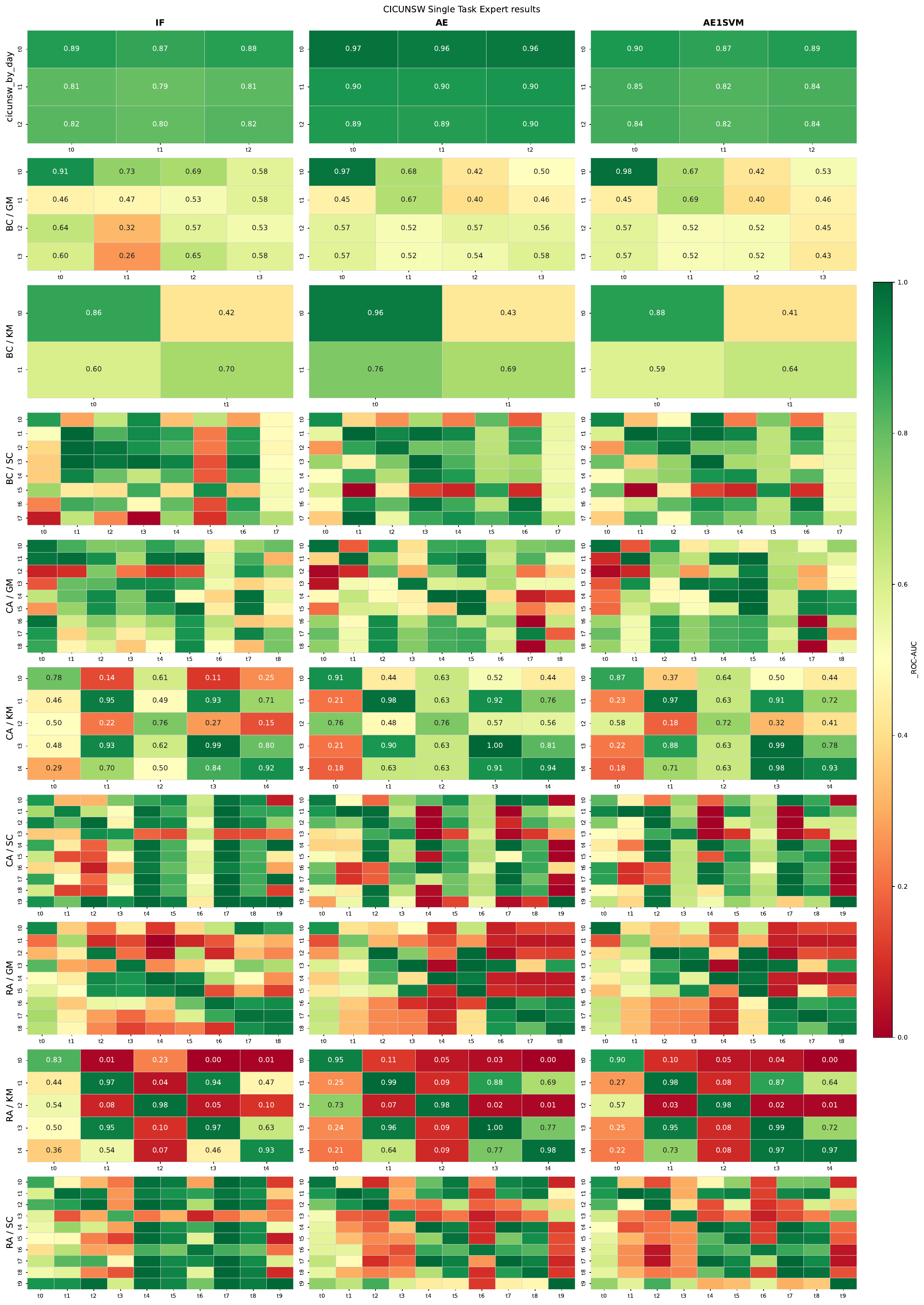}
            \caption{STE evaluation heatmaps for the CICUNSW candidate splits measured with ROC-AUC. The retained split combines diagonal learnability with sufficiently irregular off-diagonal structure to support non-trivial CAD dynamics.}
            \label{fig:app:cicunsw_ste_roc_auc}
        \end{figure}
        
        \begin{figure}[h]
            \centering
            \includegraphics[width=\linewidth]{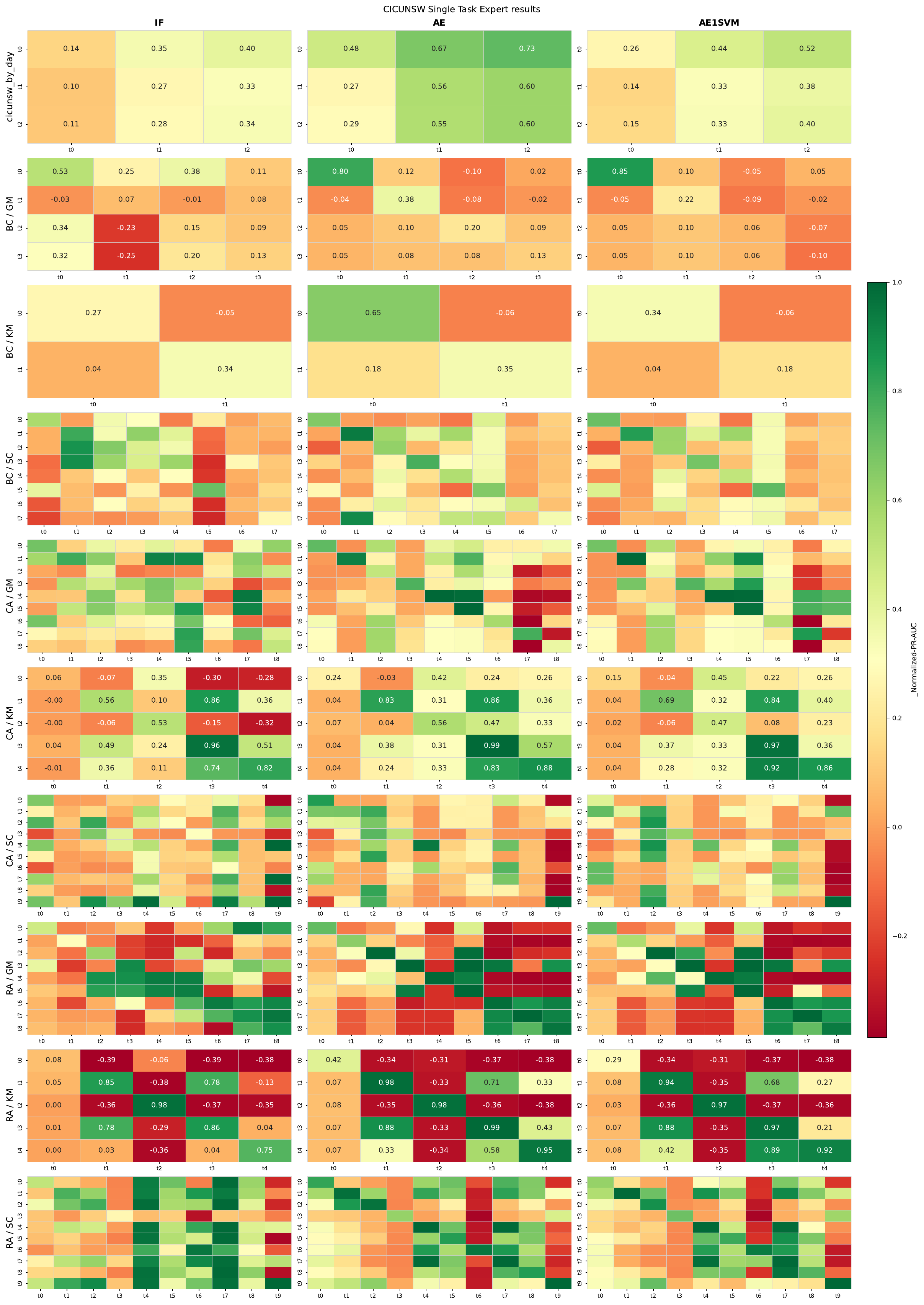}
            \caption{STE evaluation heatmaps for the CICUNSW candidate splits measured with normalized PR-AUC. The qualitative structure is preserved, indicating that the filtering decisions are not an artifact of a single metric.}
            \label{fig:app:cicunsw_ste_pr_auc}
        \end{figure}
    
    \FloatBarrier
    \subsection{MCAD-CIC-3x1}
            \begin{table}[h]
\centering
\small
\setlength{\tabcolsep}{4pt}
\caption{Scenario MCAD-CIC-3x1: Task filtering results}
\label{tab:filtering_results_MCAD-CIC-3x1}
\begin{tabular}{lcccl}
\toprule
\textbf{Config} & \textbf{IT.} & \textbf{FT} & \textbf{MT} & \textbf{Final tasks.} \\
\midrule
Datasets & 3 & 3 & \cmark & t0, t1, t2 \\
\bottomrule
\end{tabular}
\end{table}
{\color{brown}
The MCAD-CIC-3x1 setting provides a simpler multi-dataset construction report in which each source dataset acts as a single task. Even in this reduced setting, the STE structure is useful for verifying that the three resulting tasks are not trivially interchangeable. Figure~\ref{fig:app:3x1_ste_roc_auc} summarizes these cross-dataset relations.
}
        \begin{figure}[h]
            \centering
            \includegraphics[width=\linewidth]{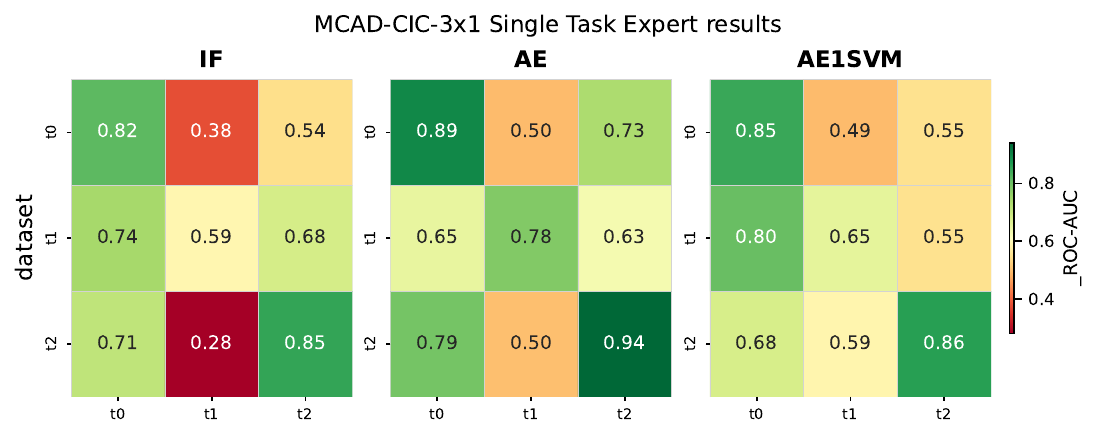}
            \caption{STE evaluation heatmaps for the MCAD-CIC-3x1 scenario measured with ROC-AUC. With one task per dataset, the figure highlights the cross-dataset transfer structure that motivates the multi-dataset continual setting.}
            \label{fig:app:3x1_ste_roc_auc}
        \end{figure}
    
    \FloatBarrier
    \subsection{MCAD-CIC-3xN}

        \begin{table}[h]
\centering
\small
\setlength{\tabcolsep}{4pt}
\caption{Scenario MCAD-CIC-3xN: Task filtering results}
\label{tab:filtering_results_MCAD-CIC-3xN}
\begin{tabular}{lcllccl}
\toprule
\textbf{Config} & \textbf{IT.} & \textbf{\makecell{FC2}} & \textbf{\makecell{FC3}} & \textbf{FT} & \textbf{MT} & \textbf{Final tasks.} \\
\midrule
Tasks from datasets & 16 & t3, t12 & t2 & 13 & \cmark & t0, t1, t4, t5, t6, t7, t8, t9, t10, t11, t13, t14, t15 \\
\bottomrule
\end{tabular}
\end{table}
{\color{brown}
The MCAD-CIC-3xN report illustrates the most complex task-construction case in the paper. Because this scenario mixes multiple datasets and multiple retained concepts per dataset, the STE heatmaps are particularly informative for showing where transfer is preserved and where regime shifts become severe. Figures~\ref{fig:app:3xn_ste_roc_auc} and \ref{fig:app:3xn_ste_pr_auc} confirm that the resulting split is far from trivial and therefore suitable as a harder benchmark extension.
The retained indices \(t0,t1,t4,\ldots,t15\) are remapped in Table~16 to dataset-prefixed identifiers \(c17_i\), \(c18_i\), and \(cu_i\) for readability.
}

        \begin{figure}[h]
            \centering
            \includegraphics[width=\linewidth]{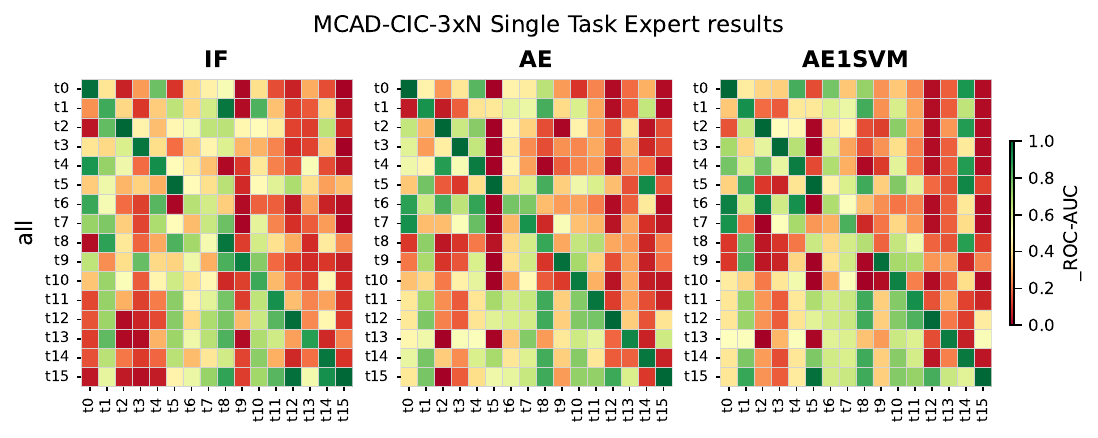}
            \caption{STE evaluation heatmaps for the MCAD-CIC-3xN scenario measured with ROC-AUC. The figure reveals a richer and harsher transfer structure than in the single-dataset scenarios, consistent with the increased difficulty of the multi-dataset setting.}
            \label{fig:app:3xn_ste_roc_auc}
        \end{figure}
        
        \begin{figure}[h]
            \centering
            \includegraphics[width=\linewidth]{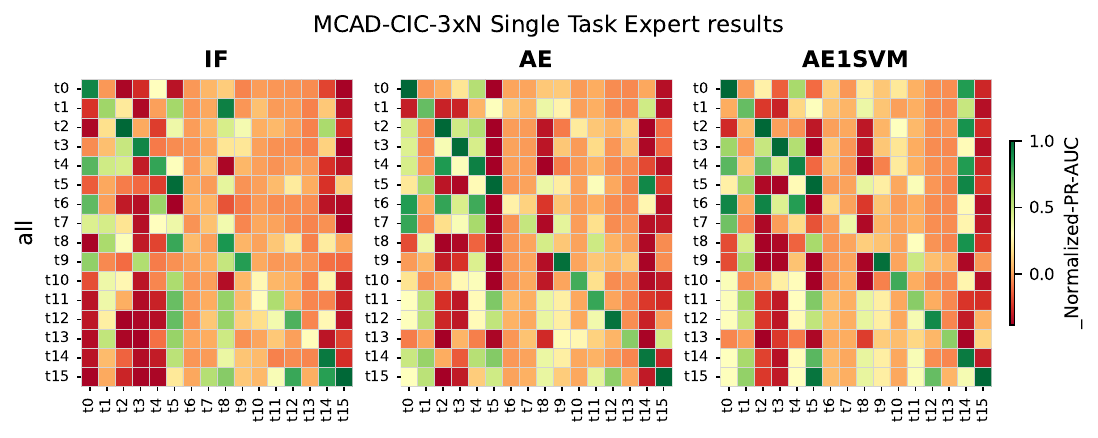}
            \caption{STE evaluation heatmaps for the MCAD-CIC-3xN scenario measured with normalized PR-AUC. The same broad structure persists under imbalance-aware evaluation, further supporting the validity of the retained split.}
            \label{fig:app:3xn_ste_pr_auc}
        \end{figure}

\FloatBarrier

\end{document}